\documentclass[a4paper, 12pt, twoside]{report}

\usepackage[left=2.5cm, right=2.5cm, top=2.5cm, bottom=2.5cm, head=15pt]{geometry} 
\usepackage{graphicx}
\usepackage{amsmath}
\usepackage{amsfonts}
\usepackage{imakeidx}
\usepackage{natbib}
\usepackage[utf8x]{inputenc}
\usepackage{fancyhdr}
\usepackage{float}
\usepackage{wrapfig}
\usepackage{tikz}
\usepackage{tikzpagenodes}
\usepackage{xcolor}
\usepackage{colortbl}
\usepackage{adjustbox}
\usepackage{breqn}	
\usepackage{soul}
\usepackage{framed}
\usepackage{multirow}
\usepackage{enumitem}
\usepackage{setspace}
\usepackage{hyperref}
\usepackage{calc}
\usepackage{makecell}
\usepackage{amssymb}
\usepackage{array}
\usepackage{tabularx}
\usepackage{arydshln}
\usepackage{siunitx}
\usepackage{algorithm}
\usepackage{csquotes}
\usepackage[noend]{algpseudocode}
\usepackage{xspace}

\algnewcommand\algorithmicforeach{\textbf{for each}}
\algdef{S}[FOR]{ForEach}[1]{\algorithmicforeach\ #1\ \algorithmicdo}

\usetikzlibrary{arrows}
\usetikzlibrary{arrows.meta}
\usetikzlibrary{shapes.geometric}
\usetikzlibrary{shapes.misc, positioning}
\usetikzlibrary{shapes.multipart, shapes.arrows}
\usetikzlibrary{decorations.pathmorphing}

\newcommand{\eg}{e.g.\@\xspace}
\newcommand{\Eg}{E.g.\@\xspace}
\newcommand{\ie}{i.e.\@\xspace}
\newcommand{\Ie}{I.e.\@\xspace}
\newcommand{\cf}{cf.\@\xspace}

\newcommand{\symbraw}[1]{[\textit{#1}\,]}
\newcommand{\attrraw}[1]{\{\textit{#1}\,\}}
\newcommand{\symb}[1]{$\symbraw{#1}$}
\newcommand{\symbind}[2]{$\symbraw{#1}_{#2}$}
\newcommand{\attr}[1]{$\attrraw{#1}$}

\makeindex

\makeatletter
\let\OldStatex\Statex
\renewcommand{\Statex}[1][3]{%
	\setlength\@tempdima{\algorithmicindent}%
	\OldStatex\hskip\dimexpr#1\@tempdima\relax}
\makeatother

\tikzset{
	treenode/.style = {align=center, inner sep=0pt, text centered,
		font=\sffamily},
	tochild/.style={draw,-latex},
	toparent/.style={draw,latex-},
	noedge/.style={draw,latex-, white},
	blacknode/.style = {treenode, regular polygon,regular polygon sides=6, white, draw=black, fill=black, text=white, text width=1.5em}, 
	greynode/.style = {treenode, regular polygon,regular polygon sides=6, white, draw=black, fill=black!50, text width=1.5em}, 
	whitenode/.style = {treenode, regular polygon,regular polygon sides=6, black, draw=black, text width=1.5em, very thick}, 
	whitenodefill/.style = {treenode, circle, black, draw=black, fill=white, text width=1.5em}, 
	rednode/.style = {treenode, regular polygon,regular polygon sides=6, red, draw=red, text width=1.5em, very thick}, 
	fullrednode/.style = {treenode, regular polygon,regular polygon sides=6, white, draw=white, fill=red, text width=1.5em, very thick},
	block/.style= {draw, rectangle, minimum width=3cm,minimum height=1cm},
	smallblock/.style= {draw, rectangle, minimum width=2cm,minimum height=0.75cm},
}

\newcommand{\vividnet}[7]
{
	\node (CapsBackground) [rounded corners, fill=#2!10!white, draw=#2,thick, minimum width=3cm, minimum height = 4.8cm, text width=6cm, align=center] at (0cm, 0.9cm) {};

	\node[text=#2, text width=3cm, align=center] at (-1.5cm, 0.8cm) {\textit{Semantic Capsules}};
	\node[text=#2, text width=6cm, align=center] at (-1.4cm, -1cm) {\textit{Primitive Capsules}};
	
	\node[text=#2, align=center] at (0cm, 2.8cm) {\textbf{Capsule Network}};
	
	\draw[#2, dashed] (-3.15cm, -0.5cm) -- (3.15cm, -0.5cm);

	\node (CCap1) [regular polygon,regular polygon sides=6, draw=#2, fill=#2!20!white, minimum width=0.5cm, minimum height=0.5cm] at (1.5cm, 1.6cm) {};
	\node (CCap2) [regular polygon,regular polygon sides=6, draw=#2, fill=#2!20!white, minimum width=0.5cm, minimum height=0.5cm] at (1.1cm, 0.8cm) {};
	\node (CCap3) [regular polygon,regular polygon sides=6, draw=#2, fill=#2!20!white, minimum width=0.5cm, minimum height=0.5cm] at (1.9cm, 0.8cm) {};
	\node (CCap4) [regular polygon,regular polygon sides=6, draw=#2, fill=#2!20!white, minimum width=0.5cm, minimum height=0.5cm] at (1.1cm, -0.0cm) {};
	\node (CCap5) [regular polygon,regular polygon sides=6, draw=#2, fill=#2!20!white, minimum width=0.5cm, minimum height=0.5cm] at (1.9cm, -0.0cm) {};
	\node (CCap6) [regular polygon,regular polygon sides=6, draw=#2, fill=#2!20!white, minimum width=0.5cm, minimum height=0.5cm] at (1.5cm, -1.0cm) {};
	\node (CCap7) [regular polygon,regular polygon sides=6, draw=#2, fill=#2!20!white, minimum width=0.5cm, minimum height=0.5cm] at (0.7cm, -1.0cm) {};
	\node (CCap8) [regular polygon,regular polygon sides=6, draw=#2, fill=#2!20!white, minimum width=0.5cm, minimum height=0.5cm] at (2.3cm, -1.0cm) {};
	\draw[#2, scale=0.1] (CCap1) -- (CCap2);
	\draw[#2, scale=0.1] (CCap1) -- (CCap3);
	\draw[#2, scale=0.1] (CCap2) -- (CCap4);
	\draw[#2, scale=0.1] (CCap2) -- (CCap5);
	\draw[#2, scale=0.1] (CCap3) -- (CCap5);
	\draw[#2, scale=0.1] (CCap4) -- (CCap6);
	\draw[#2, scale=0.1] (CCap4) -- (CCap7);
	\draw[#2, scale=0.1] (CCap5) -- (CCap8);
	\draw[#2, scale=0.1] (CCap5) -- (CCap7);

	\node (Pixels) [circle, text=#1, fill=#1!10!white, draw=#1, text width=1.2cm, thick, align=center, minimum height = 2cm] at (-1.5cm, -3.5cm) {Visual\\Input};

	\node (Other) [circle, text=#1, fill=#1!10!white, draw=#1, text width=1.2cm, thick, align=center, minimum height = 2cm] at (1.5cm, -3.5cm) {$\cdots$};
	
	\node (Oracle) [circle, text=#7, fill=#7!10!white, draw=#7, text width=1.2cm, thick, align=center, minimum height = 2cm] at (0cm, 11.5cm) {Oracle/\\Output};

	\node (IntPhys) [rounded corners, fill=#4!10!white, draw=#4, thick, text width=6cm, minimum width=6cm, minimum height = 2.3cm, align=center] at (0cm, 8.35cm) {};
	\node[text=#4, align=center] at (0cm, 9cm) {\textbf{Intuitive Physics}};
	
	\node[text=#4, text width=3cm, align=center] at (-1.4cm, 8cm) {\textit{Interaction Networks}};
	
	\node (IntCap1) [regular polygon,regular polygon sides=6, draw=#4, fill=#4!20!white, minimum width=0.5cm, minimum height=0.5cm] at (0.5cm, 8.2cm) {};
	\node (IntCap2) [regular polygon,regular polygon sides=6, draw=#4, fill=#4!20!white, minimum width=0.5cm, minimum height=0.5cm] at (2.6cm, 8.1cm) {};
	\node (IntCap3) [regular polygon,regular polygon sides=6, draw=#4, fill=#4!20!white, minimum width=0.5cm, minimum height=0.5cm] at (1.6cm, 7.5cm) {};q
	
	\draw [draw=#4,latex'-latex',decorate,decoration={snake,amplitude=.3mm,segment length=1mm,post length=1mm,pre length=1mm}] (IntCap1) -- (IntCap2);
	\draw [draw=#4,latex'-latex',decorate,decoration={snake,amplitude=.3mm,segment length=1mm,post length=1mm,pre length=1mm}] (IntCap2) -- (IntCap3);
	\draw [draw=#4,latex'-latex',decorate,decoration={snake,amplitude=.3mm,segment length=1mm,post length=1mm,pre length=1mm}] (IntCap1) -- (IntCap3);
	
	\node (EpiMem) [rounded corners, fill=#3!10!white, draw=#3, thick, text width=6cm, minimum width=6cm, minimum height = 2.5cm, align=center] at (0cm, 5.25cm) {};
	\node[text=#3, align=center] at (0cm, 6.0cm) {\textbf{Episodic Memory}};

	\node (E1) [fill=#3!20!white, draw=#3, minimum width=0.5cm, minimum height=0.5cm] at (-2.1cm, 5.3cm) {};
	\node (E2) [fill=#3!20!white, draw=#3, minimum width=0.5cm, minimum height=0.5cm] at (-1.1cm, 5.3cm) {};
	\node (E3) [fill=#3!20!white, draw=#3, minimum width=0.5cm, minimum height=0.5cm] at (-0.1cm, 5.3cm) {};
	\node (E4) [fill=#3!20!white, draw=#3, minimum width=0.5cm, minimum height=0.5cm] at (0.9cm, 5.3cm) {};
	\node (E4H) [fill=#3!20!white, dashed, draw=#3, minimum width=0.5cm, minimum height=0.5cm] at (0.9cm, 4.7cm) {}; 
	\node (E5) [fill=#3!20!white, draw=#3, minimum width=0.5cm, minimum height=0.5cm] at (1.9cm, 5.3cm) {};
	\node (E5H) [fill=#3!20!white, dashed, draw=#3, minimum width=0.5cm, minimum height=0.5cm] at (1.9cm, 4.7cm) {};
	
	\draw[tochild, #3] (E1) -- (E2);
	\draw[tochild, #3] (E2) -- (E3);
	\draw[tochild, #3] (E3) -- (E4);
	\draw[tochild, #3, dashed] (E3) -- (E4H);
	\draw[tochild, #3] (E4) -- (E5);
	\draw[tochild, #3, dashed] (E4H) -- (E5H);
	
	\node (E1Cap1) [regular polygon,regular polygon sides=6, fill=#3, minimum width=1cm, minimum height=1cm, scale=0.1] at (-2.1cm, 5.45cm) {};
	\node (E1Cap2) [regular polygon,regular polygon sides=6, fill=#3, minimum width=1cm, minimum height=1cm, scale=0.1] at (-2.25cm, 5.2cm) {};
	\node (E1Cap3) [regular polygon,regular polygon sides=6, fill=#3, minimum width=1cm, minimum height=1cm, scale=0.1] at (-1.95cm, 5.2cm) {};
	\draw[#3, scale=0.1] (E1Cap1) -- (E1Cap2);
	\draw[#3, scale=0.1] (E1Cap1) -- (E1Cap3);
	
	\node (E2Cap1) [regular polygon,regular polygon sides=6, fill=#3, minimum width=1cm, minimum height=1cm, scale=0.1] at (-1.1cm, 5.45cm) {};
	\node (E2Cap3) [regular polygon,regular polygon sides=6, fill=#3, minimum width=1cm, minimum height=1cm, scale=0.1] at (-0.95cm, 5.2cm) {};
	\draw[#3, scale=0.1] (E2Cap1) -- (E2Cap3);
	
	\node (E3Cap1) [regular polygon,regular polygon sides=6, fill=#3, minimum width=1cm, minimum height=1cm, scale=0.1] at (-0.1cm, 5.45cm) {};
	\node (E3Cap2) [regular polygon,regular polygon sides=6, fill=#3, minimum width=1cm, minimum height=1cm, scale=0.1] at (-0.25cm, 5.2cm) {};
	\node (E3Cap3) [regular polygon,regular polygon sides=6, fill=#3, minimum width=1cm, minimum height=1cm, scale=0.1] at (0.05cm, 5.2cm) {};
	\draw[#3, scale=0.1] (E3Cap1) -- (E3Cap2);
	\draw[#3, scale=0.1] (E3Cap1) -- (E3Cap3);
	
	\node (E4Cap1) [regular polygon,regular polygon sides=6, fill=#3, minimum width=1cm, minimum height=1cm, scale=0.1] at (0.9cm, 5.45cm) {};
	\node (E4Cap2) [regular polygon,regular polygon sides=6, fill=#3, minimum width=1cm, minimum height=1cm, scale=0.1] at (0.75cm, 5.2cm) {};
	\node (E4Cap3) [regular polygon,regular polygon sides=6, fill=#3, minimum width=1cm, minimum height=1cm, scale=0.1] at (1.05cm, 5.2cm) {};
	\draw[#3, scale=0.1] (E4Cap1) -- (E4Cap2);
	\draw[#3, scale=0.1] (E4Cap1) -- (E4Cap3);
	
	\node (E5Cap1) [regular polygon,regular polygon sides=6, fill=#3, minimum width=1cm, minimum height=1cm, scale=0.1] at (1.9cm, 5.45cm) {};
	\node (E5Cap2) [regular polygon,regular polygon sides=6, fill=#3, minimum width=1cm, minimum height=1cm, scale=0.1] at (1.75cm, 5.2cm) {};
	\draw[#3, scale=0.1] (E5Cap1) -- (E5Cap2);
	
	\node (E4HCap1) [regular polygon,regular polygon sides=6, fill=#3, minimum width=1cm, minimum height=1cm, scale=0.1] at (0.9cm, 4.85cm) {};
	\node (E4HCap3) [regular polygon,regular polygon sides=6, fill=#3, minimum width=1cm, minimum height=1cm, scale=0.1] at (1.05cm, 4.6cm) {};
	\draw[#3, scale=0.1] (E4HCap1) -- (E4HCap3);
	
	\node (E5HCap1) [regular polygon,regular polygon sides=6, fill=#3, minimum width=1cm, minimum height=1cm, scale=0.1] at (1.9cm, 4.85cm) {};
	\node (E5HCap2) [regular polygon,regular polygon sides=6, fill=#3, minimum width=1cm, minimum height=1cm, scale=0.1] at (1.75cm, 4.6cm) {};
	\node (E5HCap3) [regular polygon,regular polygon sides=6, fill=#3, minimum width=1cm, minimum height=1cm, scale=0.1] at (2.05cm, 4.6cm) {};
	\draw[#3, scale=0.1] (E5HCap1) -- (E5HCap2);
	\draw[#3, scale=0.1] (E5HCap1) -- (E5HCap3);
	
	\node (Agent) [rounded corners, fill=#6!10!white, draw=#6, thick, minimum width=2cm, minimum height = 11cm, align=center] at (5cm, 4cm) {};
	
	\node[text=#6, align=center, rotate=-90] at (5.5cm, 4cm) {\textbf{Querying}};
	
	\node (Predi) [text=#6, align=center, rotate=-90] at (4.3cm, 8.35cm) {\footnotesize\textit{Predict}};
	
	\node (Reca) [text=#6, align=center, rotate=-90] at (4.3cm, 5.25cm) {\footnotesize\textit{Replay}};	
	
	\node (Simula) [text=#6, align=center, rotate=-90] at (4.3cm, 0.9cm) {\footnotesize\textit{Fabricate}};

	\node (MetaLearn) [rounded corners, fill=#5!10!white, draw=#5, thick, minimum width=2cm, minimum height = 11cm, align=center] at (-5cm, 4cm) {};
	
	\node[text=#5, align=center, rotate=90] at (-5cm, 4cm) {\textbf{Meta-Learning}};

	\draw[tochild, #1] (Pixels) -- (-1.5cm, -1.5cm);
	\draw[tochild, #1] (Other) -- (1.5cm, -1.5cm);
	\draw[tochild, #2] (0cm, 3.3cm) -- (0cm, 4cm);
	\draw[tochild, #4] (0.05cm, 6.5cm) -- (0.05cm, 7.2cm);
	\draw[toparent, #4] (-0.05cm, 6.5cm) -- (-0.05cm, 7.2cm);
	
	\draw[draw, tochild, #6] (4cm, 8.4cm) -- (3.15cm,  8.4cm);
	\draw[draw, toparent, #6] (4cm, 8.3cm) -- (3.15cm,  8.3cm);
	\draw[draw, tochild, #6] (4cm, 5.3cm) -- (3.15cm,  5.3cm);
	\draw[draw, toparent, #6] (4cm, 5.2cm) -- (3.15cm,  5.2cm);
	\draw[draw, tochild, #6] (4cm, 0.95cm) -- (3.15cm,  0.95cm);
	\draw[draw, toparent, #6] (4cm, 0.85cm) -- (3.15cm,  0.85cm);
	
	\draw[draw, tochild, #5] (-4cm, 8.4cm) -- (-3.15cm,  8.4cm);
	\draw[draw, toparent, #5] (-4cm, 8.3cm) -- (-3.15cm,  8.3cm);
	\draw[draw, tochild, #5] (-4cm, 5.3cm) -- (-3.15cm,  5.3cm);
	\draw[draw, toparent, #5] (-4cm, 5.2cm) -- (-3.15cm,  5.2cm);
	\draw[draw, tochild, #5] (-4cm, 0.95cm) -- (-3.15cm,  0.95cm);
	\draw[draw, toparent, #5] (-4cm, 0.85cm) -- (-3.15cm,  0.85cm);

	\draw[draw, #5] (-5.05cm, 9.5cm) -- (-5.05cm, 11.55cm);
	\draw[draw, tochild, #5] (-5.05cm, 11.55cm) -- (-1cm, 11.55cm);
	\draw[draw, toparent, #5] (-4.95cm, 9.5cm) -- (-4.95cm, 11.45cm);
	\draw[draw, #5] (-4.95cm, 11.45cm) -- (-1cm, 11.45cm);
	
	\draw[draw, #6] (5.05cm, 9.5cm) -- (5.05cm, 11.55cm);
	\draw[draw, tochild, #6] (5.05cm, 11.55cm) -- (1cm, 11.55cm);
	\draw[draw, toparent, #6] (4.95cm, 9.5cm) -- (4.95cm, 11.45cm);
	\draw[draw, #6] (4.95cm, 11.45cm) -- (1cm, 11.45cm);
}

\title{A Neural-Symbolic Framework for Mental Simulation}

\newcommand{\thesisauthor}{Michael Kissner}
\newcommand{\supervisor}{Prof. Dr.-Ing. Helmut Mayer}
\newcommand{\cosupervisor}{Prof. Dr. rer. nat. Martin Werner}

\begin{document}
	
%%%%%%%%%%%%%%%%%%%%%%%%%%%%%%%%%%%%%%%%%%%%%%%%%%%%%%%%%%%%%
%TITLE PAGE
%%%%%%%%%%%%%%%%%%%%%%%%%%%%%%%%%%%%%%%%%%%%%%%%%%%%%%%%%%%%%
\makeatletter
\begin{titlepage}
    \begin{center}
        \vspace*{1cm}

		\begin{figure}[htbp]
			\centering
			\includegraphics[width=.5\linewidth]{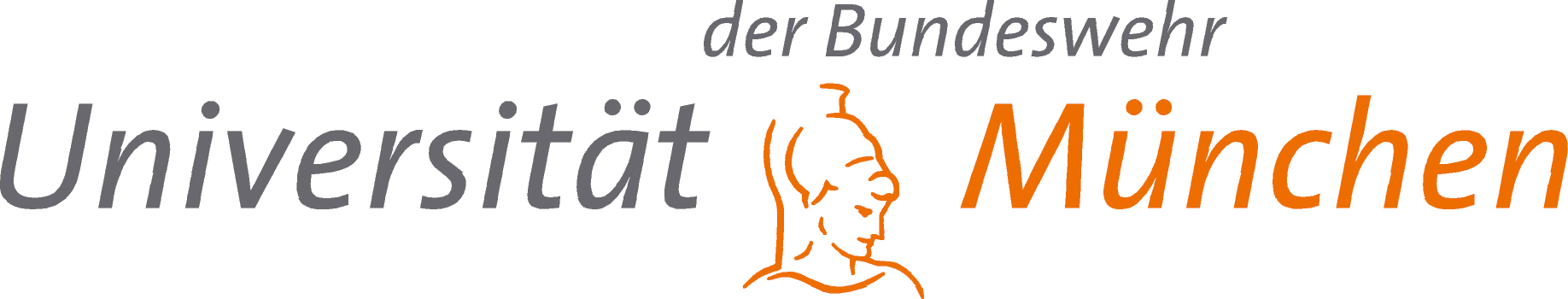}
		\end{figure}

        \Large
        \textbf{\@title}

        \vspace{1.5cm}
        
        \thesisauthor{}
        
        \vspace{1.5cm}
        
    \end{center}

    \large
    \noindent Vollständiger Abdruck der von der Fakultät für Informatik der Universität der Bundeswehr München zur Erlangung des akademischen Grades eines
    
    %\vspace{1cm}
    
    \begin{center}
    	\textbf{Doktors der Naturwissenschaften (Dr. rer. nat.)}
    \end{center}
    
    %\vspace{1cm}
    
    \large
    \noindent genehmigten Dissertation.
    
    \vspace{1cm}

    \vspace{1cm}

    \large
    \noindent Gutachter:
    
    \begin{enumerate}
    	\item \supervisor{}
    	\item \cosupervisor{}
    \end{enumerate}

    \vspace{2cm}
    
    \large
    %\noindent Dissertation wurde am XXXX bei der Universität der Bundeswehr München eingereicht und durch die Fakultät für Informatik am XXX angenommen. Die mündliche Prüfung fand am XXX statt.

\end{titlepage}
\makeatother

\newcommand{\firstIndexing}{\pagestyle{fancy}
	\fancyhf{}
	\fancyhead[LE,RO]{\thepage}}

\newcommand{\secondIndexing}{\pagestyle{fancy}
	\fancyhf{}
	\fancyhead[RE]{\rightmark}
	\fancyhead[LO]{\leftmark}
	\fancyfoot[RE,LO]{\thepage}
	\renewcommand{\footrulewidth}{0.4pt}}

\clearpage

\section*{Abstract}

We present a neural-symbolic framework for observing the environment and continuously learning visual semantics and intuitive physics to reproduce them in an interactive simulation. The framework consists of five parts, a neural-symbolic hybrid network based on capsules for inverse graphics, an episodic memory to store observations, an interaction network for intuitive physics, a meta-learning agent that continuously improves the framework and a querying language that acts as the framework's interface for simulation. By means of lifelong meta-learning, the capsule network is expanded and trained continuously, in order to better adapt to its environment with each iteration. This enables it to learn new semantics using a few-shot approach and with minimal input from an oracle over its lifetime. From what it learned through observation, the part for intuitive physics infers all the required physical properties of the objects in a scene, enabling predictions. Finally, a custom query language ties all parts together, which allows to perform various mental simulation tasks, such as navigation, sorting and simulation of a game environment, with which we illustrate the potential of our novel approach.

\clearpage

\section*{Zusammenfassung}

Wir präsentieren einen neuro-symbolischen Rahmen, für die Beobachtung der Umgebung und kontinuierliches lernen von visueller Semantik und intuitiver Physik, um diese dann in einer interaktiven Simulation wiederzugeben. Der vorgestellte Rahmen besteht aus fünf Teilen, einem neuro-symbolischen hybriden Netzwerk basierend auf Capsules für inverse Grafik, einem episodischen Gedächtnis zum Speichern von Beobachtungen, einem Interaktionsnetz für intuitive Physik, einem Meta-Lernagenten zur kontinuierlichen Verbesserung des Rahmens und einer Abfragesprache, die als Schnittstelle des Rahmens für Simulationen fungiert. Durch lebenslanges Meta-Lernen wird das Capsule Netzwerk ständig erweitert und trainiert, sodass es sich mit jedem Schritt besser an seine Umgebung anpasst. Dies ermöglicht das Lernen von neuer Semantik anhand von wenigen Beispielen und minimaler Eingabe von einem Orakel über die gesamte Lebenszeit. Aus dem aus Beobachtungen Gelernten leitet der Teil für intuitive Physik alle benötigten physischen Eigenschaften der Objekte in einer Szene ab und ermöglicht somit Vorhersagen. Schließlich werden alle Teile durch eine angepasste Abfragesprache zusammengeführt, womit verschiedene mentale Simulationen ausgeführt werden können, wie Navigation, Sortierung und die Simulation einer Computerspielumgebung, anhand derer wir das Leistungsvermögen unseres neuen Ansatzes demonstrieren.
\clearpage

\section*{Danksagung}

Besonders möchte ich meinem Doktorvater Herrn Univ.-Prof. Dr. Helmut Mayer für sein entgegengebrachtes Vertrauen und die Möglichkeit der Promotion danken. Durch seine Betreuung und Unterstützung wurde mir ein sehr interessantes Forschungsthema ermöglicht. Aus unseren Gesprächen habe ich immer viel gelernt und neue Blickwinkel entdeckt für meine Arbeit.

Ebenfalls herzlich danken möchte ich Herrn Univ.-Prof. Dr. Martin Werner für seine Bereitschaft das Zweitgutachten zu übernehmen. Sein tiefer Einblick in das Thema hat mir stets geholfen frühzeitig mögliche Probleme zu identifizieren und eine Lösung zu finden.

Natürlich möchte ich auch allen derzeitigen und ehemaligen Angehörigen der Professur für Visual Computing und Professur für Geoinformatik danken für die interessanten Diskussionen und deren Hilfe während meiner Promotion.

Mein besonderer Dank gilt meiner Partnerin, meinen Eltern, meiner Schwester und meinen Freunden auf dessen Unterstützung und Verständnis ich während dieser Zeit immer zählen konnte.

\clearpage

\newpage\null\thispagestyle{empty}\newpage

\pagenumbering{roman}

\firstIndexing

\tableofcontents

\clearpage

\setlength{\parindent}{0em}
\setlength{\parskip}{1em}

\pagenumbering{arabic}

\secondIndexing

% Open Questions (Perhaps in the Future Work section):
% 
% - On a macroscopic level, are we inverting a bayesian network? No, because bayesian network assumes full knowledge a priori?
% - How does Our approach help with Open-Set Problem? \citep{Geng:2018}
% - Intuitive Physics as Capsule? activation prob of int phys caps?
% - Causal reasoning, did A cause B?
% - \citep{Kellman:1983} - Perception of partly occluded objects in infancy
% - How does our Model generalize from Data?
% - Categorize Capsules by Type? Word2Vec-style. Maybe even integrate it for querying? Incorporate Attributes for categorization? (i.e. "can fly" -> Bird)
% - Sort capsules spacially/geometrically -> Word2Vec and Viewpoints via route on a sphere.
% - Entropy of Configuration States?
% - would \subsection{Fundamental Group} help in identifying entanglement? Seems a bit much...
% - Bayesian Formulation of p in the routing-by-agreement protocol?
% - Augmentation by degree of freedom observed, \ie, dimension?

\chapter{Introduction}\thispagestyle{empty}

In order for a robot or an intelligent agent to plan its next move, it needs to be able to simulate the environment and possible outcomes to make informed decisions. This process is referred to as mental simulation and resembles a game engine in many ways, as it allows the agent to explore and interact like a player would in a game. And just as in gaming, for very demanding tasks the simulation environment needs to accurately reflect reality, be it trough physical accuracy or semantic content, so that the agent is able to perform adequate planning. 

Our motivation and one of the open problems in deep learning is to find a way to learn to perform mental simulations based on observations, by constructing an inverse game engine \citep{Battaglia:2013, Ullman:2017} and trying to understand the world in a generative way. The ability to do so is not only interesting from an artificial intelligence (AI) point of view, but also from a cognitive science perspective, as it is believed that this might resemble and, thus, help to understand the way humans are able to perform mental simulations.

A mental simulation engine needs to perform three tasks: First, it has to be able to extract, store and reproduce all the required semantic information from its environment. Second, it needs the ability to modify and visualize properties of its observation. And third, it must be able to perform intuitive physics. While executing each of these tasks in isolation has proven to be possible with adequate results, merging them has been quite problematic. This is due to the fact that each of the tasks operates on and produces different or insufficient representations of information not suitable for the others. \Eg, modifying observations and intuitive physics work most reliably on a symbolic representation with attributes, preferably in a graph or tree structure, which current visual extraction processes can only partially deliver.

Ideally, a unified structure for visual information should be chosen on which all aspects of a mental simulation pipeline can operate reliably. Modern game engines, such as Godot by the \citet{Team:2019}, give a hint at what might be a good candidate for this structure: a scene-graph.

Scene-graphs consist of a tree-structure, where the root node represents the scene as a whole and each hierarchy of descendants the parts or parts-of-parts of the parent node, \ie, objects that appear in the scene. For a game engine to be able to render this tree, each node is equipped with attributes that describe the object it represents. This graph is easy to manipulate and apply intuitive physics on. The key we have identified in creating a mental simulation engine is, thus, defining a deep learning algorithm that is capable of producing such a scene-graph from an image.

\section{Problem Statement}

In this thesis, we approach the creation of a mental simulation framework capable of interacting with a planning agent. We subdivide this goal and identify four major problem domains that need to be solved, so that it is in fact usable by any type of planning agent. We provide a solution that addresses all of the following problems:

\begin{enumerate}
	\item \textbf{Unified Semantic Information:} A suitable vision algorithm must be devised that is capable of producing a scene-graph usable by the other systems in the framework.
	\item \textbf{Querying and Simulating:} A simulation pipeline must be defined that the agent may interface with and query.
	\item \textbf{Explainability:} The framework must be able to present explainable results, so that an agent is capable of comprehending why specific events occur in a simulation.
	\item \textbf{Lifelong Meta-Learning:} The framework must be able to continuously learn and adapt to new environments.
\end{enumerate}

We note that these points are tightly linked. The ability to simulate and query the result of a simulation requires a semantically rich data structure. As a rich data structure needs to be produced, the algorithm generating this structure should intuitively become more capable in explaining its decisions. And as the algorithm becomes more capable to explain, the path to tackle meta-learning should become more clear.

Through the intertwined nature of the problems, they represent the minimal set of issues we need to contend with in order to define an effective mental simulation framework. Thus, when approaching the definition of such a framework, we must constantly evaluate the interoperability of the domains to ensure that the final solution is coherent.

\section{Approach and Thesis Outline}

% IMPORTANT: Labels didn't work properly, so they are just hard coded

After giving a short introduction to the mathematical concepts we require throughout the thesis and a brief overview of related work, we follow the path outlined by the problem statement. Due to the broad nature of the subject, we focus mainly on toy examples throughout the thesis. 

In Chapter 4 we begin by describing a generative grammar capable of producing a parse-tree that closely resembles a classical scene-graph found in game engines. This grammar is used as a guide to design an inverse-graphics pipeline and the attempt to invert it. The result is comparable to a capsule network \citep{Hinton:2011, Sabour:2017} and we make use of the same terminology, but modify the inner workings of the individual capsules in order to generate a visual parse-tree from an image. The resulting parse-trees are stored as observations in an episodic memory.

The final structure of our neural-symbolic capsule network is highly modular and allows for a symbolic interpretation for each of its parameters. In Chapter 5 we use this explainability to devise a meta-learning algorithm that continuously improves the network with each new scene it encounters in a few-shot approach. This meta-learning process modifies existing capsules, adds new ones and re-routes the flow of information inside the network nearly autonomously, requiring only occasional feedback from an oracle.

With all observations stored in memory we are able to perform simple tracking and introduce an intuitive physics pipeline based on interaction networks \citep{Battaglia:2016} in Chapter 6 that is capable of predicting future events from past observations. We find that by closely analyzing all the semantics in memory we are able to deduce useful physical properties, increasing the versatility of intuitive physics by including complex physical interactions. With this additional information, the original parse-tree generated by the capsule network becomes a full-fledged scene-graph in the game engine sense.

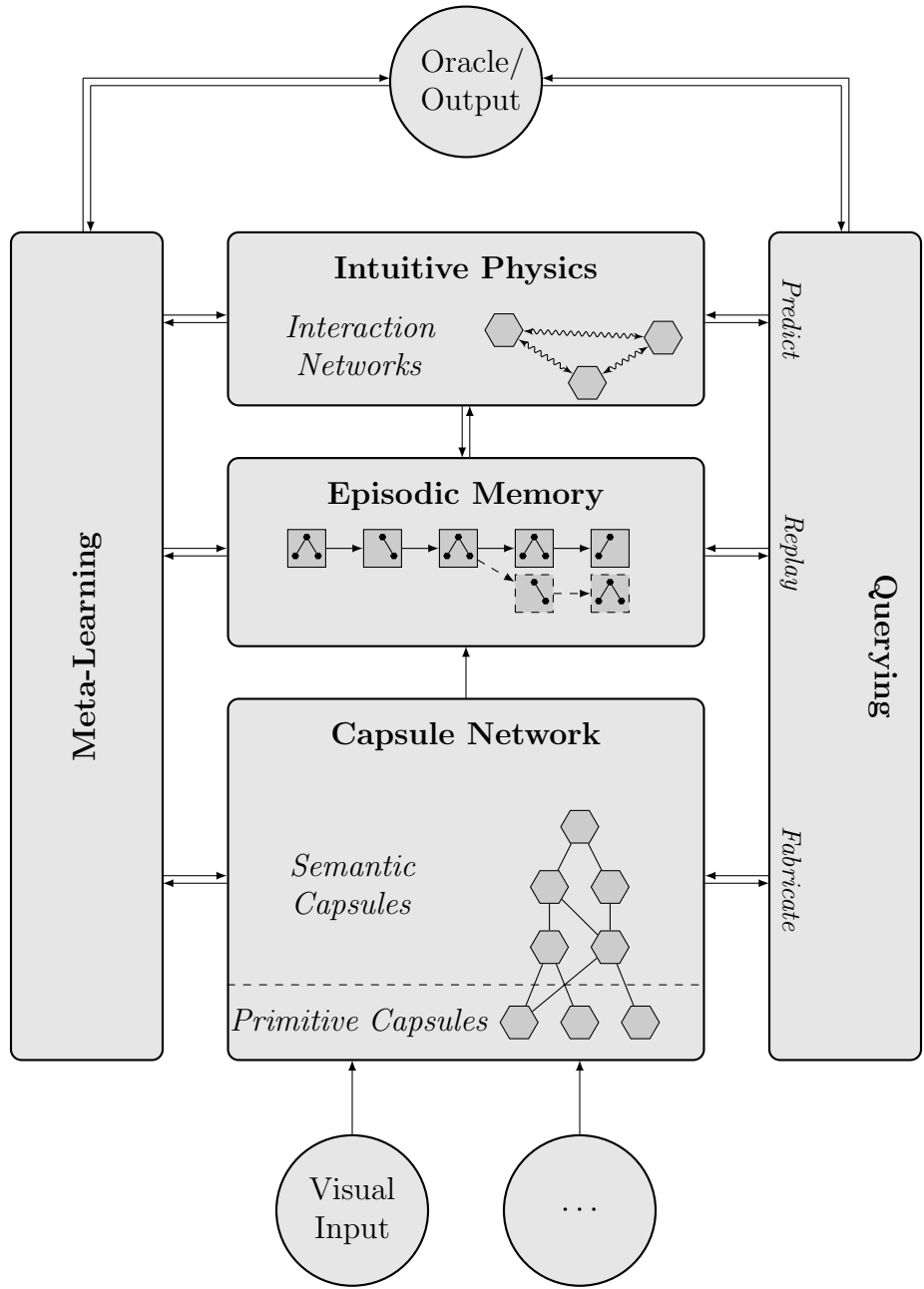
\begin{figure}[H]
	\centering
	\begin{adjustbox}{max width=1\textwidth}
		\begin{tikzpicture}

		\vividnet{black}{black}{black}{black}{black}{black}{black}
		
		\end{tikzpicture}
	\end{adjustbox}
	\caption{The final VividNet framework. Chapter 4 introduces the capsule network and episodic memory, Chapter 5 discusses the meta-learning pipeline, Chapter 6 adds intuitive physics and Chapter 7 covers the querying process.} \label{fig:AIArchitectureIntro}
\end{figure}

Capsule network, episodic memory, meta-learning pipeline and intuitive physics are tied together to form a mental simulation pipeline. However, in this raw state it is still not usable by a planning agent. In Chapter 7 we, therefore, devise a custom querying language that makes the process of mental simulation more explicit by providing a streamlined interface and demonstrate some of many possible use-cases. Our resulting framework and all its components, named VividNet, are depicted in Figure \ref{fig:AIArchitectureIntro}.

In Chapter 8 we discuss the advantages and shortcomings of our approach. We focus on the comparison to artificial neural networks and highlight how our approach addresses some of the issues found in the classical approach, such as the lack of explainability and the binding problem. Finally, a conclusion referring to our problem statement and an outlook are given.

\clearpage

\chapter{Mathematical Preliminaries}\thispagestyle{empty}

For our discussion of the meta-learning pipeline, we borrow the language used primarily in two areas of mathematics: topology and group theory. We use topology\index{Topology} for a deeper understanding of the structure of an object's configuration space, consisting of its attributes. The attributes themselves also play an important role in our analysis, for which we employ group theory to study their behavior under transformations. In this chapter we present these concepts and the notation we use.

\section{Quotient Spaces}

To introduce quotient topologies\index{Quotient Topology}, we need the notion of a topological space\index{Topological Space}. A topological space $(X,\mathcal{T}_X)$ is a set $X$ and a collection $\mathcal{T}_X$ of subsets of $X$ that satisfy the following propositions \citep{Nakahara:2003}: 

\begin{enumerate}
	\item The empty set $\emptyset$ and $X$ are in the collection $\mathcal{T}_X$ ($\emptyset, X \in \mathcal{T}_X$).
	\item The finite or infinite union of subsets found in $\mathcal{T}_X$ also belongs to $\mathcal{T}_X$ (any number of $U_i\in\mathcal{T}_X$ satisfy $\bigcup_i U_i \in \mathcal{T}_X$).
	\item The intersection of finite subsets found in $\mathcal{T}_X$ also belongs to $\mathcal{T}_X$ (finite $U_i\in\mathcal{T}_X$ satisfy $\bigcap_i U_i \in \mathcal{T}_X$).
\end{enumerate}

The collection $\mathcal{T}_X$ defines the topology and we refer to the individual subsets $U_i$ as open sets. 

\begin{figure}[H]
	\centering
	\begin{tikzpicture}
	
	\node[inner sep=0pt] (iris) at (3.0,3.0) {\includegraphics[width=0.3\textwidth]{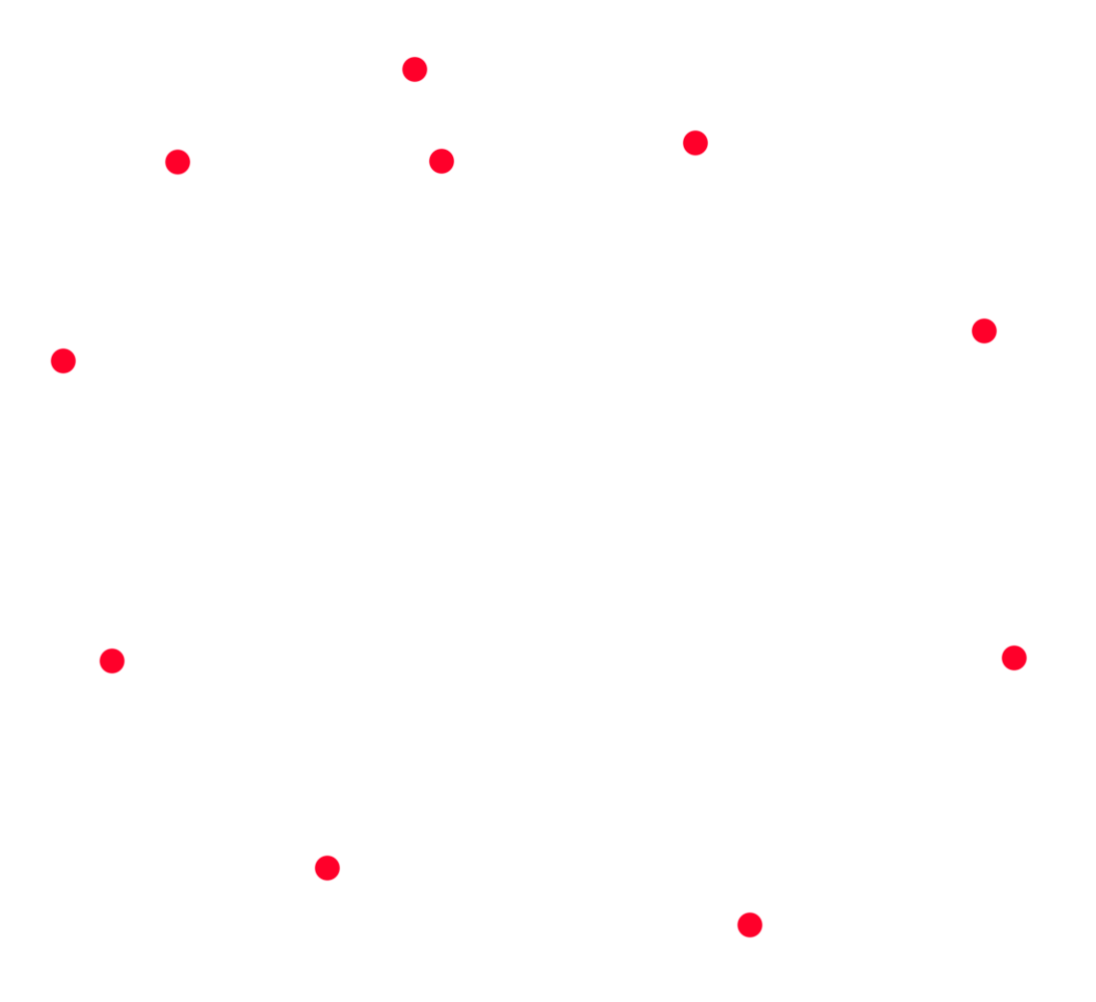}};
	
	\node[inner sep=0pt] (iris) at (11.0,3.0) {\includegraphics[width=0.3\textwidth]{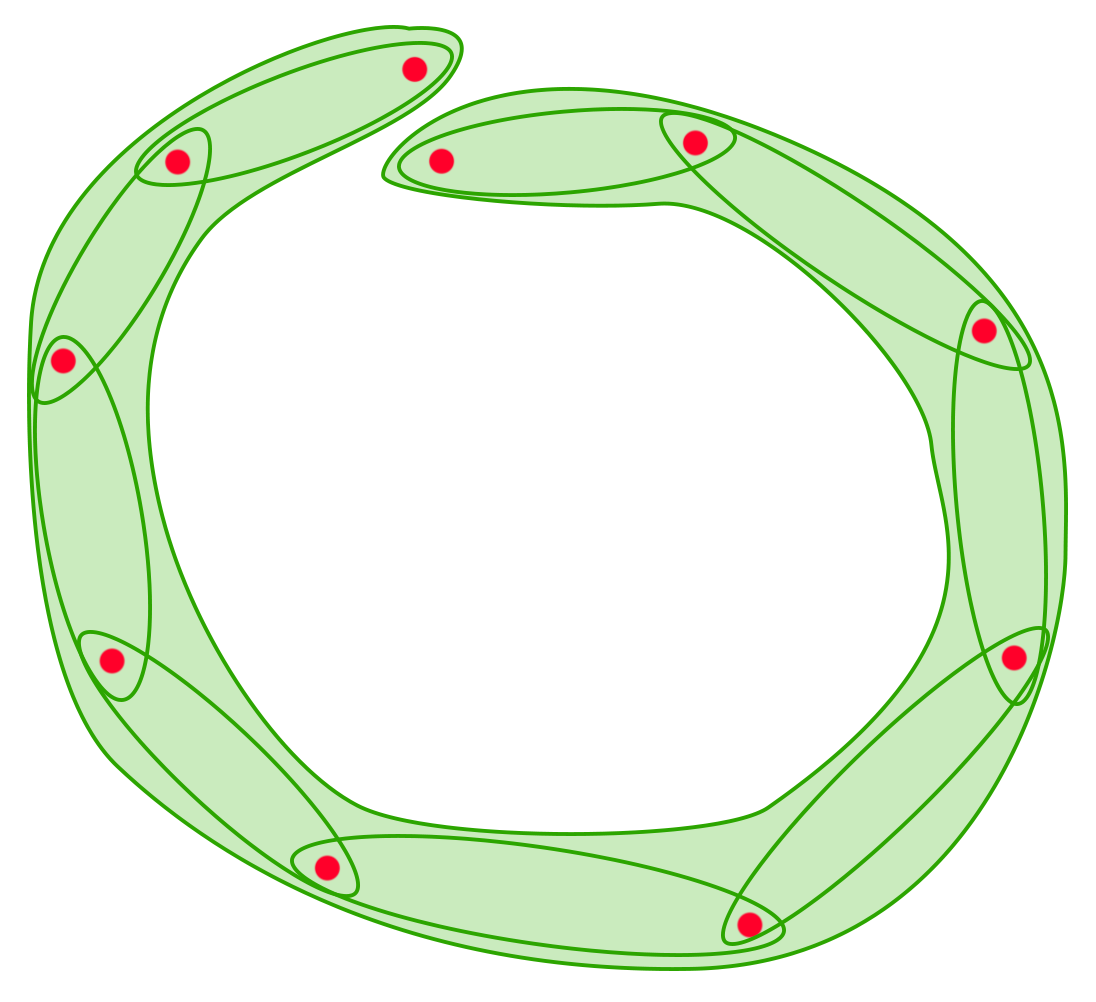}};
	
	\draw[thick,->] (0.0,0.0) -- (0.0,6.0); % node[right] {$x$};
	\draw[thick,->] (0.0,0.0) -- (6.0,0.0); % node[right] {$x$};
	
	\draw[thick,->] (8.0,0.0) -- (8.0,6.0); % node[right] {$x$};
	\draw[thick,->] (8.0,0.0) -- (14.0,0.0); % node[right] {$x$};
	
	%\draw [step=1.0,blue, thick] (0.0,0.0) grid (14.0,6.0);
	%\draw [step=0.1,gray] (0.0,0.0) grid (14.0,6.0);
	
	\node at (2.5+8.0, 3.8) {\small $a_0$};
	\node at (3.5+8.0, 3.8) {\small $a_1$};
	\node at (4.2+8.0, 3.4) {\small $a_2$};
	\node at (4.2+8.0, 2.3) {\small $a_3$};
	\node at (3.8+8.0, 0.7) {\small $a_4$};
	\node at (2.0+8.0, 0.9) {\small $a_5$};
	\node at (0.7+8.0, 2.0) {\small $a_6$};
	\node at (0.4+8.0, 3.7) {\small $a_7$};
	\node at (0.9+8.0, 4.7) {\small $a_8$};
	\node at (2.4+8.0, 5.4) {\small $a_9$};
	
	\node at (2.5, 3.8) {\small $a_0$};
	\node at (3.5, 3.8) {\small $a_1$};
	\node at (4.2, 3.4) {\small $a_2$};
	\node at (4.2, 2.3) {\small $a_3$};
	\node at (3.8, 0.7) {\small $a_4$};
	\node at (2.0, 0.9) {\small $a_5$};
	\node at (0.7, 2.0) {\small $a_6$};
	\node at (0.4, 3.7) {\small $a_7$};
	\node at (0.9, 4.7) {\small $a_8$};
	\node at (2.4, 5.4) {\small $a_9$};
	
	\node[align=center, text width=6cm] at (3.0, -1.0) {\footnotesize{Point Cloud $X$}};
	
	\node[align=center, text width=6cm] at (11.0, -1.0) {\footnotesize{Topology of $(X,\mathcal{T}_X)$ \\(only selected elements of $\mathcal{T}_X$ are shown)}};
	
	\end{tikzpicture}
	\caption{The topology of the dataset influences its interpretation. Here, one of many possible topologies for the data (left) is shown (right) in which $a_9$ can only be reached from $a_0$ by traversing through all other points first, instead of having a direct connection.} \label{fig:Quotient1}
\end{figure}
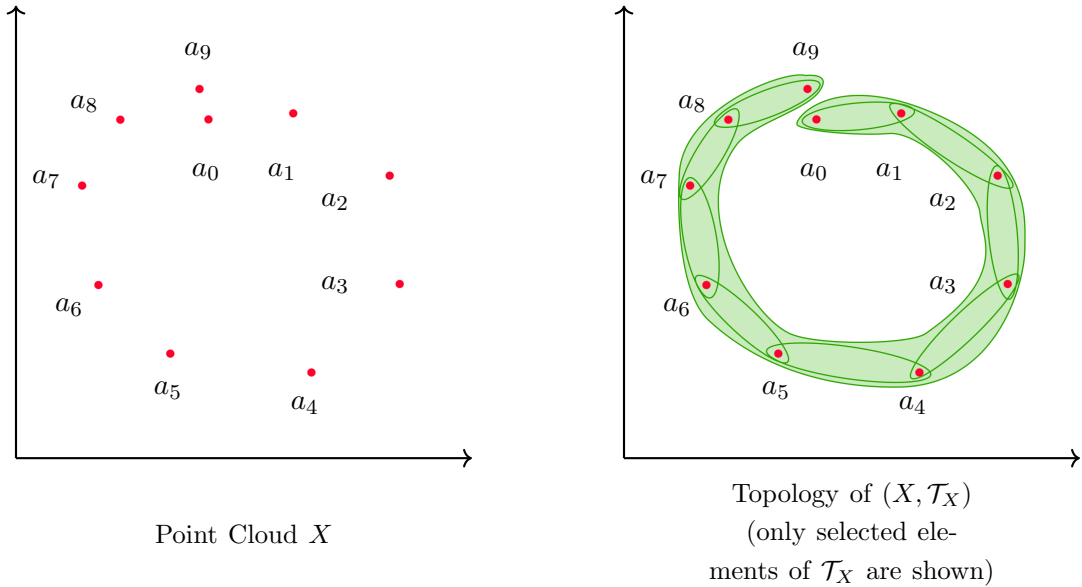

As an example, consider gathering individual data points $a_i$ for training purposes. These points form a point cloud in some $\mathbb{R}^n$ and we treat them as the set $X$. In most machine learning algorithms, such as manifold learning, one of the goals is to find the neighborhood relations between the individual points, which are vital in performing meaningful inter- and extrapolations. These neighborhood relations can be used to define a topology in the sense of $\mathcal{T}_X$ (\cf Figure \ref{fig:Quotient1}).   

The situation in Figure \ref{fig:Quotient1} is quite ambiguous from a machine learning perspective. It suggests that the data forms an open circle, but it might as well be a closed circle. While the data $X$ remains the same, the assumed topology defined by $\mathcal{T}_X$ for an open or a closed circle are very different and provide a conflicting interpretation. 

We can, however, construct a closed circle out of an open circle by gluing together its ends. The result of gluing together parts of a topological space is referred to as a quotient space\index{Quotient Space}. Such a quotient space is created by identifying points in $X$ with each other, essentially regarding them as equal. In our machine learning example of Figure \ref{fig:Quotient1}, this would mean that we assume that the end-points of the open circle, $a_0$ and $a_9$, actually describe the same point and, thus, close the loop.

\begin{figure}[H]
	\centering
	\begin{tikzpicture}
	
	\node[inner sep=0pt] (iris) at (3.0,3.0) {\includegraphics[width=0.3\textwidth]{Quotient2.png}};

	\node[inner sep=0pt] (iris) at (11.0,3.0) {\includegraphics[width=0.3\textwidth]{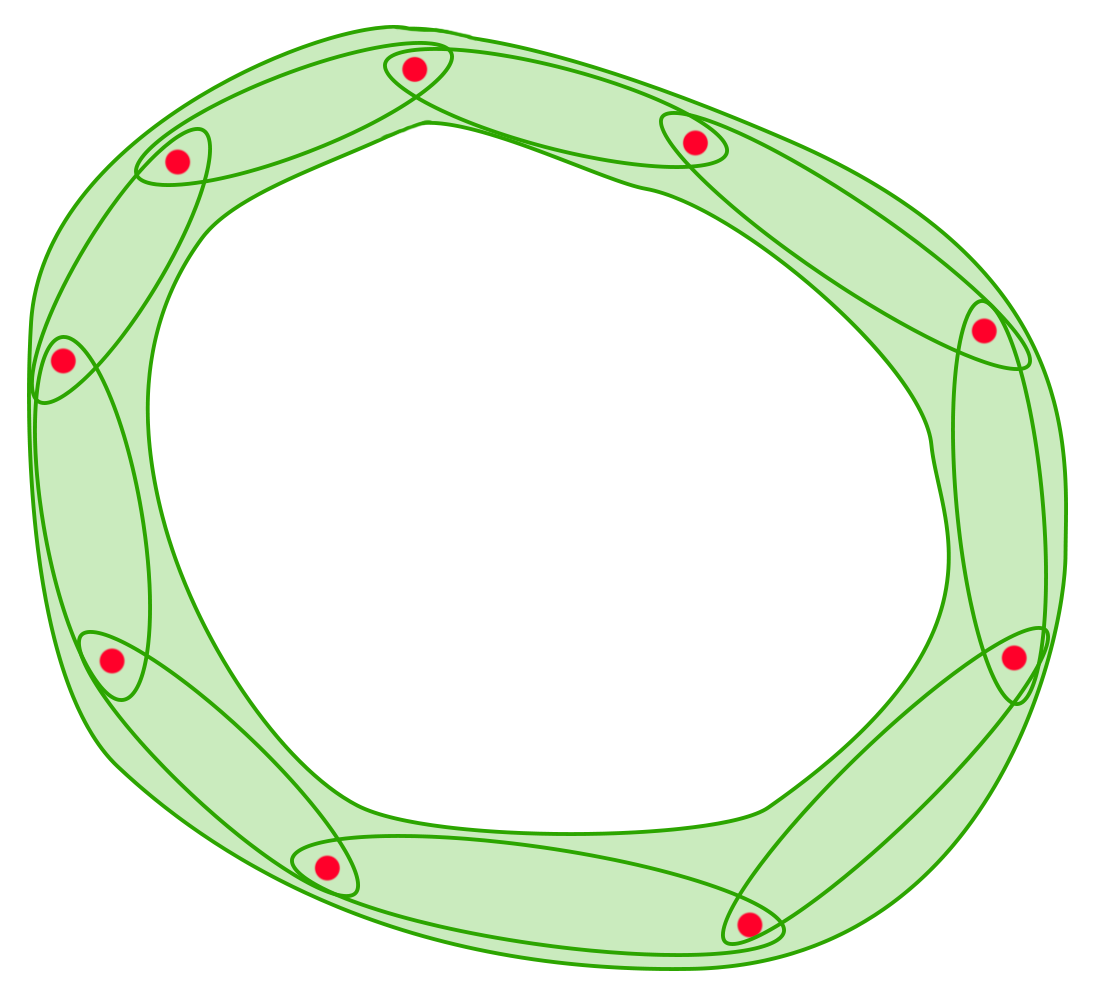}};
	
	\draw[thick,->] (0.0,0.0) -- (0.0,6.0); % node[right] {$x$};
	\draw[thick,->] (0.0,0.0) -- (6.0,0.0); % node[right] {$x$};
	
	\draw[thick,->] (8.0,0.0) -- (8.0,6.0); % node[right] {$x$};
	\draw[thick,->] (8.0,0.0) -- (14.0,0.0); % node[right] {$x$};
	
	\draw[thick,->] (6.0,3.0) -- (7.0,3.0);

	%\draw [step=1.0,blue, thick] (0.0,0.0) grid (14.0,6.0);
	%\draw [step=0.1,gray] (0.0,0.0) grid (14.0,6.0);

	\node at (2.5, 3.8) {\small $a_0$};
	\node at (3.5, 3.8) {\small $a_1$};
	\node at (4.2, 3.4) {\small $a_2$};
	\node at (4.2, 2.3) {\small $a_3$};
	\node at (3.8, 0.7) {\small $a_4$};
	\node at (2.0, 0.9) {\small $a_5$};
	\node at (0.7, 2.0) {\small $a_6$};
	\node at (0.4, 3.7) {\small $a_7$};
	\node at (0.9, 4.7) {\small $a_8$};
	\node at (2.4, 5.4) {\small $a_9$};
	
	%\node at (2.5+8.0, 3.8) {\small $a_0$};
	\node at (3.5+8.0, 3.8) {\small $a_1$};
	\node at (4.2+8.0, 3.4) {\small $a_2$};
	\node at (4.2+8.0, 2.3) {\small $a_3$};
	\node at (3.8+8.0, 0.7) {\small $a_4$};
	\node at (2.0+8.0, 0.9) {\small $a_5$};
	\node at (0.7+8.0, 2.0) {\small $a_6$};
	\node at (0.4+8.0, 3.7) {\small $a_7$};
	\node at (0.9+8.0, 4.7) {\small $a_8$};
	\node at (2.4+8.0, 5.4) {\small $a_9 \sim a_0$};
	
	\node[align=center, text width=6cm] at (3.0, -1.0) {\footnotesize{Topology of $(X,\mathcal{T}_X)$ \\(only selected elements of $\mathcal{T}_X$ are shown)}};
	
	\node[align=center, text width=6cm] at (11.0, -1.0) {\footnotesize{Topology of $(Y,\mathcal{T}_Y)$ \\(only selected elements of $\mathcal{T}_Y$ are shown)}};
	
	\end{tikzpicture}
	\caption{By identifying the points $a_0$ and $a_9$ of $(X,\mathcal{T}_X)$, the loop is closed to form a circle.} \label{fig:Quotient2}
\end{figure}

Points in $X$ are identified using an equivalence relation\index{Equivalence Relation} denoted by $a\sim b$. However, $a\sim b$ does not restrict us to only identify single points with one another as in our example. We may also have an entire equivalence class
\begin{equation}
[a] = \{ x\in X \vert x\sim a \} \;\;\;.
\end{equation}
Using these equivalence classes\index{Equivalence Class}, we introduce the quotient space $Y=X/\sim$, which is a rigorous way of defining our gluing process:
\begin{equation}
Y = \{ [x] \vert x\in X \} \;\;\;.
\end{equation}
Further, the canonical quotient map $q\colon X\to X/\sim$\index{Canonical Quotient Map} also induces a topology on $Y$
\begin{equation}
\mathcal{T}_Y = \{ U \subseteq Y \vert q^{-1}(U) \in \mathcal{T}_X \} \;\;\;,
\end{equation}
making $(Y,\mathcal{T}_Y)$ a topological space. Figure \ref{fig:Quotient2} shows the concrete example of identifying the data points $a_0$ and $a_9$ to form the quotient space $Y = X/\{a_0, a_9\}$.

\begin{figure}[H]
	\centering
	\begin{tikzpicture}
	
	\node[inner sep=0pt] (iris) at (3.0,3.0) {\includegraphics[width=0.3\textwidth]{Quotient3.png}};

	\node[inner sep=0pt] (iris) at (11.0,3.0) {\includegraphics[width=0.3\textwidth]{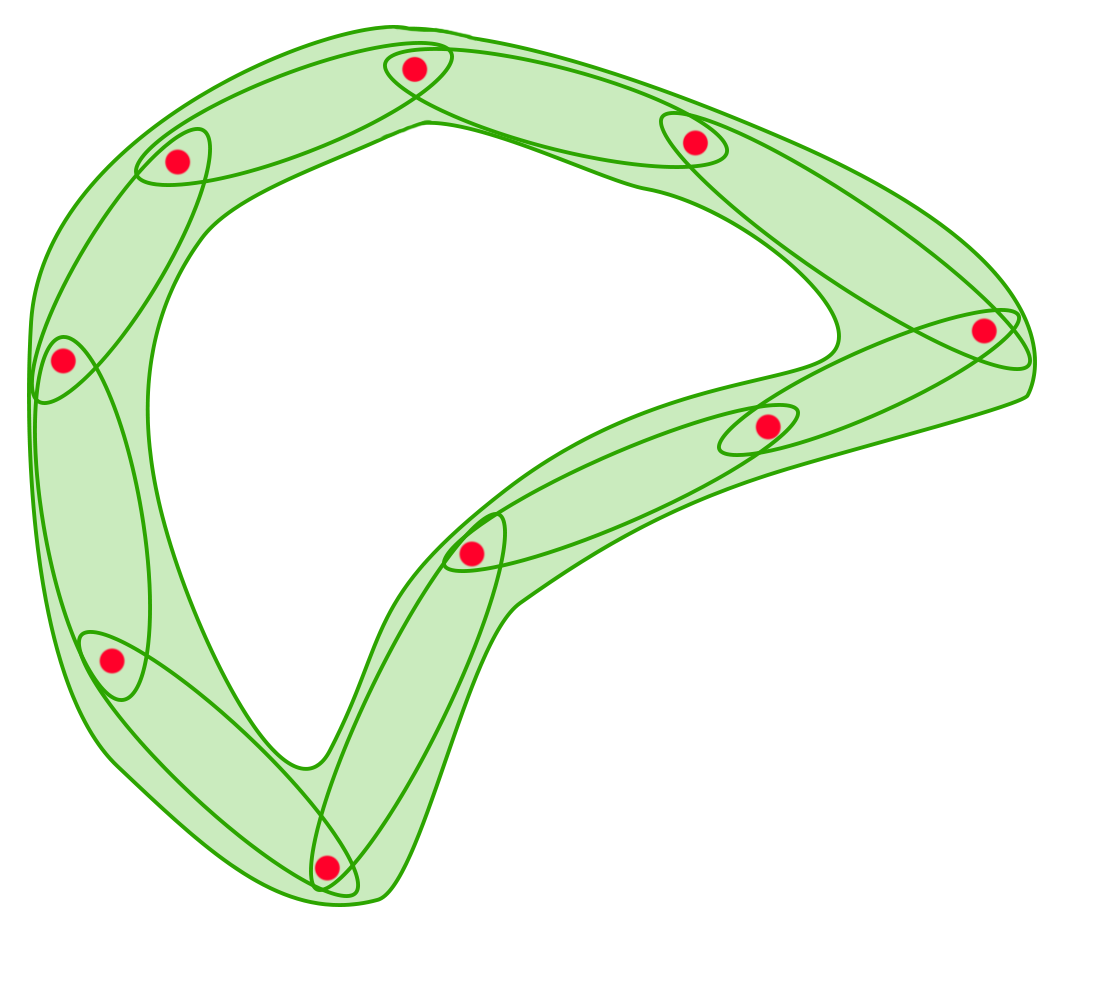}};
	
	\draw[thick,->] (0.0,0.0) -- (0.0,6.0); % node[right] {$x$};
	\draw[thick,->] (0.0,0.0) -- (6.0,0.0); % node[right] {$x$};
	
	\draw[thick,->] (8.0,0.0) -- (8.0,6.0); % node[right] {$x$};
	\draw[thick,->] (8.0,0.0) -- (14.0,0.0); % node[right] {$x$};
	
	\draw[thick,->] (6.0,3.0) -- (7.0,3.0);
	
	%\draw [step=1.0,blue, thick] (0.0,0.0) grid (14.0,6.0);
	%\draw [step=0.1,gray] (0.0,0.0) grid (14.0,6.0);

	%\node at (2.5, 3.8) {\small $a_0$};
	%\node at (3.5, 3.8) {\small $a_1$};
	%\node at (4.2, 3.4) {\small $a_2$};
	%\node at (4.2, 2.3) {\small $a_3$};
	%\node at (3.8, 0.7) {\small $a_4$};
	%\node at (2.0, 0.9) {\small $a_5$};
	%\node at (0.7, 2.0) {\small $a_6$};
	%\node at (0.4, 3.7) {\small $a_7$};
	%\node at (0.9, 4.7) {\small $a_8$};
	%\node at (2.4, 5.4) {\small $a_9 \sim a_0$};
	
	%\node at (2.5+8.0, 3.8) {\small $a_0$};
	%\node at (3.5+8.0, 3.8) {\small $a_1$};
	%\node at (4.2+8.0, 3.4) {\small $a_2$};
	%\node at (4.2+8.0, 2.3) {\small $a_3$};
	%\node at (3.8+8.0, 0.7) {\small $a_4$};
	%\node at (2.0+8.0, 0.9) {\small $a_5$};
	%\node at (0.7+8.0, 2.0) {\small $a_6$};
	%\node at (0.4+8.0, 3.7) {\small $a_7$};
	%\node at (0.9+8.0, 4.7) {\small $a_8$};
	%\node at (2.4+8.0, 5.4) {\small $a_9 \sim a_0$};
	
	\node[align=center, text width=6cm] at (3.0, -1.0) {\footnotesize{Topology of $(Y,\mathcal{T}_Y)$ \\(only selected elements of $\mathcal{T}_Y$ are shown)}};
	
	\node[align=center, text width=6cm] at (11.0, -1.0) {\footnotesize{Topology of $(\hat{Y},\mathcal{T}_{\hat{Y}})$ \\(only selected elements of $\mathcal{T}_{\hat{Y}}$ are shown)}};
	
	\end{tikzpicture}
	\caption{$(Y,\mathcal{T}_Y)$ and $(\hat{Y},\mathcal{T}_{\hat{Y}})$ are said to be homeomorphic. Both are also homeomorphic to the unit circle $S^1$, but not to the interval $[0,1]$.} \label{fig:Quotient3}
\end{figure}

Next, we discuss the concept of homeomorphism\index{Homeomorphism}. From a purely topological standpoint, the value of an individual element of $Y$ does not matter. In our example, this is equivalent to saying that their position in $\mathbb{R}^n$ is of no topological consequence. We can continuously move around the points without changing the topology and get a set $\hat{Y}$, such as in Figure \ref{fig:Quotient3}. 

The two topological spaces $(Y,\mathcal{T}_Y)$ and  $(\hat{Y},\mathcal{T}_{\hat{Y}})$ are said to homeomorphic, as there exists a function $f\colon Y \to \hat{Y}$ which satisfies:
\begin{enumerate}
	\item $f$ is bijective.
	\item $f$ is continuous.
	\item $f^{-1}$ is continuous.
\end{enumerate}
We refer to such a function $f$ as a homeomorphism and use $Y \simeq \hat{Y}$ to denote that two spaces are homeomorphic.

For machine learning purposes, however, the value of the individual elements of $Y$ does matter. We consider a special type of topological space referred to as a metric space $(M, d)$\index{Metric Space} \citep{Burago:2001}. Here, $M$ is a set and $d\colon M \times M \to \mathbb{R}$ a metric or distance function on $M$\index{Distance Function}. This metric satisfies the following conditions:
\begin{enumerate}
	\item $d(x,y) = d(y,x)$.
	\item $d(x,x) = 0$ and $d(x,y) > 0$ for $x\neq y$.
	\item $d(x,y) + d(y,z) \geq d(x,z)$.
\end{enumerate}
We see that this metric space $(M, d)$ induces a natural topology determined by $d$ and the open discs
\begin{equation}
U_{\epsilon}(M) = \{y \in M \vert d(x,y) < \epsilon\} \;\;\;,
\end{equation}
are referred to as the metric topology\index{Metric Topology}.

This induced topology also allows us to form quotient metric spaces\index{Quotient Metric Space} $M/\sim$ on which we define a quotient semi-metric $d_R$\index{Semi-Metric}. We refer to this function as a semi-metric, as it allows for two points $x$ and $y$ with $x\neq y$ to have a vanishing distance $d_R(x,y) = 0$, violating condition 2 of a regular metric. 

To construct $d_R$, we again consider Figure \ref{fig:Quotient2}. If we were to measure the distance between $a_1$ and $a_8$ on our quotient metric space, it seems obvious that we should follow the sequence $(a_1, a_0, a_9, a_8)$ and sum up the individual distances $d(a_1, a_0) + d(a_0, a_9) + d(a_9, a_8)$. However, our original metric does not take into account that $a_0$ and $a_9$ now describe the same point in the quotient space and $d(a_0, a_9) \neq 0$. Instead, to ensure that the distance between $a_0$ and $a_9$ is not counted, we perform a check if there is a shorter path by taking into account all points that are equivalent at each step in the sequence. In our case, we find $a_0 \sim a_9$ and skip this hop in the sequence. To formalize this, we define the quotient semi-metric to be
\begin{equation}
d_R([x], [y]) \colon = \inf \{ \sum_{j=1}^k d(x_j, y_j) \} \;\;\;,
\end{equation}
where we take the infimum over all possible sequences $(x_1, \cdots, x_k)$ and $(y_1, \cdots, y_k)$, with $x_1 \in [x]$, $y_k\in [y]$ and $y_j \sim x_{j+1}$. This ensures that $d_R([x], [y]) \leq d(x,y)$.

Generally, a metric space doesn't have a well defined dimension and we employ methods out of fractal geometry to make an estimate, such as the Hausdorff dimension \citep{Hausdorff:1918} or the box-counting dimension\index{Box-Counting Dimension} \citep{Falconer:1990}. We use a definition similar to the box-counting dimension for our metric space $(M,d)$. We assume $M$ is a subset of some $\mathbb{R}^n$ and consider a space $\hat{M}$, for which there exists a continuous, surjective map $\psi \colon M \to \hat{M}$, such that for every point $p\in M$ we have 
\begin{equation}\label{eq:contract}
\Vert p - \psi(p) \Vert < \rho \;\;\; ,
\end{equation}
where $\Vert \cdot \Vert$ measures the Euclidean distance in $\mathbb{R}^n$. We are essentially allowing points in $\hat{M}$ to rearrange themselves by a small margin. Finally, our contracted box-counting dimension (CBC)\index{Contracted Box-Counting Dimension} is defined as
\begin{equation}
\dim_{CBC} M \colon= \min_{\psi} \lim_{\epsilon\to 0} \frac{\log N_\psi(\epsilon)}{\log 1/\epsilon} \;\;\;,
\end{equation}
where $N_\psi(\epsilon)$ is the number of $\epsilon^n$-sized boxes on an $\epsilon$-spaced grid in $\mathbb{R}^n$ that cover $\hat{M}$. Through this construction $\hat{M}$ is a contraction of $M$ and removes small numerical errors that accidentally generate unnecessary dimensions in our dataset. For practical purposes, we take $\epsilon$ to be a small, yet computable value and continue to the limit until the point the dimension starts to converge. 

% TODO: Example of a line of dots with outliers?

% TODO: ? Image of how box-counting dimension works?

\section{Groups and Representations}

To understand how transformations of the input affects the output of a function, group and representation theory is a helpful tool. By a group\index{Group} \citep{Willwacher:2014} we mean a set $G$ together with a map (referred to as multiplication) of the form
\begin{equation}
\circ \colon G \times G \to G
\end{equation}
which satisfies:
\begin{enumerate}
	\item For $g_1, g_2, g_3 \in G$ we have $(g_1 \circ g_2) \circ g_3 = g_1 \circ (g_2 \circ g_3)$. (\textit{Associativity})
	\item There exists an identity element $\textbf{1}$, such that for any $g\in G$ holds $\textbf{1} \circ g = g \circ \textbf{1} = g$. (\textit{Identity})
	\item For all $g\in G$ there exists an inverse element $g^{-1}\in G$, such that $g \circ g^{-1} = g^{-1} \circ g = \textbf{1}$. (\textit{Inverse})
\end{enumerate}

One example for a group are invertible linear maps\index{Invertible Linear Maps} between vector spaces. In the case of $\mathbb{R}^n\to \mathbb{R}^n$, we have invertible $n \times n$ matrices. We refer to this group as $GL(\mathbb{R}^n)$ or $GL(n)$ in short and as $GL(V)$ for a general vector space $V$. When considering this group in detail, we see that there exist multiple subsets of $n \times n$ invertible matrices, which themselves form a group under the conditions above. One such example is the group of rotations $SO(n)$\index{Group of Rotations} and we refer to a group that satisfies this condition as a subgroup and write $SO(n) \subset GL(n)$\index{Subgroup}.

We also define a product of two groups, $G \times H$, by creating a set of pairs $(g, h)$ and defining the product as $(g_1, h_1)(g_2, h_2) = (g_1 g_2, h_1 h_2)$. In this case, we explicitly see that $G$ and $H$ are subgroups of $G \times H$. 

While a group element of $G$ transforming other elements of $G$ through multiplication is interesting in itself (for example, matrix-matrix multiplication), we are interested in how it behaves when applied to another space (for example, matrix-vector multiplication). We use the term action\index{Group Action} to describe how a group element acts on such a space $X$, \ie, the left group action is a function $\varphi \colon G \times X \to X$ that satisfies:
\begin{enumerate}
	\item For all $x\in X$, we have $\varphi(\textbf{1}, x) = x$. (\textit{Identity})
	\item For all $g,h \in G$ and $x \in X$, we have  $\varphi(gh, x) = \varphi(g, \varphi(h, x))$. (\textit{Compatibility})
\end{enumerate}
With an equivalent set of axioms as the left group action, the right group action $\varphi \colon X \times G \to X$ is defined. With both of these group actions in place, we simplify the $\varphi$ notation and write $g \cdot x$ or $x \cdot g$ to imply an element $g$ acting on $x$ from the left or the right. 

Next, we introduce homomorphisms\index{Homomorphism} between groups, not to be confused with homeomorphisms between topologies. A homomorphism is a map $f\colon G \to H$ that preserves the operations of the structures, \ie, $f(g_1 g_2) = f(g_1)f(g_2)$.

This allows us to define a representation\index{Representation} of a group $G$ on a vector space $V$ as a homomorphism $\rho \colon G \to GL(V)$. Group representations are useful if the group $G$ is an abstract group and we want to study its behavior on a more concrete environment, such as $\mathbb{R}^n$. As an example, if we take the group of 2-dimensional (2D) rotations $SO(2)$, we may represent it as a set of $3\times 3$ matrices that define a rotation around some specific axis in 3-dimensional (3D) space using $\rho \colon SO(2) \to GL(3)$, essentially making it a 3D rotation.

We now study how these group actions can help us to better understand data in a machine learning context following the terminology set out by \citet{Higgins:2018}. The idea of disentanglement\index{Disentanglement} in the context of deep learning is to find attributes or latent variables which act independently of the others. As an example, the \textit{position} of a table is disentangled from its \textit{color}, but \textit{color} is often tightly entangled with other material properties, such as \textit{roughness}. 

To find entanglement\index{Entanglement} in the set of all attributes $A$, we study the effect that the change of a subset $A_1$ has on the remainder $A_2$, where $A = A_1 \oplus A_2$ and by $\oplus$ we mean the direct sum of two attribute sets $\{\alpha_1, \alpha_2\} \oplus \{\alpha_3, \alpha_4\} = \{\alpha_1, \alpha_2, \alpha_3, \alpha_4\}$. Consider a group $G = G_1 \times G_2$ that acts on $A$ by a group action $\cdot \colon G\times A \to A$ and on each $A_i$ independently as $\cdot_i \colon G_i\times A_i \to A_i$, such that
\begin{equation}
(\mu_1, \mu_2) \cdot (\vec{\alpha}_1, \vec{\alpha}_2) = (\mu_1 \cdot_1 \vec{\alpha}_1, \mu_2 \cdot_2 \vec{\alpha}_2) 
\end{equation}
   
for all $\mu_1 \in G_1$ and $\mu_2 \in G_2$. In this case, we refer to $A_1$ and $A_2$ as being disentangled with respect to $G$.

Closely related to the idea of entanglement is equivariance\index{Equivariance}. Consider a map $f \colon X \to Y$. We say that $f$ is equivariant with respect to some group $G$ if 
\begin{equation}\label{eq:equivariant}
f(\rho_X(\mu)\cdot x) = \rho_Y(\mu) \cdot f(x)
\end{equation}

is true for all $\mu\in G$ and all $x \in X$ and where $\rho_X:G\to GL(X)$ and $\rho_Y:G\to GL(Y)$ are representations of $G$ on $X$ and $Y$. Thus, a transformation of the input results in an equivariant transformation of the output. 

As an example, consider a function $f$ that measures the average brightness of an image. Here, transforming an input image uniformly so that it has a lighter tone will result in a higher brightness output. With respect to this group of transformations, $f$ is equivariant. Next, we consider a group of transformations that lightens one half of the image and darkens the other by the same amount. Under such a group, the brightness output remains unchanged and $f$ would be invariant\index{Invariance} to such transformations.

\clearpage

\chapter{State of Research}\thispagestyle{empty}

The idea that a computer could understand an image, simulate what it has seen and finally visualize the result, similar to the process of vision-based mental simulation in humans, has inspired many researchers. One of the pioneers in this field has been Lawrence G. Roberts, who was able to extract basic geometric properties from simple shapes, apply transformations and re-render them in a new pose \citep{Roberts:1963}. In the following, we give a brief overview of recent research focusing on efforts that utilize deep learning \citep{Goodfellow:2016}.

\section{Inverse Graphics and Visual Question Answering}

For fully differentiable convolutional neural networks\index{Convolutional Neural Network} (CNN) \citep{LeCun:1989, Gu:2018} it isn't entirely clear what attributes of an object get encoded in its latent representation, how entangled they are and where exactly in the network they are located. There has been considerable progress in making these models more interpretable \citep{Simonyan:2014,Mahendran:2015,Ribeiro:2016,Lipton:2017,Montavon:2018,Zhang:2018} and it is generally believed that an object's representation is present in the later layers, which can be visualized by inverting the neural network \citep{Dosovitskiy:2016}. However, it has been shown that CNNs have a bias towards detecting textures instead of shapes \citep{Geirhos:2019} and are often fooled by the same object in a different pose \citep{Alcorn:2019}, indicating that the representation might be quite different to that of a human and possibly lacking some important attributes for our understanding.

During the quest for a more understandable representation for objects, there has been considerable effort to detect the poses of objects in a scene, such as finding their general orientation and bounding box \citep{Song:2016} or their rotational pose \citep{Yang:2015, Mahendran:2017, Wang2:2018}. All these approaches explicitly search for attributes. Using predictability minimization \citep{Schmidhuber:1992}, generative adversarial networks (GANs) \citep{Niemitalo:2010, Goodfellow:2014} or variational auto-encoders \citep{Kingma:2014}, these pose attributes often emerge naturally in the latent variables, but require some disentanglement \citep{Kulkarni:2015, Chen:2016, Higgins:2017, Nguyen-Phuoc:2019} in order to be usable.

While the pose is interesting, its meaning only becomes clear when the actual shape of an object is known. General image segmentation results are often not sufficient to describe a 3D shape. Classically \citep{Szeliski:2010}, a shape prior is chosen and then fit to the image \citep{Chan:2005, Calakli:2011, Dame:2013, Haene:2017}. Newer methods of shape fitting employ deep learning to aid in the fitting \citep{Izadinia:2017}. They require a set of 3D models that resemble the shape of real objects as well as possible, such as ShapeNet \citep{Chang:2015}. 

Apart from fitting individual shapes, also the outcome of an entire procedural modeling pipeline may be fitted. This process is far more complicated, as it not only involves fitting a single shape, but also requires an understanding of the relations between parts. However, a deeper understanding of object-part relations and the overall semantics does allow for more complex tasks, such as reconstructing hidden objects based on perceived symmetries \citep{Bokeloh:2010}. Also in inverse procedural modeling, deep learning is employed \citep{Felzenszwalb:2010, Tulsiani:2017, Zou:2017, Liu:2018, Romaszko:2018, Gafni:2019}. In this regard, a method related to our approach are stacked capsule autoencoders \citep{Kosiorek:2019}, which encode the object-part relation by assigning a different capsule to each.

A widely held belief is that one understands something better if one has the ability to generate it. In computer vision specifically, it is assumed that it should be easier to understand the interpretation of an image, if it can be generated using a grammar, a tree structure, a shape program, etc., which produces an image based on a series of commands. The goal is to infer these programs directly from the image using different machine learning methods \citep{Teboul:2010, Stava:2014, Wu:2017, Liu:2019, Tian:2019}. 

A structured scene representation is ideal to answer questions regarding the scene \citep{Yi:2018, Mao:2019}, such as about composition or appearance, referred to as visual question answering\index{Visual Question Answering} (VQA). On the other hand, it has also been shown, that such in-depth scene understanding is not necessarily required for VQA. It can be performed by a fully differentiable approach \citep{Hudson:2018}, using a mixture of convolutional and recurrent networks. 

Further, as an inverse to the idea of VQA, GeNeVa \citep{El-Nouby:2019}, a proposed GAN, is capable of drawing images based on text, implying that it has some hidden understanding of the scene. Such an understanding of the scene also allows the user to manipulate and re-render it \citep{Reed:2015, Yao:2018}.

\section{Intuitive Physics}

The previously discussed methods capable of describing a scene can be well suited for physical predictions, but also require a move towards video input instead of a static image. The most obvious way to perform physical simulations would be a hard-coded engine, such as PhysX \citep{Nvidia:2018}. These engines can calculate mechanical interactions accurately (billiard balls colliding) and even include advanced electro-magnetic effects (a concentrated beam of light igniting paper). Yet, this would require full knowledge of the physical properties, for which it is nearly impossible to infer them from an image alone, such as weight or temperature. However, a human is still able to perform crude predictions from images, without knowing much else in a process often referred to as intuitive physics. This insight motivates the use of deep learning in order to replicate this skill. 

We differentiate between two different approaches to intuitive physics: The first is to learn and predict physical effects directly based on image data \citep{Lerer:2016, Ehrhardt:2017, Ehrhardt:2018, Iten:2018}. This allows the model to be learned without any prior assumptions and has the added benefit that it can consider the entire image for clues about physical effects, including those that might have been missed by a more knowledge-based approach. However, such end-to-end systems have the problem of trying to solve two tasks, vision and physics, at once, each of which is difficult on its own. 

The second approach to intuitive physics is the one presented here, viewing vision and physics as two distinct domains, akin to how simulation \citep{Battaglia:2013} and game engines work \citep{Ullman:2017}. This allows for more interpretability and stability \citep{Chang:2017}, but requires prior extraction of all important information from the image into an adequate representation. One such approach are interaction networks\index{Interaction Networks} by \citet{Battaglia:2016}, which model physical effects on pairwise interactions of objects. We introduce this approach in more detail in Section 6.1. However, besides interaction networks there other models for physical scene understanding \citep{Chang:2017, Wu:2017, Wu2:2017, Steenkiste:2018, Kipf:2018}.

Interaction networks can be connected to the scene understanding methods discussed above or directly to a CNN \citep{Watters:2017} to learn from video data. There are also interesting extensions, such as automatic relation detection in scenes \citep{Raposo:2017}, integration of unsupervised methods \citep{Zheng:2018}, integrating graph networks \citep{Sanchez-Gonzalez:2018}, optimization of the flow of information to and from the network \citep{Hamrick:2017} and, finally, utilization in planning systems \citep{Pascanu:2017}, as we will do.

\section{Neural Network Architectures and Meta-Learning}

Designing neural networks is not a straightforward task and requires experience and trial and error when doing it manually. This was recognized early in the research and various genetic algorithms were developed \citep{Miller:1989, Kitano:1990} to automate the search for the design of a neural network. This process of neural architecture search\index{Neural Architecture Search} (NAS) \citep{Elsken:2019} is better understood now, with widely available libraries, such as Auto-Keras \citep{Jin:2018}, and also utilizes modern methods of reinforcement learning \citep{Baker:2017}. 

NAS generally produces a static neural network that is trained once and then used, but is not meant to learn anything new thereafter. For safety critical applications \citep{Amodei:2016} having a static is often a requirement, as the architecture was found with a specific dataset and purpose and it is possible that further training might lead to unexpected behavior \citep{McCloskey:1989, Ratcliff:1990}. In many cases transfer learning \citep{Tan:2018} is employed to take a pre-trained model and apply it to a new problem, but such an approach has a limited versatility \citep{Yosinski:2014}. To make a neural network more dynamic, lifelong meta-learning is used to let it learn and reconfigure itself even after it is already in use \citep{Chen:2018, Yoon:2018, Parisi:2019}.

Lifelong meta-learning\index{Meta-Learning} has the obvious need for more data, which is often limited. One possible remedy is the augmentation by synthetic data for which various 3D datasets have been proposed \citep{Chang:2015, Gaidon:2016, Mueller:2017, Johnson:2017}, each addressing a special use-case. Alternatively, full simulation environments \citep{Shah:2017, Johnson-Roberson:2017} based on computer games can be used. Models pre-trained on such synthetic datasets have been shown to generalize well and reduce the amount of training needed for real environments \citep{Sun:2014, Peng:2015, Tobin:2017, Tremblay:2018}, making it an ideal candidate for training set augmentation.

\clearpage

\chapter{Neural-Symbolic Capsules}\label{sec:NSC}\thispagestyle{empty}

\begin{wrapfigure}{R}{0.4\textwidth}
	\centering
	\begin{adjustbox}{max width=0.4\textwidth}
		\begin{tikzpicture}

		\vividnet{black!10!blue}{black!10!blue}{black!10!blue}{black!20!white}{black!20!white}{black!20!white}{black!20!white}

		\end{tikzpicture}
	\end{adjustbox}
\end{wrapfigure}

In neuroscience, it is argued that neurons in the cerebral cortex should not be seen as the fundamental computational unit of the brain and one should instead consider groups of about $100$ neurons organized as cortical minicolumns \citep{Buxhoeveden:2002}. This shifts the analysis of cognitive functions in the brain to an abstraction layer above that of neurons, where the minicolumns form the nodes of a network-of-networks.

For artificial neural networks the neuron is still seen as the fundamental computational unit and introducing further abstraction layers between the neuron and the final network has mainly been pursued in the field of neural-symbolic reasoning \citep{Garcez:2009, Besold:2017}. Here, we propose such an abstraction layer for vision by introducing special containers for compact machine learning algorithms, such as neural networks, which form the fundamental building blocks of our network-of-networks, loosely based on the idea of capsules proposed by \citet{Hinton:2011}.

In this chapter, we explore the inner workings of our neural-symbolic capsules and how they interact as part of a capsule network\footnotemark \footnotetext{The results in this and the following chapter were previously published at the German Conference on Pattern Recognition \citep{Kissner:2019}.}. We begin with a brief review of the classical capsule network, which inspired our neural-symbolic approach and its name, but differs strongly in its actual inner workings, as the original capsule does not act as a container for a neural network. We next introduce a generative grammar, initially assuming that it is hand-designed to procedurally generate images based on a set of attributes and show how the generative process for a single symbol in this grammar can be inverted to form a neural-symbolic capsule. While this is in reverse order of how the final capsule network works, where we invert a capsule network to find its generative grammar, it does help motivate both our naming scheme and the design process. Next, we explore how neural-symbolic capsules are connected in a network-of-networks to form a capsule network and discuss its properties. Finally, we show how our capsule network is related to the generative grammar we used as a design guide and acts as its dual representation.

\section{Classical Capsules}

The classical capsule network is an extension of artificial neurons outputting vectors instead of a single scalar. This requires a different internal mechanism, known as the routing-by-agreement protocol \citep{Sabour:2017}, to produce the desired output. The overall idea is to multiply the inputs by a weight matrix instead of a weight vector and then use the result to predict what the individual inputs should have been, in order to iteratively improve the output vector.

The output vector can be interpreted as an attribute vector and each capsule can be seen to represent some visual object. Further, as the capsules are equivariant, the information loss between the layers is reduced in comparison to, for example, pooling in CNNs \citep{Krizhevsky:2012, LeCun:1998}.

Recently, additional extensions for capsules have been proposed, such as using matrices internally \citep{Hinton:2018}, employing 3D input \citep{Zhao:2018}, improving equivariance \citep{Lenssen:2018} and even a proposal for hardware acceleration has been made \citep{Marchisio:2019}.

The neural-symbolic capsules presented here act more as a container \citep{Hinton:2014} and diverge somewhat from their original construction, but the overall idea of routing-by-agreement, interpreting the output vectors as attributes and the capsules as objects remains the same. 

\section{Attributed Generative Grammar}

Our generative grammar is inspired by \citep{Leyton:2001, Martinovic:2013, Martinovic:2015}. It procedurally generates an image based on interpretable input information. We require that our grammar is context-free, non-recursive and has only a finite number of symbols in order to avoid infinite productions. In the following, all the individual components of the grammar are explained in detail.

\begin{equation}\label{eq:GrammarFull}
G = (S, V, \Sigma, R, A, D)
\end{equation}

\begin{tabular}{ll}
	$S$: & the axiom\\
	$V$: & non-terminal symbols\\
	$\Sigma$: & terminal symbols\\
	$R$: & set of production rules $r$\\
	$A$: & set of attributes $\alpha$\\
	$D$: & set of decoders $g$ for rule $r$\\
\end{tabular}

\bigskip

\noindent\textbf{The Axiom:} Each generative grammar produces a specific type of image described using a symbol, \ie, the axiom. This symbol is a compound noun which sufficiently represents the image's content the grammar produces. For example, to generate an image of a street scene, we construct a grammar with \symb{street-scene} as its axiom. 

\bigskip

\noindent\textbf{Non-Terminal Symbols:} Continuing with the \symb{street-scene}, in order to draw this symbol, we use a set of objects that represent the parts of the axiom symbol. These parts also describe nouns and we refer to them as non-terminal symbols. For example, a street may contain a \symb{house} symbol. Generally, the axiom consists of multiple parts, which we describe by concatenating the individual symbols such as \symb{house}\symb{house}\symb{post-office}. We also consider the axiom itself as a non-terminal symbol.

Each part may also be made up of parts. We refer to the process of splitting an object into its parts as a production. As all these objects represent real-world entities, this splitting process must be finite.

\bigskip

\noindent\textbf{Terminal Symbols:} We refer to the most primitive parts reached during production as the terminal symbols. For our purpose of image generation, the terminal symbols represent renderable primitives, such as a \symb{square} or an \symb{edge}. As will become apparent later, our goal is a low number of terminal symbols. \Ie, we aim to find a small set of renderable primitives from which all non-terminal symbols can be constructed. 

\begin{figure}[H]
	\centering
	\begin{adjustbox}{max width=1.0\textwidth}
		\begin{tikzpicture}

		%- Grammar Parse-Tree
		\node [blacknode] at (0, 0)  {}  
		child[grow=right]
		{ 
			node (A1) [whitenode] at (0, 0.5) {} edge from parent[tochild]
			{}
		}
		child[grow=right]
		{ 
			node (A2) [whitenode] at (0, -0.5) {} edge from parent[tochild]
			{}
		};
	
		\draw [white] (0,0.25) -- (0.2,0.05) -- (-0.2,0.05) -- (0,0.25) ;
		\draw [white] (0.2, 0.05) -- (0.2,-0.25) -- (-0.2, -0.25) -- (-0.2, 0.05) -- (0.2, 0.05);
		
		\draw [black] (1.5,0.6) -- (1.7,0.4) -- (1.3,0.4) -- (1.5,0.6) ;
		\draw [black] (1.7, -0.35) -- (1.7,-0.65) -- (1.3, -0.65) -- (1.3, -0.35) -- (1.7, -0.35);
		
		\end{tikzpicture}
	\end{adjustbox}
	\caption{Producing terminal symbols from a \symb{house} symbol. The black hexagon represents a non-terminal symbol and the white hexagons the terminal symbols.} \label{fig:FullProduction}
\end{figure}
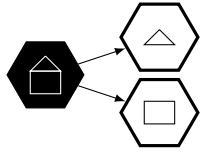

\bigskip
% TODO: Important! There is a \linebreak below!
\noindent\textbf{Production Rules:} For each non-terminal symbol in our grammar, a set of parts is produced as a mixture of non-terminal or terminal symbols. The set of produced symbols is determined by a rule $r$. We write
\begin{equation}\label{eq:Rule}
r : \Omega \to \lambda_1\cdots\lambda_n  \;\;\;\; \text{where} \; \Omega\in V, \lambda\in \bigcup_{i\in\mathbb{N}\setminus\{0\}}(V\cup\Sigma)
\end{equation}
to indicate that a non-terminal symbol $\Omega$ produces the symbols $\lambda_1$ to $\lambda_n$. Our \symb{street-scene} grammar would, thus, have a rule $r_1$ that produces \symb{house}\symb{house}\symb{post-office} from \linebreak \symb{street-scene}. We may continue this process, splitting the \symb{house} into its floors, each floor into rooms and furniture. Finally we would split the furniture using a rule such as $r_i : $\symb{table}$\to$\symb{square}\symb{triangle} for the 2D case. Here, we chose \symb{square} and \symb{triangle} as our graphical primitives, \ie, terminal symbols. By defining a similar rule for the \symb{post-office} symbol, we are able to produce all terminal symbols for the entire \symb{street-scene} axiom. We refer to the final tree-structure as the parse-tree and an example parse-tree of a \symb{house} is shown in Figure \ref{fig:FullProduction}.

\bigskip

\noindent\textbf{Attributes:} As rendering a \symb{street-scene} using the same houses over and over again would be rather pointless, we attach attributes to each symbol to allow for variety. By $\vec{\alpha}_j = (\alpha_j^1, \cdots, \alpha_j^{n} )$ we denote the vector of attributes for a symbol $\lambda_j$ and restrict the range of the attribute vector to $[0,1]^n$. These attributes describe the visual appearance of a symbol and are equivalent to attaching \textit{adjectives}, \textit{preposition} and \textit{verbs} to the corresponding noun. As we did with symbols, we introduce the notation \attr{$\cdots$} for the semantic representation of a specific attribute $\alpha_j^{\cdots}$.
 
The lexical class of each attribute becomes clear by looking at the specific example of an \symb{office-chair}. Assume this chair is made out of some metal and is able to corrode. We measure how corroded the chair is on a scale of $0$ to $1$ and encode this in the \textit{adjective} attribute \attr{corroded}. The \symb{office-chair} also has some \attr{position} attribute, which allows us to set it in relation to other objects in a scene using \textit{prepositions}, such as "on" or "near". Finally, the chair is able to swivel around its base over time, which is encoded in the \attr{swivel} \textit{verb} attribute, that defines a relative rotation between the seat and base. Note that \attr{swivel} is different from \attr{rotation}, as the former describes an internal rotation between two parts, whereas the latter describes the rotation of the chair as a whole. 

\noindent\textbf{Decoders:} The attributes of a rule $r:\Omega\to\lambda_j$ are produced by means of decoders $g$. These decoders take the attributes $\vec{\alpha}_\Omega$ of the left-hand side and generate the attributes $\vec{\alpha}_j$ of the right-hand side, using
\begin{equation}
\alpha^i_j = g^i_j(\vec{\alpha}_\Omega) \;\;\; .
\end{equation}

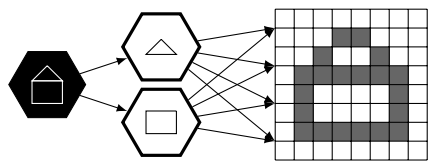
\begin{figure}[H]
	\centering
	\begin{adjustbox}{max width=1.0\textwidth}
		\begin{tikzpicture}

		%- Grammar Parse-Tree
		\node [blacknode] at (0, 0)  {}  
		child[grow=right]
		{ 
			node (A1) [whitenode] at (0, 0.5) {} edge from parent[tochild]
			{}
		}
		child[grow=right]
		{ 
			node (A2) [whitenode] at (0, -0.5) {} edge from parent[tochild]
			{}
		};
		
		\fill[black!60] (3.25,-0.75) rectangle (3.5,-0.5);
		\fill[black!60] (3.25,-0.5) rectangle (3.5,-0.25);
		\fill[black!60] (3.25,-0.25) rectangle (3.5,-0.0);
		\fill[black!60] (3.25,-0.0) rectangle (3.5,0.25);
		
		\fill[black!60] (4.5,-0.75) rectangle (4.75,-0.5);
		\fill[black!60] (4.5,-0.5) rectangle (4.75,-0.25);
		\fill[black!60] (4.5,-0.25) rectangle (4.75,-0.0);
		\fill[black!60] (4.5,-0.0) rectangle (4.75,0.25);
		
		\fill[black!60] (3.5,-0.0) rectangle (3.75,0.25);
		\fill[black!60] (3.75,-0.0) rectangle (4,0.25);
		\fill[black!60] (4.0,-0.0) rectangle (4.25,0.25);
		\fill[black!60] (4.25,-0.0) rectangle (4.5,0.25);
		
		\fill[black!60] (3.5,-0.75) rectangle (3.75,-0.5);
		\fill[black!60] (3.75,-0.75) rectangle (4,-0.5);
		\fill[black!60] (4.0,-0.75) rectangle (4.25,-0.5);
		\fill[black!60] (4.25,-0.75) rectangle (4.5,-0.5);
		
		\fill[black!60] (4.25,0.25) rectangle (4.5,0.5);
		\fill[black!60] (3.5,0.25) rectangle (3.75,0.5);
		\fill[black!60] (3.75,0.5) rectangle (4,0.75);
		\fill[black!60] (4.0,0.5) rectangle (4.25,0.75);
		
		\draw[scale=0.25] (12, -4) grid (20, 4);

		\draw[tochild, black] (A1) -- (3, 0.75);
		\draw[tochild, black] (A1) -- (3, 0.25);
		\draw[tochild, black] (A1) -- (3, -0.25);
		\draw[tochild, black] (A1) -- (3, -0.75);
		\draw[tochild, black] (A2) -- (3, 0.75);
		\draw[tochild, black] (A2) -- (3, 0.25);
		\draw[tochild, black] (A2) -- (3, -0.25);
		\draw[tochild, black] (A2) -- (3, -0.75);
		
		\draw [white] (0,0.25) -- (0.2,0.05) -- (-0.2,0.05) -- (0,0.25) ;
		\draw [white] (0.2, 0.05) -- (0.2,-0.25) -- (-0.2, -0.25) -- (-0.2, 0.05) -- (0.2, 0.05);
		
		\draw [black] (1.5,0.6) -- (1.7,0.4) -- (1.3,0.4) -- (1.5,0.6) ;
		\draw [black] (1.7, -0.35) -- (1.7,-0.65) -- (1.3, -0.65) -- (1.3, -0.35) -- (1.7, -0.35);
		
		\end{tikzpicture}
	\end{adjustbox}
	\caption{Generating an image from a \symb{house} symbol.} \label{fig:Drawing}
\end{figure}

We also introduce a special type of drawing-decoder, which takes terminal symbols and their attributes and draws them on the screen, essentially rendering the grammar parse-tree. For simplicity, we refer to this drawing function as a decoder $g$ as well, but note that it is not attached to any rule. Here the resulting attributes $\vec{\alpha}^i_j$ represent the pixel color values, as shown in Figure \ref{fig:Drawing} for a \symb{house} symbol.

\subsection{Style, Pose, Configuration and Viewpoint}

A distinct choice of rules and attributes will lead to different effects on the final image. Specifically, we can summarize these as differences in style, pose, configuration or viewpoint.

\bigskip

\noindent\textbf{Style:} The most straightforward design choice are the attributes, specifically those interpretable as adjectives, which determine the possible styles the objects in the final image can have. Only a \symb{house} with a \attr{red} attribute will allow for variety in its redness. Also more complex style-attributes are possibly, such as a grading of how \attr{rustic} a \symb{house} is.

\bigskip

\noindent\textbf{Pose:} Attributes interpretable as verbs have a different effect on the final image. Instead of styles, they describe an object's different poses. Varying a particular attribute in the range $[0,1]$ is equivalent to an animation, such as walking.

\bigskip

\noindent\textbf{Configuration:} A rule produces a fixed set of parts for an object. However, an object should still be classified correctly, even if it has a different configuration of parts. A \symb{house} is still a \symb{house}, even if it produces two \symb{floor} symbols instead of one. We, thus, may have multiple rules with the same left-hand side in Equation \ref{eq:Rule} and must choose an appropriate one when producing an image.

\bigskip

\noindent\textbf{Viewpoint:} Apart from different configurations, rules with the same left-hand side may also describe different viewpoints. Particularly in 3D, an object viewed from a different angle will have different visible parts, \ie, seem to have a different configuration of parts. These rules are, thus, equivalent to the nodes of an aspect graph.

\subsection{Rendering and Occlusion}

Producing an image using the grammar based on a symbol and corresponding attributes is equivalent to rendering an image in computer graphics. Yet, it is not necessary to begin this process at the axiom. We are free to choose any symbol and begin from there, choosing appropriate rules along the way. 

However, there can be some ambiguity in ordering the final renderable primitives according to their depth, when doing it independent of the starting point. To remedy this problem, we define the symbol order in a production rule in accordance with the depth or drawing order, also referred to as \index{Painter's Algorithm}painter's algorithm. Thus, \symb{floor}\symb{chair}\symb{table} would entail first rendering the \symb{floor}, then the \symb{chair} and finally the \symb{table}.

\subsection{Completeness}

For our discussion it is important that our grammar is complete. By complete we mean that it is able to produce any possible image of a given size $n\times n$ given an adequate set $G$ (\cf Equation \ref{eq:GrammarFull}). We will prove this explicitly.

Consider non-terminal symbols $\Omega_{m,m}$ each with a production rule of the form $r_{\Omega_{m,m}} \colon \Omega\to\lambda_{0,0}\cdots\lambda_{3,3}$, where $\lambda_{i,j}$ are terminal symbols, essentially splitting the $n\times n$ target image into $m \times m$ smaller $3 \times 3$ patches. For any image we can craft a constant decoder $g: \cdot \to \mathbb{R}^{3\times 3}$ that simply outputs the appropriate $3 \times 3$ patch and all symbols together would output the original image. Essentially $g$ defines a single point in $\mathbb{R}^{3\times 3}$. This can be expanded by letting each decoder $g$ take in one attribute, \ie, defining a line $g : \mathbb{R} \to \mathbb{R}^{3 \times 3}$ in $3 \times 3$-dimensional space instead of a point. As long as this line goes through the point that was used in the constant case, the symbols are able to detect and reproduce the original image, but also others described by points on the line.

We may continue this construction by adding further attributes and essentially reduce our proof to the following: If our terminal symbol $\lambda_{i,j}$ can produce all possible $3 \times 3$ images, then our grammar is complete. This is trivially true, as we can always choose our terminal symbols to do a lossless discrete cosine transform with a maximum of $3 \times 3$ discrete attributes, to which we will refer as \symb{DCT}. In this extreme case, our grammar is essentially a lossless JPEG-like decoding algorithm.

While this proves completeness, we are, of course, mainly interested in semantically rich cases. We split the domain of the \symb{DCT} symbols to not include all those transforms that look like an edge resulting in two terminal symbols instead, \symb{DCT-without-edges} and \symb{edge}. Our previous analysis still holds and we may repeat this splitting process as often as we deem sensible without hurting completeness. At some point, the \symb{DCT-without-edges-patches-\dots} symbols will only have a very small domain it is actually used for, only drawing noise-like sections of the image, and all the interesting features are drawn by other terminal symbols with a richer interpretation.

\section{From Symbols to Capsules}

In the following we present our neural-symbolic capsules, each of which acts as a container for a regression model and a routing-by-agreement protocol managing the flow of information from the capsule's input to its output. These capsules are interpreted as distinct visual objects, which not only output the object's attributes, but also an activation probability expressing how confident the capsule is in its detection and use these terms interchangeably.

To motivate our design, we invert the previously defined generative grammar and its produced parse-tree to form a capsule network inspired by \citep{Towell:1993,Towell:1994}. Each symbol, rule, attribute and decoder of the grammar has a dual element in the capsule network:

\bigskip

\noindent\textbf{Terminal Symbols $\to$ Primitive Capsules:} A terminal symbol is a renderable graphical primitive. In the parse-tree, these symbols are connected directly to the drawing function, \eg, shown in Figure \ref{fig:Drawing}. We refer to its inverse as a \textbf{primitive capsule}. These capsules take pixel data as input to determine if the image consists of the graphical primitive it represents and what attributes it has. 

\bigskip

\noindent\textbf{Non-Terminal Symbols $\to$ Semantic Capsules:} As for the terminal symbols, non-terminal symbols are inverted to form \textbf{semantic capsules}. These capsules are only connected to other capsules and have no link to direct pixel data. Instead, they take other capsule's attributes and confidence values as input to produce their own.

By $\Omega$ and $\lambda$ we will be referring to both the symbol and its inverse, the capsule, depending on context. The same applies to their corresponding attributes $\vec{\alpha}$, which remain unchanged in both representations.

\bigskip

\noindent\textbf{Rules $\to$ Routes:} For a rule $r:\Omega \to \lambda_i$ a decoder $g$ transforms $\vec{\alpha}_\Omega$ into $\vec{\alpha}_1,\cdots\,\vec{\alpha}_{\vert\lambda\vert}$. We refer to the inversion of the rule $r$ as a \textbf{route} and the inversion of the decoder $g$ as the \textbf{encoder}. While $g$ generally is not exactly invertible, we introduce $\gamma$ as the actual encoder and let it be our best approximation, by minimizing
\begin{equation}\label{eq:minimization}
||g(\gamma(\vec{\alpha})) - \vec{\alpha}|| \;\;\; .
\end{equation}

\begin{figure}[H]
	\centering	
	\begin{adjustbox}{max width=1.0\textwidth}
		\begin{tikzpicture}
		
		%- Grammar Parse-Tree
		\node [blacknode] at (-1, 0)  {$\Omega$}  
		child[grow=right]{ node [whitenode] at (3, 0.5) {$\lambda_1$} edge from parent[tochild, sloped] node[above] {$\alpha_j^i=g^i_j(\vec{\alpha}_\Omega$)} }
		child[grow=right]{ node [whitenode] at (3, -0.5) {$\lambda_2$} edge from parent[tochild] };
		
		\draw (1.5, -1.5) node {\begin{tabular}{c} Symbol $\Omega$ \\ with production rule $r$ \\ and decoder $g$ \end{tabular}};
		
		%- Arrow
		
		\draw [->, >=stealth, thick] (4.5,0) -- (6,0);
		
		%- Capsule Parse-Tree
		\node [blacknode] at (11.5, 0)  {$\Omega$}  
		child[grow=left]{ node [whitenode] at (-3, 0.5) {$\lambda_1$} edge from parent[toparent, sloped] node [above]{$\alpha_\Omega^i\approx\gamma^i(\vec{\alpha}_{1,\cdots,\vert\lambda\vert})$} }
		child[grow=left]{ node [whitenode] at (-3, -0.5) {$\lambda_2$} edge from parent[toparent] };
		
		\draw (9.5, -1.5) node {\begin{tabular}{c} Capsule $\Omega$ \\ with route $r$ \\ and encoder $\gamma$ \end{tabular}};
		
		\end{tikzpicture}
	\end{adjustbox}
	\caption{Inversion of a symbol of the grammar parse-tree results in a capsule. We illustrate both the symbol and the capsule using a hexagon to avoid confusion with neurons.} \label{fig:reverseNode}
\end{figure}
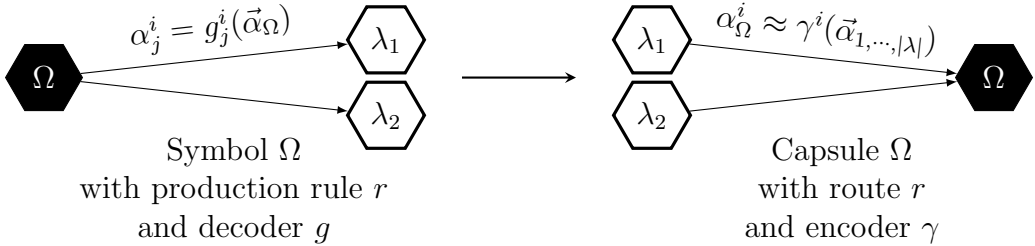

The overall inversion process is summarized in Figure \ref{fig:reverseNode}. 

By using an individual capsule for each object, we avoid early on some of the entanglement that general object recognition methods face in deep learning \citep{DiCarlo:2007, Achille:2018}. 

\subsection{Routing-by-Agreement}

For the inversion it does not suffice to simply invert the individual elements to obtain a fully functional capsule network. While such a network can accurately process an image that represents the axiom of the underlying generative grammar, even with some generalization, its output will produce nonsense for anything else. We require an additional mechanism to measure the confidence each individual capsule has in detecting the correct object and an algorithm to calculate it. We use a modified routing-by-agreement protocol introduced in \citep{Sabour:2017} to find this activation probability $p$, which is equivalent to the confidence the capsule has.

\begin{figure}[H]
	\centering
	\begin{adjustbox}{max width=1\textwidth}
		\begin{tikzpicture}
		
		%- Bounding Boxes
		\draw[draw=black!50, dashed, fill=black!5] (-30mm,40mm) rectangle (185mm, -72.5mm);
		\draw[draw=black!10!yellow, dashed, fill=yellow!10] (-22.5mm,30mm) rectangle (22.5mm, -62.5mm);
		\draw[draw=black!10!yellow, dashed, fill=yellow!10] (37.5mm,30mm) rectangle (82.5mm, -62.5mm);
		
		%- Inputs		
		\node (InA) [whitenodefill] at (-37.5mm, 35mm) {$\vec{\alpha}_i$};
		\node (InP) [whitenodefill] at (-37.5mm, -67.5mm) {$p_i$};
		
		%- First Autoencoder
		\node (A01) [trapezium, trapezium angle=-60, minimum height=15mm, draw=black!30!green, fill=black!10!green!20, thick] at (0, 15mm) {$\gamma_{r_1}$};
		\node (A02) [draw=red, fill=red!20, rounded rectangle, thick] at (0, 0) {$\vec{\alpha}_{\Omega_1}$};
		\node (A03) [trapezium, trapezium angle=60, minimum height=15mm, draw=black!30!green, fill=black!10!green!20, thick] at (0,-15mm) {$g_{r_1}$};
		\draw [tochild, black] (A01) -- (A02);
		\draw [tochild, black] (A02) -- (A03);
		\node (A0i) [rectangle, draw=black!30!green, text height=2.5mm, fill=black!10!green!20, thick] at (0, 22.5mm) {\footnotesize $\vec{\alpha}_i$};
		\node (A0o) [rectangle, draw=black!30!green, text height=2.5mm, fill=black!10!green!20, thick] at (0, -22.5mm) {\footnotesize $\vec{\tilde{\alpha}}_i$};
		
		%- First agreement
		\node (A0m) [rectangle, draw=black!30!blue, fill=black!10!blue!20, minimum height = 20mm, minimum width = 30mm, thick] at (0, -45mm) {Agreement};
		\node (A0mi1) [rectangle, draw=black!30!blue, text height=2.5mm, fill=black!10!blue!20, thick] at (0, -35mm) {\footnotesize $\vec{\tilde{\alpha}}_i$};
		\node (A0mi2) [rectangle, draw=black!30!blue, text height=2.5mm, fill=black!10!blue!20, thick] at (-15mm, -40mm) {\footnotesize $\vec{\alpha}_i$};
		\node (A0mi3) [rectangle, draw=black!30!blue, text height=2.5mm, fill=black!10!blue!20, thick] at (-15mm, -50mm) {\footnotesize $p_i$};
		\node (A0mo1) [rectangle, draw=black!30!blue, text height=2.5mm, fill=black!10!blue!20, thick] at (15mm, -45mm) {\footnotesize $p_{\Omega_1}$};
		\draw [tochild, black] (A0o) -- (A0mi1);
		
		%- ith Autoencoder
		\node (Ai1) at (27.5mm, 15mm) {$\cdots$};
		\node (Ai3) at (27.5mm, -15mm) {$\cdots$};
		
		%- nth Autoencoder
		\node (An1) [trapezium, trapezium angle=-60, minimum height=15mm, draw=black!30!green, fill=black!10!green!20, thick] at (60mm, 15mm) {$\gamma_{r_n}$};
		\node (An2) [draw=red, fill=red!20, rounded rectangle, thick] at (60mm, 0) {$\vec{\alpha}_{\Omega_n}$};
		\node (An3) [trapezium, trapezium angle=60, minimum height=15mm, draw=black!30!green, fill=black!10!green!20, thick] at (60mm,-15mm) {$g_{r_n}$};
		\draw [tochild, black] (An1) -- (An2);
		\draw [tochild, black] (An2) -- (An3);
		\node (Ani) [rectangle, draw=black!30!green, text height=2.5mm, fill=black!10!green!20, thick] at (60mm, 22.5mm) {\footnotesize $\vec{\alpha}_i$};
		\node (Ano) [rectangle, draw=black!30!green, text height=2.5mm, fill=black!10!green!20, thick] at (60mm, -22.5mm) {\footnotesize $\vec{\tilde{\alpha}}_i$};
		
		%- nth agreement
		\node (Anm) [rectangle, draw=black!30!blue, fill=black!10!blue!20, minimum height = 20mm, minimum width = 30mm, thick] at (60mm, -45mm) {Agreement};
		\node (Anmi1) [rectangle, draw=black!30!blue, text height=2.5mm, fill=black!10!blue!20, thick] at (60mm, -35mm) {\footnotesize $\vec{\tilde{\alpha}}_i$};
		\node (Anmi2) [rectangle, draw=black!30!blue, text height=2.5mm, fill=black!10!blue!20, thick] at (45mm, -40mm) {\footnotesize $\vec{\alpha}_i$};
		\node (Anmi3) [rectangle, draw=black!30!blue, text height=2.5mm, fill=black!10!blue!20, thick] at (45mm, -50mm) {\footnotesize $p_i$};
		\node (Anmo1) [rectangle, draw=black!30!blue, text height=2.5mm, fill=black!10!blue!20, thick] at (75mm, -45mm) {\footnotesize $p_{\Omega_n}$};
		\draw [tochild, black] (Ano) -- (Anmi1);
		
		%- Connecting Inputs to Autoencoders
		\draw[tochild, black] (InA) -- (0, 35mm) -- (A0i);
		\draw[tochild, black] (InA) -- (60mm, 35mm) -- (Ani);
		\draw[tochild, black] (InA) -- (-25mm, 35mm) -- (-25mm, -40mm) -- (A0mi2);
		\draw[tochild, black] (InP) -- (-25mm, -67.5mm) -- (-25mm, -50mm) -- (A0mi3);
		\draw[tochild, black] (InP) -- (35mm, -67.5mm) -- (35mm, -50mm) -- (Anmi3);
		\draw[tochild, black] (InA) -- (35mm, 35mm) -- (35mm, -40mm) -- (Anmi2);
		
		%- Outputs
		\node (OutABox) [rectangle, draw=black, text height=2.5mm, text width=30mm, thick, fill=white, minimum height = 20mm, minimum width = 30mm, align=center] at (120mm, -5mm) {Route\\Selection};
		\node (OutAi1) [rectangle, draw=black, fill=white, text height=2.5mm, thick] at (105mm, 0) {\footnotesize $\vec{\alpha}_{\Omega_j}$};
		\node (OutAi2) [rectangle, draw=black, fill=white, text height=2.5mm, thick] at (105mm, -10mm) {\footnotesize $p_{\Omega_j}$};
		\node (OutAo1) [rectangle, draw=black, fill=white, text height=2.5mm, thick] at (135mm, 0mm) {\footnotesize $\vec{\alpha}_\Omega$};
		\node (OutPo1) [rectangle, draw=black, fill=white, text height=2.5mm, thick] at (135mm, -10mm) {\footnotesize $p_\Omega$};
		
		\node (OutA) [whitenodefill] at (195mm, -0mm) {$\vec{\alpha}_\Omega$};
		\node (OutP) [whitenodefill] at (195mm, -10mm) {$p_\Omega$};
		
		%- Connecting Outputs
		\draw[tochild, blue, line width=1mm] (Anmo1) -- (90mm,-45mm) -- (90mm,-10mm) -- (OutAi2);
		\draw[blue, line width=1mm] (A0mo1) -- (Anm);
		\draw[tochild, red, line width=1mm] (An2) -- (OutAi1);
		\draw[red, line width=1mm] (A02) -- (An2);
		\draw[tochild, black] (150mm, -0mm) -- (OutA);
		\draw[tochild, black] (150mm, -10mm) -- (OutP);
		\draw[tochild, black] (OutAo1) -- (150mm, -0mm);
		\draw[tochild, black] (OutPo1) -- (150mm, -10mm);
		
		%- Labels
		\node[text=black!50] at (177mm, 35.5mm) {Capsule};
		\node[text=black!50!yellow] at (14.5mm, -60.5mm) {Route 1};
		\node[text=black!50!yellow] at (74.5mm, -60.5mm) {Route n};
		
		%- Observation List
		
		\draw[draw=black, fill=white, thick, dashed] (145mm, -15mm) rectangle (180mm, -62.5mm);
		\draw[draw=black, fill=white, thick] (145mm, 5mm) rectangle (180mm, -15mm);
		\node[text=black, text width=30mm, align=center] at (162.5mm, -5mm) {Observation\\Table};
		\node[text=black] at (162.5mm, -19mm) {$(p_\Omega, \vec{\alpha}_\lambda, \vec{\alpha}_\Omega)^{(1)}$};
		\node[text=black] at (162.5mm, -27mm) {$(p_\Omega, \vec{\alpha}_\lambda, \vec{\alpha}_\Omega)^{(2)}$};
		\node[text=black] at (162.5mm, -35mm) {$(p_\Omega, \vec{\alpha}_\lambda, \vec{\alpha}_\Omega)^{(3)}$};
		\node[text=black] at (162.5mm, -43mm) {\textbf{$\vdots$}};
		\draw[draw=black, dashed] (145mm, -23mm) -- (180mm, -23mm);
		\draw[draw=black, dashed] (145mm, -31mm) -- (180mm, -31mm);
		\draw[draw=black, dashed] (145mm, -39mm) -- (180mm, -39mm);
		
		\node (InAL) [rectangle, draw=black, fill=white, text height=2.5mm, thick] at (145mm, 0mm) {\footnotesize $\vec{\alpha}_\Omega$};
		\node (InPL) [rectangle, draw=black, fill=white, text height=2.5mm, thick] at (145mm, -10mm) {\footnotesize $p_\Omega$};
		\node (OutAL) [rectangle, draw=black, fill=white, text height=2.5mm, thick] at (180mm, 0mm) {\footnotesize $\vec{\alpha}_\Omega$};
		\node (OutPL) [rectangle, draw=black, fill=white, text height=2.5mm, thick] at (180mm, -10mm) {\footnotesize $p_\Omega$};
		
		\end{tikzpicture}
	\end{adjustbox}
	\caption{The inner structure of a capsule $\Omega$ with inputs $\vec{\alpha}_i, p_i$ and outputs $\vec{\alpha}_\Omega, p_\Omega$, representing our routing-by-agreement protocol (Equations \ref{eq:CapsuleInternals0} to \ref{eq:CapsuleInternals4}). The outputs are stored in an observation table and individual routes are highlighted in yellow. } \label{fig:CapsuleBlockDiagram}
\end{figure}
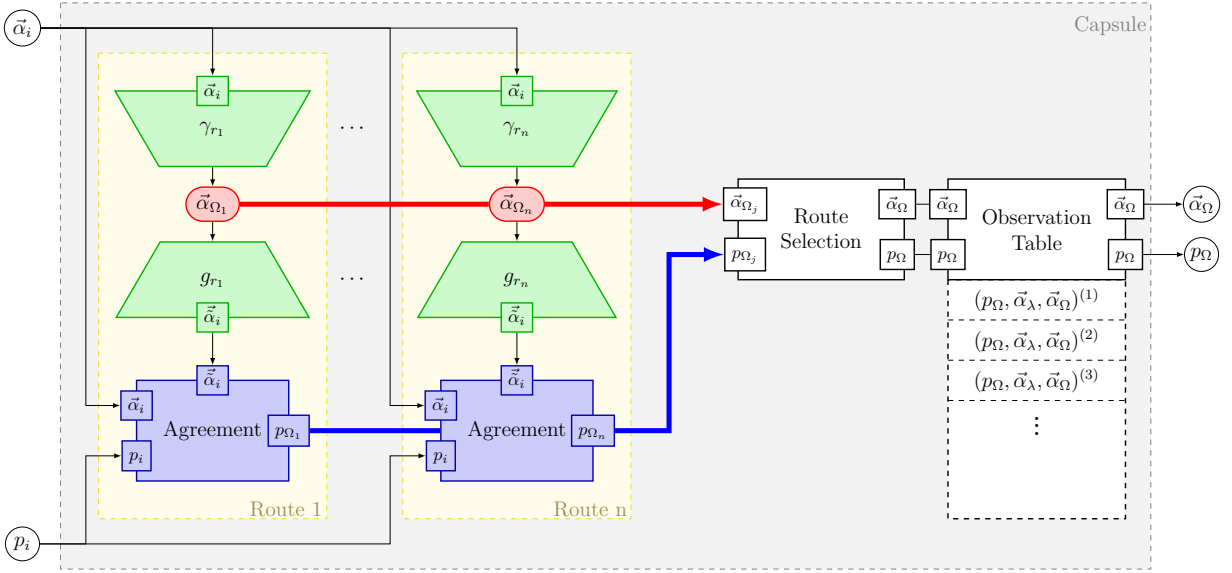

\begin{figure}[H]
	\centering
	\begin{adjustbox}{max width=0.5\textwidth}
		\begin{tikzpicture}

		%- Capsule Parse-Tree
		\node [blacknode] at (0, 0)  {$\Omega$}  
		child[grow=left]{ node [whitenode] at (-2, 1.75) {$\lambda_1$} edge from parent[toparent] }
		child[grow=left]{ node [whitenode] at (-2, 0.75) {$\lambda_2$} edge from parent[toparent] }
		child[grow=left]{ node [rednode] at (-2, -0.75) {$\lambda_1$} edge from parent[toparent, red] }
		child[grow=left]{ node [rednode] at (-2, -1.75) {$\lambda_2$} edge from parent[toparent, red] }
		child[grow=left]{ node [rednode] at (-2, -2.75) {$\lambda_3$} edge from parent[toparent, red] };
		
		%- Divider
		
		\draw [dotted] (-9,0) -- (-1,0);
		
		%- Labels
		
		\draw (-7, 1.25) node {$r_1$ with probability $p_{\Omega_1}$};
		\draw (-7, -1.75) node {$r_2$ with probability $p_{\Omega_2}$};
		
		\end{tikzpicture}
	\end{adjustbox}
	\caption{Actual routes taken by routing-by-agreement. Here, $p_{\Omega_2} > p_{\Omega_1}$, thus, the output of $r_2$  is used for the capsule's attributes $\vec{\alpha}_\Omega$.} \label{fig:twoSetsToOneNode}
\end{figure}
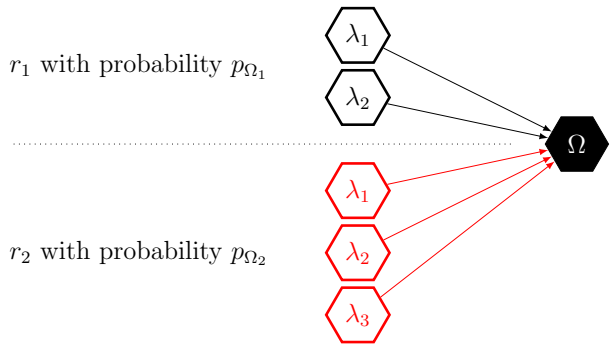

In addition, this routing-by-agreement protocol is able to determine which route is the most sensible, \ie, what production rule could have led to the configuration or viewpoint represented in the image. The overall idea behind the algorithm is simple: We take a symbol $\Omega$ and use the decoder $\gamma$ to determine its attributes based on the input capsules or pixels. Using these attributes, we generate what the capsule expected the input should have been with encoder $g$ and compare these results with the actual input. If both input and expected input agree, the capsule has a high activation probability $p$, otherwise $p$ will be small. This is repeated for all routes of a capsule (\ie, rules with the same left-hand side) to find the route with highest activation, which is taken to be the correct one. In detail, our protocol works as follows (\cf Figure \ref{fig:CapsuleBlockDiagram}):

\begin{enumerate}
	
	\item The output $\vec{\alpha}_{\Omega_r}$ for a route $r$ is calculated using
	
	\begin{equation}\label{eq:CapsuleInternals0}
	\vec{\alpha}_{\Omega_r}=\gamma_r(\vec{\alpha}_{1,\cdots,\vert\lambda\vert})\;\;\; .
	\end{equation}
	
	\item For each input $\vec{\alpha}_{j_r}$ for a route $r$, we estimate the expected input value $\vec{\tilde{\alpha}}_{j_r}$ as if $\vec{\alpha}_{j_r}$ were unknown, using the following equation:
	\begin{equation}\label{eq:CapsuleInternals1}
	\vec{\tilde{\alpha}}_{j} = g_{r,j}(\vec{\alpha}_{\Omega_r})\;\;\; .
	\end{equation}
	
	\item The activation probability $p_{\Omega_r}$ of a route $r$ is calculated as 
	\begin{equation}\label{eq:CapsuleInternals2}
	p_{\Omega_r} = \frac{1}{\vert(\lambda)_r\vert}\;\;\sum_{(\lambda)_r}\; \frac{\lVert Z\left(\vec{\alpha}_i, \vec{\tilde{\alpha}}_i\right)\rVert_1}{\vert Z \vert} \cdot w\left(\frac{p_{i}}{\bar{p}_i} - 1 \right)\;\;\; ,
	\end{equation}
	where $(\lambda)_r$ denotes the set of all inputs that contribute to a route $r$, $p_{i}$ the route's input capsule's activation probability, $Z$ an agreement-function comparing the actual input to the expected input that outputs a vector of size $\vert Z \vert$, $\lVert \cdot \rVert_1$ the $l_1$-norm, $w$ some window function with $w(0) = 1$, $\sup\{w\} = 1$ and $\bar{p}_i$ the mean of all past probabilities the route has calculated using Equation \ref{eq:CapsuleInternals2} for that input. 
	
	\item Steps 1. - 3. are repeated for each $r\in R(\Omega)$.
	
	\item Find the route that was most likely used (\cf Figure \ref{fig:twoSetsToOneNode})
	\begin{equation}\label{eq:CapsuleInternals3}
	r_{\textit{final}} = \max_r \{ p_{\Omega_r} \}
	\end{equation}
	and set the final output to
	
	\begin{align}\label{eq:CapsuleInternals4}
	&p_\Omega = p_{\Omega_{r_{\textit{final}}}} \\
	&\vec{\alpha}_\Omega = \vec{\alpha}_{\Omega_{r_{\textit{final}}}} \;\;\; . 
	\end{align}
	
\end{enumerate}

Note that $g(\gamma(\vec{\alpha})) = \vec{\tilde{\alpha}}$ is architecturally similar to an autoencoder \citep{Hinton:2006}, hence the names for $g$ and $\gamma$. However, unlike an autoencoder, we have a known interpretation for the latent variables (attributes).

\subsection{Agreement Function} 

In the following we discuss the agreement function found in Equation \ref{eq:CapsuleInternals2} in detail, which plays an integral role in both the capsule network as well as the meta-learning pipeline. Its main purpose is to give a measure of similarity between the input attributes of the encoder $\gamma$ and the output attributes of the decoder $g$, but on a semantic level. In this sense it is closely related to the reconstruction loss used in auto-encoders.  

Similarity of semantics implies the ability to be invariant to symmetries. A perfect square is indistinguishable from one that is rotated by $\frac{\pi}{2}$. In this case, a general $L^2$ loss would find a discrepancy numerically between attributes of two such squares, where there should be none semantically. Instead, we introduce a set $R_{\vec{\alpha}}$ of rotationally equivalent attributes $\vec{\alpha}$, which allows to propose an agreement function for semantic capsules:
\begin{equation}\label{eq:SemantsicAgreementFunction1}
Z\left(\vec{\alpha}_a, \vec{\alpha}_b\right) = \max\{ w\left(\vec{\alpha}_{b} - \vec{\hat{\alpha}}_{a} \right) \; : \; \vec{\hat{\alpha}}_a \in R_{\vec{\alpha}_{a}} \} \;\;\;,
\end{equation}
where $w \colon \mathbb{R}^n \to \mathbb{R}^n$ describes a separate window function for each individual attribute in the vector. 

The symmetries $R_{\vec{\alpha}}$ are found by explicitly by iterating over the possible $n$-fold rotational symmetries of $\frac{2\pi}{n}$. We choose some random set of attributes $\vec{\alpha}$, from which we produce two identical sets $\vec{\alpha}_{a}$ and $\vec{\alpha}_{b}$, but one with rotation $\vec{\alpha}_{a}^{\text{rot}} = 0$ and the other with $\vec{\alpha}_{b}^{\text{rot}} = \frac{2\pi}{n}$. The capsule is said to have a rotational symmetry, if
\begin{equation}\label{eq:SemantsicAgreementFunction2}
Z\left(g(\vec{\alpha}_a), g(\vec{\alpha}_b)\right) > \epsilon \;\;\;,
\end{equation}
holds true, where $\epsilon$ is some threshold and we use the agreement functions of the capsule's parts. In essence, we check if the attributes produce parts that are symmetric themselves by $\frac{2\pi}{n}$, which in turn check if their parts are symmetric all the way down the network until we reach the primitive capsules. In the case of our two squares, this would entail rendering them both and checking if the resulting images are similar.

For primitive capsules, finding an appropriate $Z$ is dependent on the decoder $g$, as we are dealing with pixel data. However, as $g$ is known for these capsules, we can incorporate the symmetries explicitly as well as invariance to semantically irrelevant lighting effects and noise.

With the primitive capsules' agreement function known, we recursively build up all the agreement functions of higher level capsules using Equation \ref{eq:SemantsicAgreementFunction2} for thresholding the equivalence class $[\alpha^i]$ of symmetries. This allows to generalize the agreement function of Equation \ref{eq:SemantsicAgreementFunction1} to all possible symmetries, for example a repetitive cloth texture with translational symmetries, by using
\begin{equation}\label{eq:SemantsicAgreementFunction3}
Z\left(\vec{\alpha}_a, \vec{\alpha}_b\right) = \max\{ w\left(\vec{\alpha}_b - \vec{\hat{\alpha}}_a \right) \; : \; \vec{\hat{\alpha}}_a \in [\vec{\alpha}_a] \} \;\;\;.
\end{equation}
We further discuss finding equivalence classes in the following chapter.

\subsection{Completeness}

As for the grammar, we must show that our capsule network is complete as well. By completeness we mean that the neural-symbolic capsule network must have the ability to detect every meaningful feature in the image and be able to give semantic meaning to any set of such features.

We again show this explicitly by following the proof of our grammar's completeness in reverse. The discrete cosine transform is an invertible function, thus a \symb{DCT} primitive capsule is able to decode any feature found in the image into a set of coefficients. To add some meaning, we again split the \symb{DCT} capsule into a \symb{DCT-without-edges-patches-\dots} primitive capsule and \symb{edge}\symb{patch}\dots primitive capsules. We, thus, construct feature detectors for edges, patches, etc., but also those we did not expect, such as noise, covering the entire spectrum of possibilities.

The feature detectors can be recombined as objects with a semantic meaning in a semantic capsule. Consider an encoder $\gamma \colon \mathbb{R}^m \to \mathbb{R}$, where $m$ is the total number of attributes from the feature detectors, that projects the whole $\mathbb{R}^m$ onto a line that passes through a point $p$ in $\mathbb{R}^m$. The point $p$ is defined by the configuration of features that were extracted. A semantic capsule with such an encoder $\gamma$ will detect this exact set of features as an object and even have a single attribute that allows it to generalize along the aforementioned line of configurations. We may always add a capsule, if we encounter a new object, enabling the capsule network to learn to detect every possible object and proving our requirement for completeness.

In the following chapter on meta-learning, we will explore in detail how this process can be performed in a more elegant way.

\subsection{Observation Table} 

Our final capsule network differs from other feed-forward networks by lacking a specific output layer. Instead, we refer to the set of capsules that have the same distance from the input as a hierarchy level, to make the difference to layers clear. Every capsule in the network acts as a network output and describes important semantics, even if it would be considered a hidden node in the sense of traditional neural networks. We use the notion of hierarchy level merely for orientation when talking about a network.

We also extend our capsule design by another functionality. As the first level of primitive capsules uses pixels as input attributes and this set may have a smaller size than the actual input image, we must handle scanning the entire image in a sensible way, \ie, sliding filters. To this end, each capsule in the network is amended with a list of objects with valid attributes it has detected, the observation table (\cf Figure \ref{fig:CapsuleBlockDiagram}). As the sliding filter moves across the image, valid detections are added. An entry is admitted, if its activation probability $p_\Omega$ is larger than a threshold $\rho$. This observation table is not persistent and the objects are stored in the table only per image. 

After detection by sliding filters, only the capsules in the first hierarchy level have entries in their observation tables. To perform a forward pass, we move up level-by-level and try out every permutation of input for the subsequent capsules, including setting them to $0$-capsules with vanishing activation. Adding $0$-capsules is equivalent to occlusion and, thus, to missing input. Every permutation that leads to a sufficiently large activation probability exceeding threshold $\rho$ adds an entry to the observation table. Once every permutation was tested for each capsule in the current hierarchy level, the process continues until the last level. This avoids the need for duplicates of the same capsule in the capsule network.

We refer to the set of observations of a capsule as the set $\Lambda_\Omega$. As these tables are reset at the beginning of each forward pass, we assume that all $\Lambda_\Omega$ of past observation tables are stored in more persistent memory as
\begin{equation}\label{eq:memoryTrainingSet}
\left((p_\lambda)^{(i)}, (\vec{\alpha}_\lambda)^{(i)},(\vec{\alpha}_\Omega)^{(i)}\right)_r \;\;\; .
\end{equation} 
This allows to calculate the mean value $\bar{p}_\Omega$ required by our routing-by-agreement protocol in Equation \ref{eq:CapsuleInternals2}. 

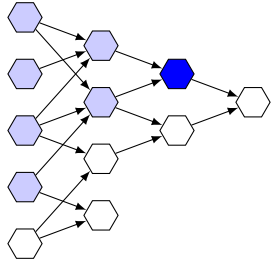
\begin{figure}[H]
	\centering
	\begin{adjustbox}{max width=1.0\textwidth}
		\begin{tikzpicture}
		
		%- First Caps Net
		\node (C01) [regular polygon,regular polygon sides=6, draw=black, fill=white] at (0,-1.5) {};
		\node (C02) [regular polygon,regular polygon sides=6, draw=black, fill=blue!20!white] at (0,-0.75) {};
		\node (C03) [regular polygon,regular polygon sides=6, draw=black, fill=blue!20!white] at (0,0) {};
		\node (C04) [regular polygon,regular polygon sides=6, draw=black, fill=blue!20!white] at (0,0.75) {};
		\node (C05) [regular polygon,regular polygon sides=6, draw=black, fill=blue!20!white] at (0,1.5) {};
		
		\node (C11) [regular polygon,regular polygon sides=6, draw=black, fill=white] at (1,-1.125) {};
		\node (C12) [regular polygon,regular polygon sides=6, draw=black, fill=white] at (1,-0.375) {};
		\node (C13) [regular polygon,regular polygon sides=6, draw=black, fill=blue!20!white] at (1,0.375) {};
		\node (C14) [regular polygon,regular polygon sides=6, draw=black, fill=blue!20!white] at (1,1.125) {};
		
		\node (C22) [regular polygon,regular polygon sides=6, draw=black, fill=white] at (2,0.0) {};
		\node (C23) [regular polygon,regular polygon sides=6, draw=black, fill=blue] at (2,0.75) {};
		
		\node (C31) [regular polygon,regular polygon sides=6, draw=black, fill=white] at (3,0.375) {};
		
		\draw[tochild, black] (C01) -- (C11);
		\draw[tochild, black] (C01) -- (C12);
		\draw[tochild, black] (C02) -- (C11);
		\draw[tochild, black] (C02) -- (C13);
		\draw[tochild, black] (C03) -- (C13);
		\draw[tochild, black] (C03) -- (C14);
		\draw[tochild, black] (C04) -- (C14);
		\draw[tochild, black] (C05) -- (C14);
		\draw[tochild, black] (C05) -- (C13);
		\draw[tochild, black] (C03) -- (C12);
		
		\draw[tochild, black] (C12) -- (C22);
		\draw[tochild, black] (C13) -- (C22);
		\draw[tochild, black] (C13) -- (C23);
		\draw[tochild, black] (C14) -- (C23);
		
		\draw[tochild, black] (C22) -- (C31);
		\draw[tochild, black] (C23) -- (C31);
		
		\end{tikzpicture}
	\end{adjustbox}
	\caption{A capsule network with an observed parse-tree formed by the activated capsules (blue). The top most activated capsule (dark blue) acts as the observed axiom of the observed grammar.} \label{fig:ObservedGrammar}
\end{figure}

During a forward pass, the subset of activated capsules and routes, \ie, the entries in the observation tables, form one or multiple parse-trees. The topmost activated capsule's symbol, however, is not necessarily an axiom of the capsule network's inverse grammar (\cf Figure \ref{fig:ObservedGrammar}). This is especially true if multiple parse-trees with distinct root nodes can be formed from the observation tables, as a grammar only produces one. To differentiate them, we refer to these parse-trees as observed parse-trees $\Lambda$, which incorporate all capsule observations $\Lambda_\Omega$. Each observed parse-tree also induces its own grammar with the root symbol being its axiom and we refer to these as the observed grammar and observed axiom.

\subsection{Episodic Memory and  Video}\label{sec:Video}

Verb attributes encode a sequence of poses on a time-frame of $0$ to $1$. Each pose in this sequence has to be seen in context of the previous, as a single pose may repeat itself in a sequence or in another verb attribute. In order to do this, we must extend the formulation of the capsules from detection based on single images to video data. 

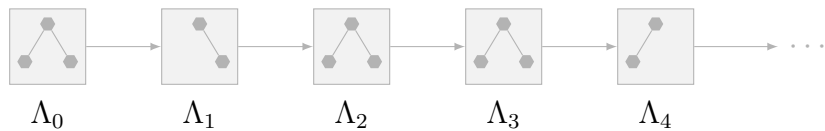
\begin{figure}[H]
	\centering
	\begin{adjustbox}{max width=1\textwidth}
		\begin{tikzpicture}

		%- Episodic Memory
		\node (E4) [fill=black!05!white, draw=black!30!white, minimum width=1cm, minimum height=1cm] at (1.8cm, 0.6cm) {};
		\node (E5) [fill=black!05!white, draw=black!30!white, minimum width=1cm, minimum height=1cm] at (3.8cm, 0.6cm) {};
		\node (E6) [fill=black!05!white, draw=black!30!white, minimum width=1cm, minimum height=1cm] at (5.8cm, 0.6cm) {};
		\node (E7) [fill=black!05!white, draw=black!30!white, minimum width=1cm, minimum height=1cm] at (7.8cm, 0.6cm) {};
		\node (E8) [fill=black!05!white, draw=black!30!white, minimum width=1cm, minimum height=1cm] at (9.8cm, 0.6cm) {};
		\node (EX) [text=black!30!white, minimum width=0.5cm, minimum height=0.5cm] at (11.8cm, 0.6cm) {$\cdots$};

		\node (X4) at (1.8cm, -0.3cm) {$\Lambda_{0}$};
		\node (X5) at (3.8cm, -0.3cm) {$\Lambda_{1}$};
		\node (X6) at (5.8cm, -0.3cm) {$\Lambda_{2}$};
		\node (X7) at (7.8cm, -0.3cm) {$\Lambda_{3}$};
		\node (X8) at (9.8cm, -0.3cm) {$\Lambda_{4}$};

		\draw[tochild, black!30!white] (E4) -- (E5);
		\draw[tochild, black!30!white] (E5) -- (E6);
		\draw[tochild, black!30!white] (E6) -- (E7);
		\draw[tochild, black!30!white] (E7) -- (E8);
		\draw[tochild, black!30!white] (E8) -- (EX);
				
		\node (E4Cap1) [regular polygon,regular polygon sides=6, fill=black!30!white, minimum width=1cm, minimum height=1cm, scale=0.2] at (1.8cm, 0.9cm) {};
		\node (E4Cap2) [regular polygon,regular polygon sides=6, fill=black!30!white, minimum width=1cm, minimum height=1cm, scale=0.2] at (1.5cm, 0.4cm) {};
		\node (E4Cap3) [regular polygon,regular polygon sides=6, fill=black!30!white, minimum width=1cm, minimum height=1cm, scale=0.2] at (2.1cm, 0.4cm) {};
		\draw[black!30!white, scale=0.2] (E4Cap1) -- (E4Cap2);
		\draw[black!30!white, scale=0.2] (E4Cap1) -- (E4Cap3);
		
		\node (E5Cap1) [regular polygon,regular polygon sides=6, fill=black!30!white, minimum width=1cm, minimum height=1cm, scale=0.2] at (3.8cm, 0.9cm) {};
		\node (E5Cap2) [regular polygon,regular polygon sides=6, fill=black!30!white, minimum width=1cm, minimum height=1cm, scale=0.2] at (4.1cm, 0.4cm) {};
		\draw[black!30!white, scale=0.2] (E5Cap1) -- (E5Cap2);
		
		\node (E6Cap1) [regular polygon,regular polygon sides=6, fill=black!30!white, minimum width=1cm, minimum height=1cm, scale=0.2] at (5.8cm, 0.9cm) {};
		\node (E6Cap2) [regular polygon,regular polygon sides=6, fill=black!30!white, minimum width=1cm, minimum height=1cm, scale=0.2] at (5.5cm, 0.4cm) {};
		\node (E6Cap3) [regular polygon,regular polygon sides=6, fill=black!30!white, minimum width=1cm, minimum height=1cm, scale=0.2] at (6.1cm, 0.4cm) {};
		\draw[black!30!white, scale=0.2] (E6Cap1) -- (E6Cap2);
		\draw[black!30!white, scale=0.2] (E6Cap1) -- (E6Cap3);
		
		\node (E7Cap1) [regular polygon,regular polygon sides=6, fill=black!30!white, minimum width=1cm, minimum height=1cm, scale=0.2] at (7.8cm, 0.9cm) {};
		\node (E7Cap2) [regular polygon,regular polygon sides=6, fill=black!30!white, minimum width=1cm, minimum height=1cm, scale=0.2] at (7.5cm, 0.4cm) {};
		\node (E7Cap3) [regular polygon,regular polygon sides=6, fill=black!30!white, minimum width=1cm, minimum height=1cm, scale=0.2] at (8.1cm, 0.4cm) {};
		\draw[black!30!white, scale=0.2] (E7Cap1) -- (E7Cap2);
		\draw[black!30!white, scale=0.2] (E7Cap1) -- (E7Cap3);
		
		\node (E8Cap1) [regular polygon,regular polygon sides=6, fill=black!30!white, minimum width=1cm, minimum height=1cm, scale=0.2] at (9.8cm, 0.9cm) {};
		\node (E8Cap2) [regular polygon,regular polygon sides=6, fill=black!30!white, minimum width=1cm, minimum height=1cm, scale=0.2] at (9.5cm, 0.4cm) {};
		\draw[black!30!white, scale=0.1] (E8Cap1) -- (E8Cap2);
		
		\end{tikzpicture}
	\end{adjustbox}
	\caption{Schematic look at Episodic Memory. All past observed parse-trees $\Lambda_t$ are chronologically ordered.} \label{fig:EpisodicMemory}
\end{figure}

In order to make sense of video data, the capsules require the ability to track objects across frames, to be able to represent concepts such as continuity and permanence of objects. We expand the observation table $\Lambda_{\Omega}$ of a capsule $\Omega$ to include observations at various time-steps $t$, which we denote as $\Lambda_{\Omega, t}$, and store these past observations in Episodic Memory (\cf Figure \ref{fig:EpisodicMemory}), as part of the observed parse-tree $\Lambda_t$.

We denote the attributes of the $i$th observed object in $\Lambda_{\Omega}$ as $(\vec{\alpha}_\Omega)^{(i)}_t$, in accordance with Equation \ref{eq:memoryTrainingSet}. If two observations,  $(\vec{\alpha}_\Omega)^{(i)}_{t-1}$ at time $t-1$ and  $(\vec{\alpha}_\Omega)^{(j)}_{t}$ at time $t$, describe the same object, \ie, it is tracked across the two frames, we denote this relation by $(i,j)$. If an object is not tracked from the previous frame, or has entered the frame, we denote this as $(i, \cdot)$ and $(\cdot, j)$ respectively. All relations of a specific object class form the set $P_{\Omega,t}$. 

For now, we assume that the camera moves smoothly and the time between frames $\Delta t$ is small. This allows to describe an object's position and orientation in relation to the last frame based on small changes of the camera. We encode these changes in a camera rotation matrix $\textbf{R}_t$ and a translation vector $\vec{x}_t$ in relation to the last frame. Both, the angles and the distance described by these quantities are small. 

Further, we are able to predict the movement of objects based on their past locations with sufficient accuracy, by using a finite Taylor expansion of the form:
\begin{equation}\label{eq:Taylor}
(\vec{\alpha}_\Omega)^{(i)}_t \approx (\vec{\alpha}_\Omega)^{(i)}_{t-1} + \frac{\delta (\vec{\alpha}_\Omega)^{(i)}_{t-1}}{\delta t}\Delta t + \frac{1}{2} \frac{\delta^2 (\vec{\alpha}_\Omega)^{(i)}_{t-1}}{\delta t^2}\Delta t^2 + \cdots \;\;\;.
\end{equation}
However, as we use the camera as the frame of reference, we need to transform it from the previous frame of reference
\begin{equation}\label{eq:CameraTransform}
(\vec{\alpha}_\Omega)^{(i)}_t \to\; \textbf{R}_t \cdot (\vec{\alpha}_\Omega)^{(i)}_t + \vec{x}_t \;\;\;,
\end{equation}
where we let $\textbf{R}_t$ and $\vec{x}_t$ only act on the position and rotation attributes.

Finally, we can formulate our tracking problem. By $P_{\Omega,t}$ we denoted the set of true relations between objects in frame $t-1$ and $t$. Assuming that $P_{\Omega,t}$ is unknown, let $\tilde{P}_{\Omega,t}^k \in \mathcal{P}_\Omega$ denote a set of unique relations, \ie, each index may only appear on the left-hand-side and right-hand-side once, including those of the type $(i, \cdot)$, $(\cdot, j)$ and let $\mathcal{P}_\Omega$ be the set of all such possible sets. Note that the true relations $P_{\Omega,t}$ are also in $\mathcal{P}_\Omega$, which we approximate using the first order expansion of Equation \ref{eq:Taylor} combined with Equation \ref{eq:CameraTransform}:
\begin{equation}\label{eq:Tracking}
P_{\Omega,t} = \min_{\tilde{P}_{\Omega,t}^k\in \mathcal{P}_\Omega, \textbf{R}_t, \vec{x}_t} \left\{ \sum_{(i,j) \in \tilde{P}_{\Omega,t}^k} \left\lVert \vec{w} \cdot \left( (\vec{\alpha}_\Omega)^{(i)}_{t-1} + \frac{\delta (\vec{\alpha}_\Omega)^{(i)}_{t-1}}{\delta t}\Delta t - \left[\textbf{R}_t\cdot(\vec{\alpha}_\Omega)^{(j)}_t + \vec{x}_t \right] \right) \right\lVert \right\} \;\;\;,
\end{equation}
where $\vec{w}$ assigns a weight to each attribute in accordance with its importance for tracking and we again let $\textbf{R}_t$ and $\vec{x}_t$ only act on the position and rotation attributes. For example, adjective attributes are less likely to change across frames and, thus, provide a good indicator for tracking, whereas verb attributes are prone to change.

By limiting ourselves to the first order expansion, we introduce an error of the order $(\Delta t)^2$. Thus, reducing the time-steps between frames does not only reduce the movement of the camera and the effect of Equation \ref{eq:CameraTransform}, but also the error of the prediction of the object's movement. Minimizing Equation \ref{eq:Tracking} is, thus, dominated by checking all possible combinations and of complexity $\mathcal{O}(n!)$. However, each search is limited to a single object class with $n_\Omega=\vert\Lambda_\Omega\vert$ entries, considerably reducing the search space.

Particularly, we use the fact that each search is limited to $n_\Omega$ and begin tracking objects that are are in the scene, by sorting the classes $n_{\Omega_1} < n_{\Omega_2} < \cdots$. This allows to get an early estimate for the camera movement ($\textbf{R}_t$ and $\vec{x}_t$), which we refine as we track more frequent objects to combat the problem of crowding. Once the complexity of $\mathcal{O}(n!)$ becomes an issue, we may use our refined knowledge of the camera movement and further reduce the complexity to $\mathcal{O}(n^2)$, by performing an initial prediction
\begin{equation}\label{eq:nextStep1}
(\tilde{\vec{\alpha}}_\Omega)_t = \textbf{R}^{-1}_t \cdot  \left[(\vec{\alpha}_\Omega)^{(i)}_{t-1} + \frac{\delta (\vec{\alpha}_\Omega)^{(i)}_{t-1}}{\delta t}\Delta t - \vec{x}_t \right] 
\end{equation}
and, assuming that the tracked object must be close, search individual relations directly using
\begin{equation}\label{eq:nextStep2}
\lVert (\vec{\alpha}_\Omega)^{(j)}_t - (\tilde{\vec{\alpha}}_\Omega)_t \lVert < \epsilon \cdot \Delta t^2 \;\;\;.
\end{equation}

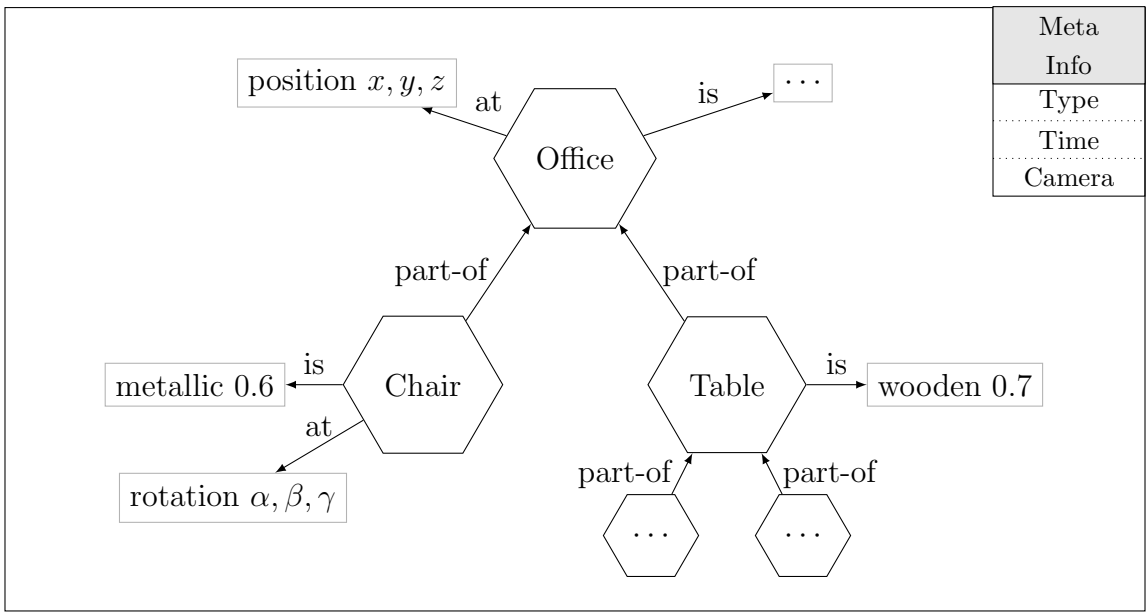
\begin{figure}[H]
	\centering
	\begin{adjustbox}{max width=1\textwidth}
		\begin{tikzpicture}		
		
		\node (Box) [draw, minimum width=15cm, minimum height=8cm] at (0.0, -1.0) {};
		\node (MetaBox) [draw, minimum width=2cm, minimum height=2.5cm] at (6.5, 1.75) {};
		\node (MetaText) [draw, fill=black!10!white, minimum width=2cm, minimum height=1cm, text width=1.7cm, align=center] at (6.5, 2.5) {\footnotesize{Meta\\Info}};
		\node (MetaText) [minimum width=2cm, minimum height=0.5cm, text width=1.7cm, align=center] at (6.5, 1.75) {\footnotesize{Type}};
		\draw [draw, dotted] (5.5,1.5) -- (7.5,1.5);
		\node (MetaText) [minimum width=2cm, minimum height=0.5cm, text width=1.7cm, align=center] at (6.5, 1.25) {\footnotesize{Time}};
		\draw [draw, dotted] (5.5,1.0) -- (7.5,1.0);
		\node (MetaText) [minimum width=2cm, minimum height=0.5cm, text width=1.7cm, align=center] at (6.5, 0.75) {\footnotesize{Camera}};
		
		\node (Office) [regular polygon,regular polygon sides=6, draw, minimum width=0.5cm, minimum height=0.5cm] at (0.0, 1.0) {Office};		
		\node (Table) [regular polygon,regular polygon sides=6, draw, minimum width=0.5cm, minimum height=0.5cm] at (2.0, -2.0) {Table};
		\node (Chair) [regular polygon,regular polygon sides=6, draw, minimum width=0.5cm, minimum height=0.5cm] at (-2.0, -2.0) {Chair};
		\node (TableP1) [regular polygon,regular polygon sides=6, draw, minimum width=0.5cm, minimum height=0.5cm] at (3.0, -4.0) {$\cdots$};
		\node (TableP2) [regular polygon,regular polygon sides=6, draw, minimum width=0.5cm, minimum height=0.5cm] at (1.0, -4.0) {$\cdots$};
		
		\path [tochild] (Table) -- node [right,midway] {part-of} (Office);
		\path [tochild] (Chair) -- node [left,midway] {part-of} (Office);
		\path [tochild] (TableP1) -- node [right,midway] {part-of} (Table);
		\path [tochild] (TableP2) -- node [left,midway] {part-of} (Table);
		%\path [draw,latex'-latex'] (Chair) -- node [above,midway] {collides} (Table);

		\node (Adj1) [draw=black!30!white, minimum width=0.5cm, minimum height=0.5cm] at (5.0, -2.0) {wooden $0.7$};
		\node (Adj2) [draw=black!30!white, minimum width=0.5cm, minimum height=0.5cm] at (-5.0, -2.0) {metallic $0.6$};
		\node (Adj3) [draw=black!30!white, minimum width=0.5cm, minimum height=0.5cm] at (3.0, 2.0) {$\cdots$};
		\node (Prep1) [draw=black!30!white, minimum width=0.5cm, minimum height=0.5cm] at (-3.0, 2.0) {position $x,y,z$};
		\node (Prep2) [draw=black!30!white, minimum width=0.5cm, minimum height=0.5cm] at (-4.5, -3.5) {rotation $\alpha,\beta,\gamma$};
		
		\path [tochild] (Chair) -- node [above,midway] {is} (Adj2);
		\path [tochild] (Table) -- node [above,midway] {is} (Adj1);
		\path [tochild] (Office) -- node [above,midway] {is} (Adj3);
		\path [tochild] (Office) -- node [above right,midway] {at} (Prep1);
		\path [tochild] (Chair) -- node [above,midway] {at} (Prep2);
		
		\end{tikzpicture}
	\end{adjustbox}
	\caption{Example scene graph for an office scene extracted from capsule observation tables and meta-information.} \label{fig:SemanticNetwork}
\end{figure}

Intuitive physics introduced later in chapter 6 allows additionally to incorporate physical predictions into Equation \ref{eq:Tracking}. This improves tracking objects that may be occluded for a short while and reappear, \ie, object permanence. With all this information available for each observation, each entry in episodic memory has a form similar to the example shown in Figure \ref{fig:SemanticNetwork}, promoting the parse-tree to a scene-graph.

\subsection{Graphics Engine}

The conversion of a grammar into a capsule network works both ways. We may always invert the network back into a generative grammar to produce an image and vice versa. In this sense, both representations are dual to each other. However, there can only be one grammar axiom, but as discussed previously, each capsule itself may be an observed axiom. We must, thus, choose one as the starting point, the compound noun that describes the scene we want to produce.

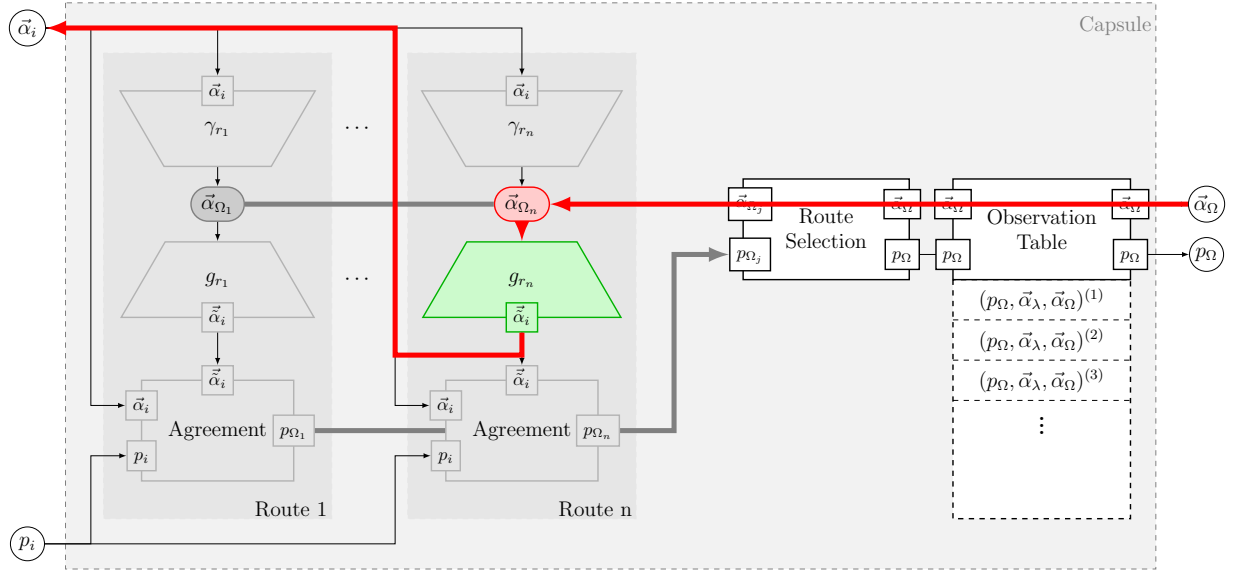
\begin{figure}[H]
	\centering
	\begin{adjustbox}{max width=1\textwidth}
		\begin{tikzpicture}
		
		%- Bounding Boxes
		\draw[draw=black!50, dashed, fill=black!5] (-30mm,40mm) rectangle (185mm, -72.5mm);
		\draw[draw=black!10!, dashed, fill=black!10] (-22.5mm,30mm) rectangle (22.5mm, -62.5mm);
		\draw[draw=black!10!, dashed, fill=black!10] (37.5mm,30mm) rectangle (82.5mm, -62.5mm);
		
		%- Inputs		
		\node (InA) [whitenodefill] at (-37.5mm, 35mm) {$\vec{\alpha}_i$};
		\node (InP) [whitenodefill] at (-37.5mm, -67.5mm) {$p_i$};
		
		%- First Autoencoder
		\node (A01) [trapezium, trapezium angle=-60, minimum height=15mm, draw=black!30, fill=black!10, thick] at (0, 15mm) {$\gamma_{r_1}$};
		\node (A02) [draw=black!50, fill=black!20, rounded rectangle, thick] at (0, 0) {$\vec{\alpha}_{\Omega_1}$};
		\node (A03) [trapezium, trapezium angle=60, minimum height=15mm, draw=black!30, fill=black!10, thick] at (0,-15mm) {$g_{r_1}$};
		\draw [tochild, black] (A01) -- (A02);
		\draw [tochild, black] (A02) -- (A03);
		\node (A0i) [rectangle, draw=black!30, text height=2.5mm, fill=black!10, thick] at (0, 22.5mm) {\footnotesize $\vec{\alpha}_i$};
		\node (A0o) [rectangle, draw=black!30, text height=2.5mm, fill=black!10, thick] at (0, -22.5mm) {\footnotesize $\vec{\tilde{\alpha}}_i$};
		
		%- First agreement
		\node (A0m) [rectangle, draw=black!30, fill=black!10, minimum height = 20mm, minimum width = 30mm, thick] at (0, -45mm) {Agreement};
		\node (A0mi1) [rectangle, draw=black!30, text height=2.5mm, fill=black!10, thick] at (0, -35mm) {\footnotesize $\vec{\tilde{\alpha}}_i$};
		\node (A0mi2) [rectangle, draw=black!30, text height=2.5mm, fill=black!10, thick] at (-15mm, -40mm) {\footnotesize $\vec{\alpha}_i$};
		\node (A0mi3) [rectangle, draw=black!30, text height=2.5mm, fill=black!10, thick] at (-15mm, -50mm) {\footnotesize $p_i$};
		\node (A0mo1) [rectangle, draw=black!30, text height=2.5mm, fill=black!10, thick] at (15mm, -45mm) {\footnotesize $p_{\Omega_1}$};
		\draw [tochild, black] (A0o) -- (A0mi1);
		
		%- ith Autoencoder
		\node (Ai1) at (27.5mm, 15mm) {$\cdots$};
		\node (Ai3) at (27.5mm, -15mm) {$\cdots$};
		
		%- nth Autoencoder
		\node (An1) [trapezium, trapezium angle=-60, minimum height=15mm, draw=black!30, fill=black!10, thick] at (60mm, 15mm) {$\gamma_{r_n}$};
		\node (An2) [draw=red, fill=red!20, rounded rectangle, thick] at (60mm, 0) {$\vec{\alpha}_{\Omega_n}$};
		\node (An3) [trapezium, trapezium angle=60, minimum height=15mm, draw=black!30!green, fill=black!10!green!20, thick] at (60mm,-15mm) {$g_{r_n}$};
		\draw [tochild, black] (An1) -- (An2);
		\node (Ani) [rectangle, draw=black!30, text height=2.5mm, fill=black!10, thick] at (60mm, 22.5mm) {\footnotesize $\vec{\alpha}_i$};
		\node (Ano) [rectangle, draw=black!30!green, text height=2.5mm, fill=black!10!green!20, thick] at (60mm, -22.5mm) {\footnotesize $\vec{\tilde{\alpha}}_i$};
		
		%- nth agreement
		\node (Anm) [rectangle, draw=black!30, fill=black!10, minimum height = 20mm, minimum width = 30mm, thick] at (60mm, -45mm) {Agreement};
		\node (Anmi1) [rectangle, draw=black!30, text height=2.5mm, fill=black!10, thick] at (60mm, -35mm) {\footnotesize $\vec{\tilde{\alpha}}_i$};
		\node (Anmi2) [rectangle, draw=black!30, text height=2.5mm, fill=black!10, thick] at (45mm, -40mm) {\footnotesize $\vec{\alpha}_i$};
		\node (Anmi3) [rectangle, draw=black!30, text height=2.5mm, fill=black!10, thick] at (45mm, -50mm) {\footnotesize $p_i$};
		\node (Anmo1) [rectangle, draw=black!30, text height=2.5mm, fill=black!10, thick] at (75mm, -45mm) {\footnotesize $p_{\Omega_n}$};
		\draw [tochild, black] (Ano) -- (Anmi1);
		
		%- Connecting Inputs to Autoencoders
		\draw[tochild, black] (InA) -- (0, 35mm) -- (A0i);
		\draw[tochild, black] (InA) -- (60mm, 35mm) -- (Ani);
		\draw[tochild, black] (InA) -- (-25mm, 35mm) -- (-25mm, -40mm) -- (A0mi2);
		\draw[tochild, black] (InP) -- (-25mm, -67.5mm) -- (-25mm, -50mm) -- (A0mi3);
		\draw[tochild, black] (InP) -- (35mm, -67.5mm) -- (35mm, -50mm) -- (Anmi3);
		\draw[tochild, black] (InA) -- (35mm, 35mm) -- (35mm, -40mm) -- (Anmi2);
		
		%- Outputs
		\node (OutABox) [rectangle, draw=black, text height=2.5mm, text width=30mm, thick, fill=white, minimum height = 20mm, minimum width = 30mm, align=center] at (120mm, -5mm) {Route\\Selection};
		\node (OutAi1) [rectangle, draw=black, fill=white, text height=2.5mm, thick] at (105mm, 0) {\footnotesize $\vec{\alpha}_{\Omega_j}$};
		\node (OutAi2) [rectangle, draw=black, fill=white, text height=2.5mm, thick] at (105mm, -10mm) {\footnotesize $p_{\Omega_j}$};
		\node (OutAo1) [rectangle, draw=black, fill=white, text height=2.5mm, thick] at (135mm, 0mm) {\footnotesize $\vec{\alpha}_\Omega$};
		\node (OutPo1) [rectangle, draw=black, fill=white, text height=2.5mm, thick] at (135mm, -10mm) {\footnotesize $p_\Omega$};
		
		\node (OutA) [whitenodefill] at (195mm, -0mm) {$\vec{\alpha}_\Omega$};
		\node (OutP) [whitenodefill] at (195mm, -10mm) {$p_\Omega$};
		
		%- Connecting Outputs
		\draw[tochild, black!50, line width=1mm] (Anmo1) -- (90mm,-45mm) -- (90mm,-10mm) -- (OutAi2);
		\draw[black!50, line width=1mm] (A0mo1) -- (Anm);
		\draw[black!50, line width=1mm] (A02) -- (An2);
		\draw[tochild, black] (150mm, -0mm) -- (OutA);
		\draw[tochild, black] (150mm, -10mm) -- (OutP);
		\draw[tochild, black] (OutAo1) -- (150mm, -0mm);
		\draw[tochild, black] (OutPo1) -- (150mm, -10mm);
		
		%- Labels
		\node[text=black!50] at (177mm, 35.5mm) {Capsule};
		\node[text=black!50!black] at (14.5mm, -60.5mm) {Route 1};
		\node[text=black!50!black] at (74.5mm, -60.5mm) {Route n};
		
		%- Observation Table
		
		\draw[draw=black, fill=white, thick, dashed] (145mm, -15mm) rectangle (180mm, -62.5mm);
		\draw[draw=black, fill=white, thick] (145mm, 5mm) rectangle (180mm, -15mm);
		\node[text=black, text width=30mm, align=center] at (162.5mm, -5mm) {Observation\\Table};
		\node[text=black] at (162.5mm, -19mm) {$(p_\Omega, \vec{\alpha}_\lambda, \vec{\alpha}_\Omega)^{(1)}$};
		\node[text=black] at (162.5mm, -27mm) {$(p_\Omega, \vec{\alpha}_\lambda, \vec{\alpha}_\Omega)^{(2)}$};
		\node[text=black] at (162.5mm, -35mm) {$(p_\Omega, \vec{\alpha}_\lambda, \vec{\alpha}_\Omega)^{(3)}$};
		\node[text=black] at (162.5mm, -43mm) {\textbf{$\vdots$}};
		\draw[draw=black, dashed] (145mm, -23mm) -- (180mm, -23mm);
		\draw[draw=black, dashed] (145mm, -31mm) -- (180mm, -31mm);
		\draw[draw=black, dashed] (145mm, -39mm) -- (180mm, -39mm);
		
		\node (InAL) [rectangle, draw=black, fill=white, text height=2.5mm, thick] at (145mm, 0mm) {\footnotesize $\vec{\alpha}_\Omega$};
		\node (InPL) [rectangle, draw=black, fill=white, text height=2.5mm, thick] at (145mm, -10mm) {\footnotesize $p_\Omega$};
		\node (OutAL) [rectangle, draw=black, fill=white, text height=2.5mm, thick] at (180mm, 0mm) {\footnotesize $\vec{\alpha}_\Omega$};
		\node (OutPL) [rectangle, draw=black, fill=white, text height=2.5mm, thick] at (180mm, -10mm) {\footnotesize $p_\Omega$};
		
		%- REVERSE:
		
		\draw[toparent, red, line width=1mm] (An2) -- (OutA);
		\draw[tochild, red, line width=1mm] (An2) -- (An3);
		\draw[tochild, red, line width=1mm] (Ano) -- (60mm,-30mm) -- (35mm,-30mm) -- (35mm,35mm) -- (InA);
		
		\end{tikzpicture}
	\end{adjustbox}
	\caption{Reversing the direction of a capsule to generate instead to detect. The feed-backward process is highlighted in red.} \label{fig:CapsuleInReverse}
\end{figure}

To produce the image, we follow the production rules, \ie, the inverted routes. A rule requires attributes, which must be provided alongside the chosen axiom and a decoder $g$, which we obtain directly from the route as shown in Figure \ref{fig:CapsuleInReverse}. The starting axiom with these attributes is a compound noun and a collection of verbs and adjectives. It can also be formulated as a full description of the final image. As an example, a \symb{car} with high values for \attr{rustic} and \attr{red} may be translated into a simple description, such as "a rustic red car".

Capsules, however, may have multiple routes, each expressing a different viewpoint, style or configuration. Thus, during production, a choice must be made between the valid rules to determine which to use. This can be either the original image's selection stored in memory or a new one. This  choice alters the final image, but not its semantic content as laid out by the chosen axiom and its attributes. Which rules/routes are valid is explored in more detail in the next chapter. 

Once the generation process of the grammar has reached the terminal symbols, the final image is rendered using the decoders of the primitive capsules. However, for rendering, depth ordering is required. This is provided by the mean activation probability $\bar{p}_\Omega$ stored for each symbol, which was determined according to the objects' visibility in feed-forward and now provides depth information for feed-backward operation. 

Finally, by repeating this rendering process over multiple time-steps and smoothly varying the verb attributes of the axiom accordingly, video output is produced. By encoding a scene into a symbolic representation with the ability to render it later, we are essentially compressing the visual data for video playback.

\clearpage

\chapter{Training and Meta-Learning}\label{sec:Meta}\thispagestyle{empty}

\begin{wrapfigure}{R}{0.4\textwidth}
	\centering
	\begin{adjustbox}{max width=0.4\textwidth}
		\begin{tikzpicture}
		
		\vividnet{black!20!white}{black!20!white}{black!20!white}{black!20!white}{black!10!blue}{black!20!white}{black!10!blue}

		\end{tikzpicture}
	\end{adjustbox}
\end{wrapfigure}

Statically defining an entire neural-symbolic capsule network from scratch is not feasible, as it would require encoding a lot of information beforehand and would not take advantage of the benefits of learning. Instead, we propose a meta-learning algorithm that designs the architecture of the network automatically based on unknown image sequences it is exposed to and introduce an accompanying training algorithm to quickly learn from new environments over its entire lifespan.

The general idea and motivation for our lifelong meta-learning algorithm \citep{Chen:2018} is based on that our capsule network works as an inverted grammar and as such should also only have a single observed axiom that describes the scene. We postulate having a single observed axiom as the overall goal of the meta-learning pipeline and deduce further training rules, which we show to be sufficient to keep improving the capsule network indefinitely with each new scene it encounters.
\clearpage
\section{Choosing and Training Primitive Capsules}

Our capsule network follows the philosophical idea of innatism, \ie, it does not start out as a blank slate (or tabula rasa), but rather has a preexisting idea of the world and its structure. Specifically, we assume that all primitive capsules are predefined.

We will ignore the discrete cosine transform capsules introduced in our completeness proof for our analysis here and focus on those that extract specific features. Without a \symb{DCT} capsule, the initial set of primitive capsules limits what the network can actually analyze and learn. For example, if we only define \symb{circle} primitive capsules, the network will have a hard time to make sense of objects made up of squares (it would assume squares are made out of multiple $1\times 1$-sized circles, which is highly inefficient). Ideally, we would define primitive capsules for the most basic set of primitives from which we are able to construct every kind of object. Such a set might include different types of surface-patches and edges, from which we then can infer more complex structures on subsequent hierarchy levels. Such an approach requires a lot of time to learn, as the network would need to understand, for example, that multiple patches bounded by edges form a surface, multiple surfaces define a square and multiple squares form a table. 

Alternatively, we may begin with more complex primitives from which we can infer objects using fewer hierarchy levels. For example, if we directly begin with a \symb{square}, the following level may already contain a semantic capsule describing a \symb{table} without the need for all the intermediary capsules. This is adequate for an environment full of squares, such as a tetris-like game. By adding a \symb{triangle} capsule, it would then be possible to detect further compound shapes, such as a simple \symb{house}. Should the environment become so complex (general polygons, organic shapes, \dots) that the initial set of geometric shapes isn't sufficient, we can remedy this by either adding more primitive capsules or moving down one level of abstraction, promoting the previous primitives to semantic capsules and introducing new symbols, such as an \symb{edge}. This is possible through the modularity of the network and apart from those modified, no other capsule needs to be retrained.

While we do not restrict the complexity of the primitive, it should be noted that there is a strong relation between the complexity of a renderable primitive and the design of $\gamma$ and $g$. $\gamma$ is only bounded by current regression learning methods and particularly $n$-dimensional pose estimation. For $g$, we rely on the current state of computer graphics. Here we have access to a near endless supply of 2D and 3D art assets \citep{Quilez:2017, Zhou:2018} and physically-based rendering pipelines \citep{Pharr:2016}. Alternatively, we could also make use of previously trained CNNs, as they often have a latent representation for primitives in the early layers. However, it has proven difficult to assign a specific attribute to each such latent variable due to entanglement.

Once a primitive has been chosen for $g$ to render, we are able to train $\gamma$. Let $\chi_{i,j} \sim U( [0,1])$ be a uniform random variable for the $j$th attribute and $f(\cdot)$ describe the application of random backgrounds, occlusions and special effects to an image. We train $\gamma$ using the synthetic dataset
\begin{equation}
\left((f(g(\chi_{i,j})))^{(i)}, (\chi_{i,j})^{(i)} \right)\;\;\;.
\end{equation}

To speed up the training, we may also use quantile functions $Q_j$ for each attribute that better represent their actual distribution. However, this is neither required nor always possible and we set them to $Q_j(p) = p$ in this case. Using these quantile functions, $\gamma$ is trained with the set
\begin{equation}
\left((f(g(Q_j(\chi_{i,j}))))^{(i)}, (Q_j(\chi_{i,j}))^{(i)} \right)\;\;\;.
\end{equation}

\section{Training Semantic Capsules}

Before we describe our meta-learning pipeline for semantic capsules, we explore some mathematical details. Each semantic capsule takes a configuration of its parts as input and outputs the corresponding object's attributes. Both, the input configuration of the parts as well as the object's attributes lie in a configuration space, also referred to as the latent space. However, as the exact structure of this space is unknown, our goal is to fit the capsule such that it best approximates this configuration space. This is not without ambiguity, but by exploring the possible topologies, we derive methods to perform the best approximation and introduce our meta-learning pipeline. 

\subsection{Configuration Spaces}

The configuration space of the input attributes as well as the output attributes of a semantic capsule describe the plausible configurations of an object and this is often referred to as the manifold hypothesis. However, in our case, this name is misleading, as the underlying space is not always a manifold and rather a more general metric space. Here, we explore the topological structure of these spaces and the consequences for our meta-learning pipeline they entail. 

We denote the configuration space of the inputs as $A_\lambda$ and the configuration space of the outputs as $A_\Omega$. They are embedded in some larger spaces $A_\lambda\subseteq M_\lambda$ and $A_\Omega\subseteq M_\Omega$, which describe all possible attributes, but may contain un-plausible configurations that lie outside of $A_\lambda$ and $A_\Omega$. Using this notation, the capsule's decoder is a map $g : M_\Omega \to M_\lambda$. The image of this decoder describes an approximated input configuration space $g(M_\Omega) = \tilde{A}_\lambda$, with $\tilde{A}_\lambda \subseteq M_\lambda$. Thus, ideally we would find that $\tilde{A}_\lambda = A_\lambda$ by minimizing Equation \ref{eq:minimization} and introduce a distance measure for attributes
\begin{equation}\label{eq:dist}
d(\vec{\alpha}_{\lambda,k}) = ||g(\gamma(\vec{\alpha}_{\lambda,k})) - \vec{\alpha}_{\lambda,k}|| \;\;\; ,
\end{equation}
of how far off we are of the correct configuration.

This also allows us to find an upper bound $d_{\tilde{A}}$ for the distance between the approximated configuration space $\tilde{A}_\lambda$ and any point in the true configuration space $A_\lambda$ as

\begin{equation}\label{eq:maxdistance}
d_\max = \sup_{\vec{\alpha} \in A_\lambda} d(\vec{\alpha}) \;\;\; .
\end{equation}

\subsection{Topology of $M_\Omega$ and $M_\lambda$}

In the simplest case, there are two types of attributes, those that lie in the range of $[0, 1]$ (such as position) and those that lie in the range of $[0, 1]$ but are circular (such as rotation). The latter can be seen as $0$ and $1$  glued together to form $[0,1]/ \{0,1\}$, which is homeomorphic to the unit circle $S^1$. 

For this simple case, the spaces for all input and output attributes take on the following form:
\begin{equation}\label{eq:flatConfigSpace}
M_\Omega = \prod_i [0,1] \times \prod_j S^1 \;\;\; ,
\end{equation}
\begin{equation}
M_\lambda = \prod_\lambda \left[ \prod_{i_\lambda} [0,1] \times \prod_{j_\lambda} S^1 \right] \;\;\; .
\end{equation}

\begin{figure}[H]
	\centering
	\begin{adjustbox}{max width=0.6\textwidth}
	\begin{tikzpicture}

	% Square
	\node at (-9.5, 8.0) {$[0,1] \times [0,1]$};
	\node at (-7.75, 8.0) {$\simeq$};
	\node at (-7.0, 8.0) {$\alpha^j_\Omega$};
	\node at (-5.5, 6.5) {$\alpha^i_\Omega$};
	\draw (-6.5, 7.0) rectangle (-4.5, 9.0);

	% Cylinder - Left
	\node at (-9.5, 4.0) {$[0,1] \times S^1$};
	\node at (-7.75, 4.0) {$\simeq$};
	\node at (-7.0, 4.0) {$\alpha^j_\Omega$};
	\node at (-5.5, 2.5) {$\alpha^i_\Omega$};	
	\draw (-6.5, 3.0) rectangle (-4.5, 5.0);
	\draw[-{latex[length=2mm]}, red, very thick] (-6.5, 3.0) -- (-6.5, 5.0);
	\draw[-{latex[length=2mm]}, red, very thick, dashed] (-4.5, 3.0) -- (-4.5, 5.0);
		
	\node at (-3.75, 4.0) {$\simeq$};
	
	% Cylinder - Right
	\draw (-1.0,4.75) arc(0:360:1cm and 0.5cm);
	\draw (-3.0,3.25) arc(180:360:1cm and 0.5cm);
	\draw (-3.0,3.25) -- (-3.0,4.75);
	\draw (-1.0,3.25) -- (-1.0,4.75);
	\draw[red, very thick] (-1.8,2.77) -- (-1.8,4.27);

	% Torus - Left
	\node at (-9.5, 0.0) {$S^1 \times S^1$};
	\node at (-7.75, 0.0) {$\simeq$};
	\node at (-7.0, 0.0) {$\alpha^j_\Omega$};
	\node at (-5.5, -1.5) {$\alpha^i_\Omega$};	
	\draw (-6.5, -1.0) rectangle (-4.5, 1.0);
	\draw[-{latex[length=2mm]}, red, very thick] (-6.5, -1.0) -- (-6.5, 1.0);
	\draw[-{latex[length=2mm]}, red, very thick, dashed] (-4.5, -1.0) -- (-4.5, 1.0);
	\draw[-{latex[length=2mm]}, blue, very thick] (-4.5, 1.0) -- (-6.5, 1.0);
	\draw[-{latex[length=2mm]}, blue, very thick, dashed] (-4.5, -1.0) -- (-6.5, -1.0);
	
	\node at (-3.75, 0.0) {$\simeq$};
	
	% Torus - Right
	\draw[red, very thick] (-0.7, -0.26) arc(90:270:-0.14cm and 0.36cm);
	\draw[blue, very thick] (0.5,0.2) arc(0:360:1.5cm and 0.6cm);	
	\draw (1.0,0.0) arc(0:360:2cm and 1cm);
	\draw[rounded corners=28pt] (-2.1,0.1)--(-1.0,-0.6)--(0.1,0.1);
	\draw[rounded corners=24pt] (-1.9,0.0)--(-1.0,0.6)--(-0.1,0.0);

	\end{tikzpicture}
	\end{adjustbox}
	
	\caption{The three different combinations of $[0,1]$ and $S^1$, a closed surface, a cylinder section and a torus. Filled lines are identified with the dashed line of the same color and their equivalent cut is highlighted on the quotient space. Arrows depict orientation.} \label{fig:Combinations}
\end{figure}
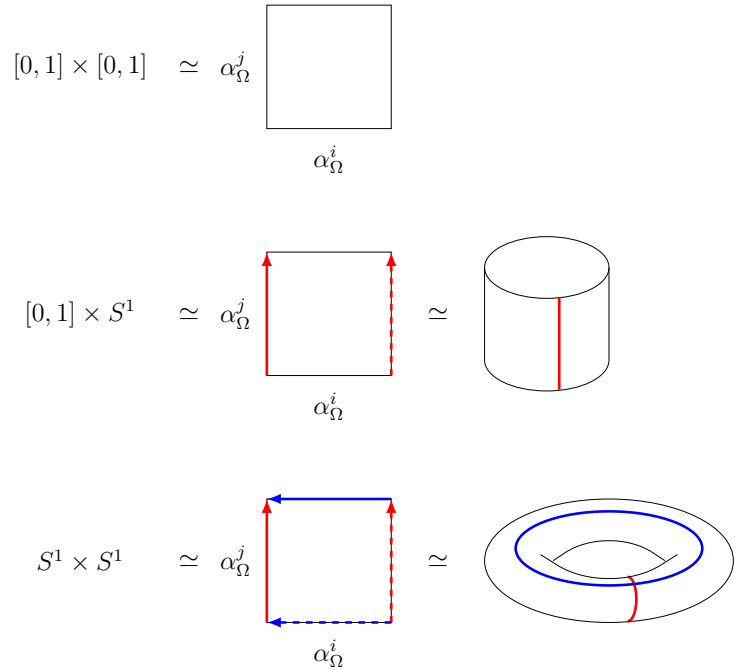

Using this construction of $M_\Omega$, we begin our analysis by looking at pairs of attributes $\alpha^i_\Omega, \alpha^j_\Omega$.  The three possible combinations of $[0,1]$ and $S^1$ result in either a closed surface, a cylinder section or a torus (\cf Figure \ref{fig:Combinations}).

% TODO: Abbildung kleiner machen

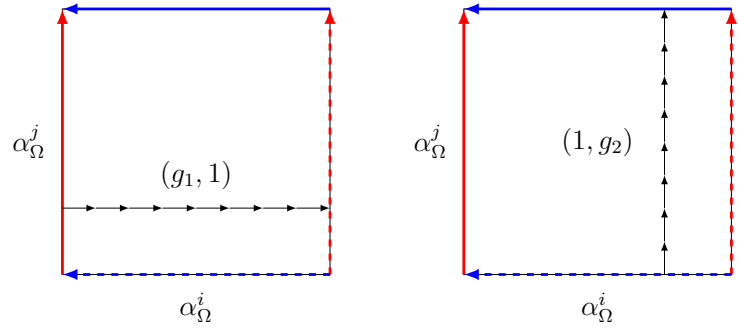
\begin{figure}[H]
	\centering
	\begin{adjustbox}{max width=0.6\textwidth}
	\begin{tikzpicture}
		
	% Torus LEFT
	\node at (-6.5, 2.0) {$\alpha^j_\Omega$};
	\node at (-4.0, -0.5) {$\alpha^i_\Omega$};	
	\draw (-6.0, 0.0) rectangle (-2.0, 4.0);
	\draw[-{latex[length=2mm]}, red, very thick] (-6.0, 0.0) -- (-6.0, 4.0);
	\draw[-{latex[length=2mm]}, red, very thick, dashed] (-2.0, 0.0) -- (-2.0, 4.0);
	\draw[-{latex[length=2mm]}, blue, very thick] (-2.0, 4.0) -- (-6.0, 4.0);
	\draw[-{latex[length=2mm]}, blue, very thick, dashed] (-2.0, 0.0) -- (-6.0, 0.0);

	% Groups
	\node at (-4.0, 1.5) {$(g_1, 1)$};	
	\draw[-{latex[length=1mm]}, black] (-6.0, 1.0) -- (-5.5, 1.0);
	\draw[-{latex[length=1mm]}, black] (-5.5, 1.0) -- (-5.0, 1.0);
	\draw[-{latex[length=1mm]}, black] (-5.0, 1.0) -- (-4.5, 1.0);
	\draw[-{latex[length=1mm]}, black] (-4.5, 1.0) -- (-4.0, 1.0);
	\draw[-{latex[length=1mm]}, black] (-4.0, 1.0) -- (-3.5, 1.0);
	\draw[-{latex[length=1mm]}, black] (-3.5, 1.0) -- (-3.0, 1.0);
	\draw[-{latex[length=1mm]}, black] (-3.0, 1.0) -- (-2.5, 1.0);
	\draw[-{latex[length=1mm]}, black] (-2.5, 1.0) -- (-2.0, 1.0);

	% Torus RIGHT
	\node at (-0.5, 2.0) {$\alpha^j_\Omega$};
	\node at (2.0, -0.5) {$\alpha^i_\Omega$};	
	\draw (0.0, 0.0) rectangle (4.0, 4.0);
	\draw[-{latex[length=2mm]}, red, very thick] (0.0, 0.0) -- (0.0, 4.0);
	\draw[-{latex[length=2mm]}, red, very thick, dashed] (4.0, 0.0) -- (4.0, 4.0);
	\draw[-{latex[length=2mm]}, blue, very thick] (4.0, 4.0) -- (0.0, 4.0);
	\draw[-{latex[length=2mm]}, blue, very thick, dashed] (4.0, 0.0) -- (0.0, 0.0);

	% Groups
	\node at (2.0, 2.0) {$(1, g_2)$};	
	\draw[-{latex[length=1mm]}, black] (3.0, 1.0) -- (3.0, 1.5);
	\draw[-{latex[length=1mm]}, black] (3.0, 1.5) -- (3.0, 2.0);
	\draw[-{latex[length=1mm]}, black] (3.0, 2.0) -- (3.0, 2.5);
	\draw[-{latex[length=1mm]}, black] (3.0, 2.5) -- (3.0, 3.0);
	\draw[-{latex[length=1mm]}, black] (3.0, 3.0) -- (3.0, 3.5);
	\draw[-{latex[length=1mm]}, black] (3.0, 3.5) -- (3.0, 4.0);
	\draw[-{latex[length=1mm]}, black] (3.0, 0.0) -- (3.0, 0.5);
	\draw[-{latex[length=1mm]}, black] (3.0, 0.5) -- (3.0, 1.0);
	
	\end{tikzpicture}
	\end{adjustbox}
	\caption{Visualization of disentanglement of translations, where $G$ is decomposable as $G_1 \times G_2$. $(g_1, 1) \in G_1 \times G_2$ only acts on $\alpha^i_\Omega$ (left side) and $(1, g_2) \in G_1 \times G_2$ only act on $\alpha^j_\Omega$ (right).} \label{fig:NoInterfering}
\end{figure}

To check the disentanglement of $\alpha^i_\Omega$ and $\alpha^j_\Omega$ with respect to a group $G$, we must show that there exists some $G=G_1\times G_2$ under which they transform independently of each other. We take $G$ to be translations in $M_\Omega$ and make the assumption that $G_1$ translates $\alpha^i_\Omega$ and $G_2$ translates $\alpha^j_\Omega$ independently. These translations can be performed explicitly on our gluing instructions in Figure \ref{fig:NoInterfering}, to check whether our assumed behavior of $G_1$ and $G_2$ is indeed correct and that then is disentanglement.

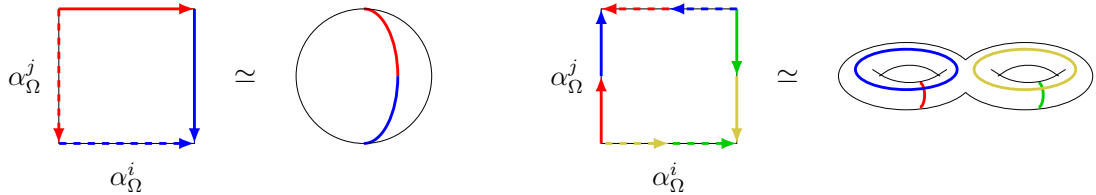
\begin{figure}[H]
	\centering
	\begin{adjustbox}{max width=0.9\textwidth}
	\begin{tikzpicture}

	% Sphere - Left
	\node at (-7.0, 0.0) {$\alpha^j_\Omega$};
	\node at (-5.5, -1.5) {$\alpha^i_\Omega$};
	%\draw[-{Latex[length=2mm]}] (-6.5, 1.0) -- (-6.5, 1.3);
	%\draw[-{Latex[length=2mm]}] (-4.5, -1.0) -- (-4.2, -1.0);
	
	\draw (-6.5, -1.0) rectangle (-4.5, 1.0);
	\draw[{latex[length=2mm]}-, red, very thick, dashed] (-6.5, -1.0) -- (-6.5, 1.0);
	\draw[{latex[length=2mm]}-, blue, very thick] (-4.5, -1.0) -- (-4.5, 1.0);
	\draw[{latex[length=2mm]}-, red, very thick] (-4.5, 1.0) -- (-6.5, 1.0);
	\draw[{latex[length=2mm]}-, blue, very thick, dashed] (-4.5, -1.0) -- (-6.5, -1.0);
	
	\node at (-3.75, -0.0) {$\simeq$};
	
	% Sphere - Right
	\draw[red, very thick] (-2.0,1.0) arc(90:180:-0.5cm and 1.0cm);
	\draw[blue, very thick] (-1.5,0.0) arc(180:270:-0.5cm and 1.0cm);
	\draw (-1.0,0.0) arc(0:360:1.0cm);

	% Gen3 Torus - Left
	\node at (1.0, 0.0) {$\alpha^j_\Omega$};
	\node at (2.5, -1.5) {$\alpha^i_\Omega$};
	%\draw[-{Latex[length=2mm]}] (1.5, 1.0) -- (1.5, 1.3);
	%\draw[-{Latex[length=2mm]}] (3.5, -1.0) -- (3.8, -1.0);
	
	\draw (1.5, -1.0) rectangle (3.5, 1.0);
	\draw[-{latex[length=2mm]}, red, very thick] (1.5, -1.0) -- (1.5, 0.0);
	\draw[{latex[length=2mm]}-, yellow!80!black, very thick] (3.5, -1.0) -- (3.5, 0.0);
	\draw[-{latex[length=2mm]}, blue, very thick] (1.5, 0.0) -- (1.5, 1.0);
	\draw[{latex[length=2mm]}-, green!80!black, very thick] (3.5, 0.0) -- (3.5, 1.0);
	\draw[-{latex[length=2mm]}, blue, very thick, dashed] (3.5, 1.0) -- (2.5, 1.0);
	\draw[{latex[length=2mm]}-, green!80!black, very thick, dashed] (3.5, -1.0) -- (2.5, -1.0);
	\draw[-{latex[length=2mm]}, red, very thick, dashed] (2.5, 1.0) -- (1.5, 1.0);
	\draw[{latex[length=2mm]}-, yellow!80!black, very thick, dashed] (2.5, -1.0) -- (1.5, -1.0);
	
	\node at (4.25, 0.0) {$\simeq$};
	
	% Gen3 Torus - Right
	\draw (8.75,0.0) arc(0:360:1cm and 0.5cm);
	\draw (7.0,0.0) arc(0:360:1cm and 0.5cm);
	
	\draw[draw=white, fill=white] (6.6,-0.23) rectangle (7.25,0.23);
	
	\draw[red, very thick] (6.2, -0.10) arc(90:270:-0.07cm and 0.18cm);
	\draw[blue, very thick] (6.75,0.1) arc(0:360:0.75cm and 0.3cm);	
	\draw[rounded corners=14pt] (5.5,0.1)--(6.05,-0.25)--(6.6,0.1);
	\draw[rounded corners=12pt] (5.6,0.0)--(6.05,0.3)--(6.5,0.0);
	
	\draw[green!80!black, very thick] (7.95, -0.10) arc(90:270:-0.07cm and 0.18cm);
	\draw[yellow!80!black, very thick] (8.5,0.1) arc(0:360:0.75cm and 0.3cm);	
	\draw[rounded corners=14pt] (7.25,0.1)--(7.8,-0.25)--(8.35,0.1);
	\draw[rounded corners=12pt] (7.35,0.0)--(7.8,0.3)--(8.25,0.0);
	
	\end{tikzpicture}
	\end{adjustbox}
	\caption{Gluing instructions on $\alpha^i_\Omega \times \alpha^j_\Omega = [0,1]\times[0,1]$ and their homeomorphic surfaces. The sphere and a torus with genus 2 are shown as examples. Filled lines are identified with the dashed line of the same color and their equivalent cut is highlighted on the resulting quotient space. Arrows depict orientation.} \label{fig:Surfaces}
\end{figure}

Repeating this check for all attributes in $M_\Omega$ gives us a fully disentangled set of attributes. However, in general, $M_\Omega$ is not just a direct product of individual attributes. We continue our analysis by looking at pairs of attributes $\alpha^i_\Omega$ and $\alpha^j_\Omega$ with more complex gluing instructions of the general form $[0,1]^2 / \sim$. Two such examples are given in Figure \ref{fig:Surfaces}, the sphere and the torus with genus $>1$.

\begin{figure}[H]
	\centering
	\begin{adjustbox}{max width=0.28\textwidth}
	\begin{tikzpicture}

	% Torus RIGHT
	\node at (-0.5, 2.0) {$\alpha^j_\Omega$};
	\node at (2.0, -0.5) {$\alpha^i_\Omega$};	
	\draw (0.0, 0.0) rectangle (4.0, 4.0);
	\draw[-{latex[length=2mm]}, red, very thick, dashed] (0.0, 4.0) -- (0.0, 0.0);
	\draw[-{latex[length=2mm]}, blue, very thick] (4.0, 4.0) -- (4.0, 0.0);
	\draw[-{latex[length=2mm]}, red, very thick] (0.0, 4.0) -- (4.0, 4.0);
	\draw[-{latex[length=2mm]}, blue, very thick, dashed] (0.0, 0.0) -- (4.0, 0.0);

	% Groups
	\node at (2.5, 3.0) {$g$};	
	\draw[-{latex[length=1mm]}, black] (3.0, 2.5) -- (3.0, 3.0);
	\draw[-{latex[length=1mm]}, black] (3.0, 3.0) -- (3.0, 3.5);
	\draw[-{latex[length=1mm]}, black] (3.0, 3.5) -- (3.0, 4.0);
	
	\draw[-{latex[length=1mm]}, black] (0.0, 1.0) -- (0.5, 1.0);
	\draw[-{latex[length=1mm]}, black] (0.5, 1.0) -- (1.0, 1.0);
	\draw[-{latex[length=1mm]}, black] (1.0, 1.0) -- (1.5, 1.0);
	
	\end{tikzpicture}
	\end{adjustbox}
	\caption{Visualization of entanglement, where $G$ can't be decomposed and an element $g\in G$ may act on $\alpha^i_\Omega$, as well as $\alpha^j_\Omega$.} \label{fig:Entangled}
\end{figure}
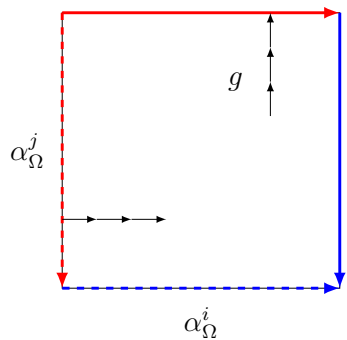

Our assumption that $G$ can be decomposed into $G_1$ and $G_2$ breaks and we quickly find that these transformations no longer act independently on $\alpha^i_\Omega$ and $\alpha^j_\Omega$, as shown in Figure \ref{fig:Entangled} for the sphere.

The failure to decompose $G$ shows that $\alpha^i_\Omega$ and $\alpha^j_\Omega$ are indeed entangled. Using this explicit method, the analysis can be extended to all sets of attributes $[0,1]^l / \sim$. We write $M_\Omega$ in the more general form
\begin{equation}\label{eq:mstructure}
M_\Omega = [0,1]^{n_1} \times \prod_{k=2}^n \left[ \prod_{l=1}^{n_k} \left([0,1]^{l} / \sim_{k,l}\right) \right] \;\;\; ,
\end{equation}
where the total number of attributes is $n=\sum_i n_i$. Now, if the equivalence relations are known exactly, we know which sets of attributes are entangled and which are not. 

However, usually the exact topological structure of $M_\Omega$ as in Equation \ref{eq:mstructure} is not known. Instead, we must probe it to find its structure. As done visually earlier, we examine curves mapped from $\tilde{A}_\lambda$ to $M_\Omega$ and, particularly, their behavior at the most likely identification regions on the boundary of $M_\Omega$, \ie, the values ${0}$ and ${1}$ on each interval. Let $f : [0,1]^n \to M_\Omega$ describe the quotient map for $M_\Omega$, which we are looking for. Here $[0,1]^n$ can be seen as the raw attribute vector $\vec{\alpha}_{\Omega}$ of the capsule, without any topological considerations. We choose some point on the boundary $\partial([0,1]^n)$ and create a curve with, ideally, constant velocity $c_p:[0,2]\to M_\Omega$, so that $f^{-1}(c_p(0)) \notin f(\partial ([0,1]^n))$ and $f^{-1}(c_p(1)) = p$. This $c_p$ describes a curve that crosses the region of identification $\partial ([0,1]^n)$.

Next, we find the two points $p_1 = \lim_{t \to 0} f^{-1}(c_p(1 - t))$ and $p_2 = \lim_{t \to 0} f^{-1}(c_p(1 + t))$. Note that $p_1 \neq p_2$, but $p_1 = p$. The two points form an equivalence relation $p_1 \sim p_2$, since $g(p_1) = g(p_2)$. If we repeat this analysis for every point on $f(\partial ([0,1]^n)) \setminus \partial M_\Omega$, we can reconstruct all the equivalence relations inherent in our quotient space.

\begin{figure}[H]
	\centering
	\begin{tikzpicture}
	
	% Cylinder - Left
	\node at (-9.0, 4.0) {$\alpha^j_\Omega$};
	\node at (-7.5, 2.5) {$\alpha^i_\Omega$};
	
	\draw (-8.5, 3.0) rectangle (-6.5, 5.0);
	
	\draw[blue, thick] (-7.0,3.8) -- (-6.5,3.9);

	% Arrow
	\draw[-{latex[length=2mm]}] (-5.5, 4.0) -- (-4.0, 4.0);
	\node at (-4.75, 3.5) {$g$};

	% Cylinder - Mid
	\draw (-1.0,4.75) arc(0:360:1cm and 0.5cm);
	\draw (-3.0,3.25) arc(180:360:1cm and 0.5cm);
	\draw (-3.0,3.25) -- (-3.0,4.75);
	\draw (-1.0,3.25) -- (-1.0,4.75);
	\draw[red, dashed, thick] (-1.8,2.77) -- (-1.8,4.27);

	\draw[blue, thick] (-2.3,3.4) arc(270:300:1cm and 0.5cm);
	\draw[blue, dashed, thick] (-1.7,3.5) arc(300:330:1cm and 0.5cm);
	
	\draw[red, fill=red] (-1.8,3.475) circle (0.05);

	% Arrow
	\draw[-{latex[length=2mm]}] (0.0, 4.0) -- (1.5, 4.0);
	\node at (0.75, 3.5) {$\gamma$};

	% Cylinder - Left
	\node at (2.5, 4.0) {$\alpha^j_\Omega$};
	\node at (4.0, 2.5) {$\alpha^i_\Omega$};
	
	\draw (3.0, 3.0) rectangle (5.0, 5.0);
	
	\draw[blue, thick] (4.5,3.8) -- (5.0,3.9);
	\draw[blue, thick, dashed] (3.0,3.9) -- (3.5,4.0);
	
	\draw[red, fill=red] (5.0,3.9) circle (0.05);
	\draw[red, fill=red] (3.0,3.9) circle (0.05);
	
	\end{tikzpicture}
	
	\caption{Curve crossing the boundary. $c_p([0,1])$ is depicted as a blue line and  $c_p((1,2])$ as a dashed blue line. The red points describe $p$ and the equivalent points $p_1$ and $p_2$.} \label{fig:Curves}
\end{figure}
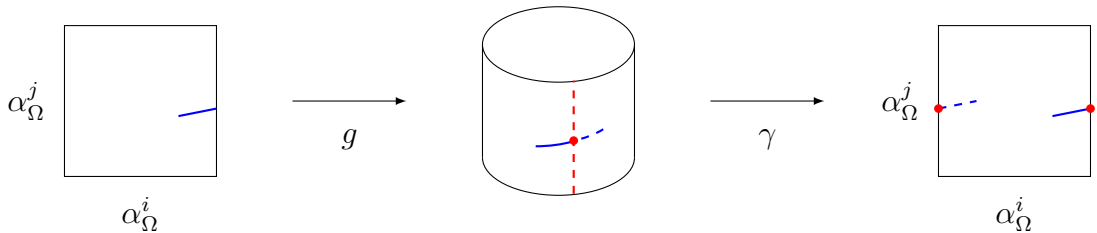

However, we must identify $p_2$ in a practical setting, as $f$ is unknown. We know $c_p$ on the interval $[0,1]$. To find $c_p$ on $(1, 2]$, we push it forward onto $M_\lambda$ as $\tilde{c}_p = g(c_p)$, where we know how the curve evolves on $\tilde{c}_p([0,1]) = g(c_p([0,1]))$. We then use the velocity and acceleration vector of the curve to advance it slightly forward on $M_\lambda$:
\begin{equation}
p_2 = \lim_{\Delta t\to 0} \gamma\left(\tilde{c}_p(1) + \frac{\partial \tilde{c}_p}{\partial t}(1) \cdot \Delta t  + \frac{1}{2}\frac{\partial^2 \tilde{c}_p}{\partial t^2}(1) \cdot \Delta t^2 + \cdots \right) \;\;\;.
\end{equation}
This process is shown in Figure \ref{fig:Curves}.

The above examination gives us a better understanding of the equivalence relations, orientation and entanglement on the boundary of $M_\Omega$ and ideally this would be sufficient. However, this is not always the case. For example, consider again the rotation attribute of a 2D \symb{square}. We saw that such a square is invariant to rotations of $\frac{\pi}{2}$, \ie, it loops back around before reaching the boundary. This means that it is perfectly possible to have equivalence relations inside the domain.  

To find such equivalence relations for the interior of $(0,1)$, we are inclined to follow the same procedure as before and construct and probe multiple geodesics. In practice, we reserve this expensive method using curves for completely unknown cases and perform simpler checks on those that are known, such as the progression of angles $\beta_n = \frac{\pi}{n}$, which we test explicitly for the rotation attribute as ($\pi, \frac{\pi}{2}, \frac{\pi}{3}, \cdots$).

\subsection{Covering the Input Configuration Space}\label{sec:Covering}

With our construction of $M_\Omega$ in Equation \ref{eq:mstructure} and the map $g:M_\Omega \to \tilde{A}_\lambda$, we have limited the possible configuration spaces $\tilde{A}_\lambda$ we are actually able to construct. Our goal is to find an $\tilde{A}_\lambda$ that resembles the true configuration space $A_\lambda$ as closely as possible. However, we have two potential topological problems that need to be addressed:

\begin{enumerate}
	\item $\tilde{A}_\lambda$ is always connected and $A_\lambda$ might not be.
	\item $\tilde{A}_\lambda$ and $A_\lambda$ might not be homeomorphic.
\end{enumerate} 

Each of these differences between $\tilde{A}_\lambda$ and $A_\lambda$ is solvable using one of these two methods:

\begin{enumerate}
	\item We introduce multiple $\tilde{A}_{\lambda, i}$ and patch them together to cover all of $A_\lambda$. For our capsules this is straightforward, as it is nothing more than adding a new route $r$ to the capsule, because each route $r_i$ produces its own $\tilde{A}_{\lambda, r_i}$ using the route specific decoder $g_{r_i}$.
	
	\item We add additional dimensions to $M_\Omega$ by expanding the attribute space, so that $\tilde{A}_\lambda$ itself becomes higher-dimensional and is able to overcome topological shortcomings.
\end{enumerate} 

\begin{figure}[H]
	\centering
	\begin{adjustbox}{max width=0.85\textwidth}
	\begin{tikzpicture}
	
	% backgrounds
	\draw [draw=white, fill=blue!10!white] (-3.75, -2.75) rectangle (-0.25, 3.5);

	\draw [draw=white, fill=black!5!white] (-3.75, -2.75) rectangle (-10.0, 3.5);

	\draw [draw=white, fill=black!5!white] (-0.25, -2.75) rectangle (6.0, 3.5);
	
	%\draw [draw=blue!10!white, fill=blue!10!white] (-3.75, -2.0) rectangle (-0.25, 2.0);
	
	\node at (-3.0, 0.0) {$A_\lambda$};
	\draw[very thick] (-3.0,1.0) -- (-2.0,0.0);
	\draw[very thick] (-3.0,-1.0) -- (-2.0,0.0);
	\draw[very thick] (-2.0,0.0) -- (-0.5,0.0);
	
	%% Left Side

	\draw[-{latex[length=2mm]}] (-4.0, 0.75) -- (-5.5, 1.5);
	\node at (-4.75, 0.6) {$r_1$};

	\node at (-9.5, 1.5) {$\tilde{A}_{\lambda,r_1}$};
	\draw[very thick, dashed] (-8.5,0.5) -- (-7.5,1.5);
	\draw[very thick, dashed] (-7.5,1.5) -- (-6.0,1.5);
	\draw[green, very thick] (-8.5,2.5) -- (-7.5,1.5);
	
	\draw[-{latex[length=2mm]}] (-4.0, -0.75) -- (-5.5, -1.5);
	\node at (-4.75, -1.6) {$r_2$};
	
	\node at (-9.5, -1.5) {$\tilde{A}_{\lambda,r_2}$};
	\draw[green, very thick] (-8.5,-2.5) -- (-7.5,-1.5);
	\draw[very thick, dashed] (-8.5,-0.5) -- (-7.5,-1.5);
	\draw[green, very thick] (-7.5,-1.5) -- (-6.0,-1.5);
	
	\node[align=center] at (-6.875,3.0) {Covering with Routes};
	
	%% Right Side
	
	\draw[-{latex[length=2mm]}] (0.0, 0.0) -- (1.5, 0.0);
	\node at (0.75, -0.3) {$r_1$};

	\node at (5.5, 0.0) {$\tilde{A}_{\lambda,r_1}$};
	\draw[very thick, dashed] (2.0,-1.0) -- (3.0,0.0);
	\draw[very thick, dashed] (3.0,0.0) -- (4.5,0.0);
	\draw[dashed, very thick] (2.0,1.0) -- (3.0,0.0);
	
	\draw[green, very thick] (1.9,0.9) -- (2.8,0.0);
	\draw[green, very thick] (2.1,1.1) -- (3.1,0.1);
	\draw[green, very thick] (2.1,1.1) -- (1.9,0.9);	
	\draw[green, very thick] (1.9,-0.9) -- (2.8,0.0);
	\draw[green, very thick] (2.1,-1.1) -- (3.1,-0.1);
	\draw[green, very thick] (2.1,-1.1) -- (1.9,-0.9);
	
	\draw[green, very thick] (3.1,-0.1) -- (4.5,-0.1);
	\draw[green, very thick] (3.1,0.1) -- (4.5,0.1);
	\draw[green, very thick] (4.5,-0.1) -- (4.5,0.1);
	
	\node[align=center] at (2.875,3.0) {Covering with Attributes};
	
	\end{tikzpicture}
	\end{adjustbox}
	\caption{Example of a problematic configuration space $A_\lambda$ and options to cover it. Left-hand-side shows how $A_\lambda$ is covered by two routes $r_1$ and $r_2$. Right-hand-side presents $A_\lambda$ covered by a higher-dimensional space $\tilde{A}_\lambda$.} \label{fig:Covering}
\end{figure}
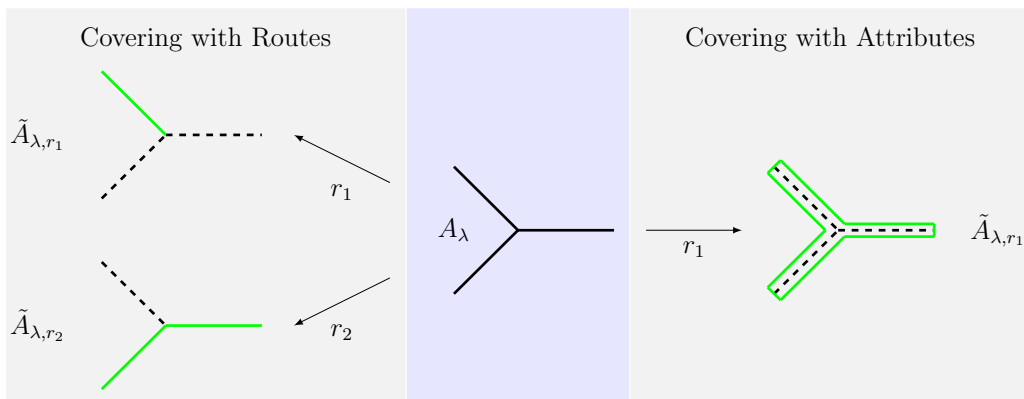

These methods are illustrated in Figure \ref{fig:Covering}. Note, that both methods can be used in any combination to solve the topological issues presented above.

These covering methods, however, can lead to an overparameterization of the capsule and we might end up with redundant attributes. Let $\vec{\alpha}_\Omega\vert_{\alpha^k=[0,1]}$ denote an attribute vector with the $k$-th entry varied in the interval $[0,1]$. By redundant we mean an attribute $\alpha^k_\Omega$ for which there exists another attribute $\alpha^l_\Omega$, such that $g(\vec{\alpha}_\Omega\vert_{\alpha^k=[0,1]})$ and $g(\vec{\alpha}_\Omega\vert_{\alpha^l=[0,1]})$ describe the same subset for any $\vec{\alpha}_\Omega$.

For example, a \symb{door} capsule can have two verb attributes \attr{close} and \attr{slam}. Both trace out the same curve in configuration space, describing the same movement, only at different speeds. Thus, they are numerically redundant attributes, yet encode semantically interesting information. This should motivate us to potentially keep some redundant attributes.

\subsection{Fitting the Input Configuration Space}\label{sec:Fitting}

We assume for now that some decision was made in regard to a remedy to cover $A_\lambda$ with $\tilde{A}_\lambda$. Yet, we are still faced with the problem that we only have a sparse idea of what $A_\lambda$ actually looks like. Our goal is to fit $\tilde{A}_\lambda$ as well to the data and with as much generalization potential as possible as shown in Figure \ref{fig:Covering}. 

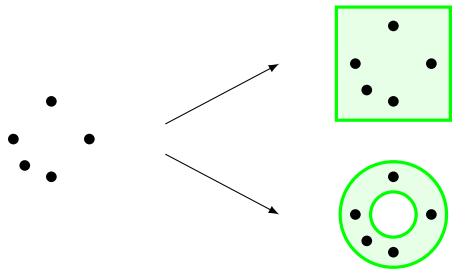
\begin{figure}[H]
	\centering
	\begin{tikzpicture}

	\draw [draw=green,fill=green, fill opacity = 0.1, very thick] (0.75, 0.25) rectangle (2.25, 1.75);
	\path [draw=green, very thick,fill=green, fill opacity = 0.1,even odd rule] (1.5, -1.0) circle (0.7) (1.5, -1.0) circle (0.3);

	\foreach \point in {(-3.0, 0.5),(-3.0,-0.5),(-2.5,-0.0),(-3.5,-0.0),(-3.35,-0.35)}{
		\fill \point circle (2pt);
		\fill \point + (4.5, 1.0) circle (2pt);
		\fill \point + (4.5, -1.0) circle (2pt);
	}

	%% Right Side
	
	\draw[-{latex[length=2mm]}] (-1.5, 0.2) -- (0.0, 1.0);
	\draw[-{latex[length=2mm]}] (-1.5, -0.2) -- (0.0, -1.0);

	\end{tikzpicture}
	
	\caption{A dataset with unknown $A_\lambda$ and two possible coverings with different topology.} \label{fig:ProblemFit}
\end{figure}

To perform the fitting, we employ machine learning techniques, such as deep learning for regression problems. These work perfectly well if we have a lot of data. However, we might be faced with a sparse situation as shown in Figure \ref{fig:ProblemFit}, leading to a lot of ambiguity. Fortunately, this is a common problem in the field of dimensionality reduction and manifold learning \citep{Tenenbaum:2000, Saul:2001, Maaten:2009}.

To reduce ambiguity, we make use of the fact that we are dealing with high-level semantic information at this point, namely attributes with a known interpretation, instead of raw input data. Our approach is to assume some prior configuration space topology for $\tilde{A}_\lambda$ and as more data becomes available, let learning transform this space to slowly converge towards the true configuration space $A_\lambda$.
 
Defining such a prior configuration space is straightforward. We begin with the assumption that most attributes are disentangled, thus $M_\Omega$ has a similar flat topology as presented in Equation \ref{eq:flatConfigSpace} or shown at the top of Figure \ref{fig:ProblemFit}. This allows us to treat groups of attributes of the same type individually. In accordance with this assumption, an augmented training set is created to encode all prior knowledge of $\tilde{A}_\lambda$ discussed here and both the encoder $\gamma$ and decoder $g$ are pre-trained with it. 

For augmentation, our starting point is some observed input configuration, $\vec{\alpha}_{\lambda,0} \in A_\lambda$, which triggered the entire process. From this set of attributes, we must infer the capsule's own attributes $\vec{\alpha}_{\Omega,0}\in M_\Omega$. Generally, we are able to start from any point in $M_\Omega$, but we choose the arithmetic mean of the individual attribute classes of the inputs $\vec{\alpha}_{\lambda,0}$, weighted by their size. We point out that the size is a purely visual measure and does not reflect the actual size of the object. This is done, as it is our best guess, but make exceptions for position, rotation and size. We propose the following equations to calculate the initial  $\vec{\alpha}_{\Omega,0}$:
\begin{equation}\label{eq:Gamma1}
\vec{\alpha}^{\; k}_\Omega = \tilde{\gamma}^{\; k}(\vec{\alpha}_\lambda) = \frac{1}{\sum_\lambda \Vert\vec{\alpha}^{\; size}_\lambda\Vert} \sum_\lambda \vec{\alpha}^{\; k}_\lambda \cdot \Vert\vec{\alpha}^{\; size}_\lambda\Vert   \;\;\; .
\end{equation}
\begin{equation}\label{eq:Gamma2}
\vec{\alpha}^{\; rot}_\Omega = \tilde{\gamma}^{\; rot}(\vec{\alpha}_\lambda) = \frac{1}{\sum_\lambda \Vert\vec{\alpha}^{\; size}_\lambda\Vert} \sum_\lambda \vec{\alpha}^{\; rot}_\lambda \cdot \Vert\vec{\alpha}^{\; size}_\lambda\Vert   \;\;\; .
\end{equation}
\begin{equation}\label{eq:Gamma3}
\begin{split}
\vec{\alpha}^{\; size}_\Omega =  \tilde{\gamma}^{\; size}(\vec{\alpha}_\lambda) =  & \max_{\lambda, i} \left( \textbf{R}^{-1}_\Omega \cdot (\vec{\alpha}^{\; pos}_\lambda +  \textbf{R}_\lambda\cdot \vec{B}_{\lambda,i}) \right)  \\ 
& - \min_{\lambda, i} \left( \textbf{R}^{-1}_\Omega \cdot (\vec{\alpha}^{\; pos}_\lambda +  \textbf{R}_\lambda\cdot \vec{B}_{\lambda,i}) \right)
\end{split}
\end{equation}
\begin{equation}\label{eq:Gamma4}
\begin{split}
\vec{\alpha}^{\; pos}_\Omega =  \tilde{\gamma}^{\; pos}(\vec{\alpha}_\lambda) = \textbf{R}_\Omega \cdot \frac{1}{2}  & \left[ \max_{\lambda, i} \left( \textbf{R}^{-1}_\Omega \cdot (\vec{\alpha}^{\; pos}_\lambda +  \textbf{R}_\lambda\cdot \vec{B}_{\lambda,i}) \right) \right. \\ 
& \left. + \min_{\lambda, i} \left( \textbf{R}^{-1}_\Omega \cdot (\vec{\alpha}^{\; pos}_\lambda +  \textbf{R}_\lambda\cdot \vec{B}_{\lambda,i}) \right) \right]  \;\;\;,
\end{split}
\end{equation}
where $\tilde{\gamma}$ is our prior for $\gamma$, $\textbf{R}_\lambda$ and $\textbf{R}_\Omega$ indicate the Euler rotation matrix calculated from the rotation attributes $\vec{\alpha}_\lambda^{rot}$ and $\vec{\alpha}_\Omega^{rot}$ and $\vec{B}_{\lambda, i}$ the $i$th corner position vector of the bounding box of $\lambda$, \ie, pairwise permutations of $\vec{\alpha}^{\; size}_\lambda / 2$ and $-\vec{\alpha}^{\; size}_\lambda / 2$. Note, that the choice is arbitrary and we are free to use a different set of equations. These equations were chosen to have as few attribute collisions as possible without encoding too much knowledge a priori. A different choice, such as using the centroid of the rendered primitive instead of its bounding box, will lead to a different starting point in $M_\Omega$, but will not impede our overall process. The chosen Equations \ref{eq:Gamma1}-\ref{eq:Gamma4} introduce some entanglement of the position and rotation attributes. 

Now, let $T_i$ indicate a random translation, rotation and scaling of the object that affects $\vec{\alpha}_\lambda^{pos}$, $\vec{\alpha}_\lambda^{rot}$ and $\vec{\alpha}_\lambda^{size}$ only. Further, let $U_i$ indicate a random style transformation that sets all adjective attributes of the same type in $\vec{\alpha}_\lambda$ to the same random value in $[0,1]$, if they are unused (\ie, for all observations in episodic memory $\Lambda_\Omega$ we have $\vec{\alpha}_\lambda^j < \epsilon$). Using these random transformations, the starting point $\vec{\alpha}_{\lambda,0}$ and assumed $\tilde{\gamma}$, we create the first augmented training set as
\begin{equation}\label{eq:AugmentedTrainingSet}
((\tilde{\gamma}(T_i \circ U_i \circ \vec{\alpha}_{\lambda,0}))^{(i)}, (T_i \circ U_i \circ \vec{\alpha}_{\lambda,0})^{(i)}) \;\;\; .
\end{equation}
In essence, we are encoding some of our information about how the world looks by augmentation with $T_i$ and $U_i$. $T_i$ contains our knowledge that an object is still the same type of object, even if it is moved, rotated or at a different size (if there are no reference objects). With $U_i$ we actually encode our lack of knowledge about the possible styles of a new object. $U_i$ assumes that an unused attribute could theoretically be of any value, in some sense mimicking style transfers. An \symb{apple}, for example, should have an adjective attribute \attr{metal} of 0. Through augmenting it with $U_i$, we assume that a metallic apple is possible, as we have never seen it before ($\vec{\alpha}_\lambda^j < \epsilon$). However, as we don't know in what ratio the apple's \symb{stem} and \symb{fruit} are metallic, $U_i$ sets them both to the same value and assumes that this is a metallic apple. We are free to introduce other augmentation strategies apart from $T_i$ and $U_i$, that encode further prior knowledge of the world. Augmentation allows us to start training based on a single example in a one-shot manner.

Once more observations are made and there are more entries $\vec{\alpha}_{\lambda,k}$ in memory, we retrain $\gamma$ using 
\begin{equation}\label{eq:AugmentedTrainingSet2}
((\tilde{\gamma}(T_i \circ U_i \circ \vec{\alpha}_{\lambda,k}))^{(i)}, (T_i \circ U_i \circ \vec{\alpha}_{\lambda,k})))^{(i,k)}) \;\;\; .
\end{equation}
As additional observations with different sets of attributes become available, less and less augmentation via $U_i$ occurs and the resulting $\tilde{A}_\lambda$ begins to loose its disentangled topology induced by Equation \ref{eq:Gamma1} and converges towards a topology that covers its respective part of $A_\lambda$ more closely. However, the augmentation by $T_i$ remains unchanged and we, thus, do not allow that position, rotation and size get entangled with the other attributes, apart from the entanglement introduced in Equations \ref{eq:Gamma2}-\ref{eq:Gamma4}. 

Avoiding the augmentation we introduced here or introducing a different set, the meta-learning algorithm will still eventually cover the observed data with $\tilde{A}_\lambda$. However, this might take longer and more observations are needed. Our choices were made to increase the speed and allow for a better generalization through one-shot learning.

From our discussion thus far, we see that it is crucial to have a suitable distribution of points $\vec{\alpha}_{\lambda,i}$, from which the topological structure of $A_\lambda$ can be inferred. In real environments, where there is no large amount of data with the same object in different styles, this is not always possible, thus, our approximated configuration space $\tilde{A}_\lambda$  represents a best guess until it has encountered a sufficient number of examples.

\subsection{Verb Training}

Thus far, our augmented training set for semantic capsules has ignored verb attributes. While we could simply allow $U_i$ to augment verb attributes as it did the adjective attributes, we propose a better solution that can further increase the autonomy of the capsule network.

We assume that at some point in the life of a capsule it makes a new observation $\vec{\alpha}_{\lambda,l}$ and is prompted to train a verb attribute $\alpha^{\text{new}}$. This indicates, that this observation is the continuation of a previous one, such as $\vec{\alpha}_{\lambda,j}$, but in a different pose. We'll refer to these as the original pose $(t=0, \vec{\alpha}_{\lambda,j})$ and the new pose at $(t=1, \vec{\alpha}_{\lambda,l})$, where $t\in [0,1]$ indicates the change of pose over the entire animation. Now we have a set of $n=2$ key-frames for the animation $\{(t=0, \vec{\alpha}_{\lambda,j}), (t=1, \vec{\alpha}_{\lambda,l})\}$. As new poses for $\alpha^{\text{new}}$ are observed, this list grows and is renormalized back to $t\in [0,1]$. For a given verb attribute, we finally have a sequence of poses
\begin{equation}
\{(t_j=0, \vec{\alpha}_{\lambda,j}), (t_{l_{1}}=1/(n-1), \vec{\alpha}_{\lambda,l_{1}}), (t_{l_{2}}=2/(n-1), \vec{\alpha}_{\lambda,l_{2}}), \cdots, (t_{l_{n-1}}=1, \vec{\alpha}_{\lambda,l_{n-1}})\} \;\;\; ,
\end{equation}
which are stored in the capsule's episodic memory.

At this point we introduce a special verb augmentation transformation $W_i$. If $W_i$ acts on any observation $\vec{\alpha}_{\lambda,j}$ which is an original pose (\ie, $(t=0, \vec{\alpha}_{\lambda,j})$), then $W_i$ is just the identity and performs no transformation for that attribute. If $W_i$ acts on any observation $\vec{\alpha}_{\lambda,l_{i}}$ which is a pose (\ie, $(t_i > 0, \vec{\alpha}_{\lambda,l_{i}})$ with $t>0$), $W_i$ takes the previous pose in the sequence $(t_{i-1}, \vec{\alpha}_{\lambda,l_{i-1}})$, chooses a random $t\in [t_{i-1},t_i]$ and linearly interpolates between $\vec{\alpha}_{\lambda,l_{i-1}}$ and $\vec{\alpha}_{\lambda,l_{i}}$ to produce some new set of attributes $\vec{\alpha}_{\lambda}$ with $\alpha^{\text{new}}=t$. We thus have our new training regime
\begin{equation}\label{eq:AugmentedTrainingSet3}
((\tilde{\gamma}(T_i \circ U_i \circ W_i \circ \vec{\alpha}_{\lambda,k}))^{(i)}, (T_i \circ U_i \circ W_i \circ \vec{\alpha}_{\lambda,k})))^{(i,k)}) \;\;\; .
\end{equation}

\subsection{Expanding Memory}

In the previous section we discussed briefly how to expand memory for verb attributes, by re-normalizing the sequence of poses. We perform a similar resizing of adjective attributes in memory, so that their range always lies in $[0,1]$, as we have up to now always assumed that the entries in memory are perfectly normalized and correctly parameterized.

For this, we note that each time a new observation is made that triggered the meta-learning pipeline, some initial value is provided for the adjectives. However, this initial value, as will become clear in the following sections, is not necessarily correct, but rather a binary "\textit{yes, it has this attribute}" $=1$ or "\textit{no, it does not have this attribute}" $=0$.

The first time we are concerned with an observation $\vec{\alpha}_{\lambda,j}$ with an adjective attribute $\alpha^{\text{adj}}$, where the meta-learning pipeline reported it needs to be trained ($\alpha^{\text{adj}}=1$), we us it as a reference point and keep it at $\alpha^{\text{adj}}_{\lambda,j}=1$. For scaling, we require a second reference point. For this, we can choose any previous observation, such as $\vec{\alpha}_{\lambda,l}$, with $\alpha^{\text{adj}}_{\lambda,l}=0$. 

We can measure the distance between these two reference points as $d_{\text{ref}}=\Vert \vec{\alpha}_{\lambda,j} - \vec{\alpha}_{\lambda,l} \Vert$. Now, each time a new observation $\vec{\alpha}_{\lambda,\text{new}}$ is added to memory, for which the meta-learning pipeline reports that $\alpha^{\text{adj}}$ needs to be trained, we measure its distance $d_{\text{new}}=\Vert \vec{\alpha}_{\lambda,\text{new}} - \vec{\alpha}_{\lambda,l} \Vert$ and set the adjective to $\alpha^{\text{adj}}_{\text{new}} = d_{\text{new}} / d_{\text{ref}}$. If $d_{\text{new}} > d_{\text{ref}}$, then the new attribute value $\vec{\alpha}_{\lambda,\text{new}}$ becomes the new reference value and all adjectives of this type in memory are rescaled  back to $[0,1]$ using $\alpha^{\text{adj}}_{\text{new}}$. Note that this only happens when the meta-learning pipeline reports that this specific attribute needs to be trained.

\subsection{Equivariance}

Our training regime introduces equivariance into $\gamma$ and $g$ for semantic capsules. This helps us to gain insight into how well the attributes parameterize the capsule. Underparameterization means that a capsule is not able to describe every possible configuration, style or pose of its parts ($\gamma$ is not equivariant enough). This is a problem during feed-forward operation as objects might not get detected. Opposed to this, overparameterization means that some attributes have no effect on the capsule's parts ($g$ is not equivariant enough). Here the situation is reversed, the capsule network is able to detect, but in a feed-backward operation, it is not able to correctly generate the described scene. 

Checking equivariance explicitly is costly, as it needs to be performed with respect to some group and finding the correct representation can be difficult. However, we have different measures at our disposal to obtain a good picture of the situation. The overall performance of $\gamma$ and $g$ in the auto-encoder configuration $g \circ \gamma$ is an indication of how well the parameterization works. If the measure
\begin{equation}
Z\left(\tilde{A}_\lambda, (g \circ \gamma)(\tilde{A}_\lambda) \right)
\end{equation}
over the complete $\tilde{A}_\lambda$ is low, it is most likely due to underparameterization. From this we devise another measure that can be used to check more specifically if over- or underparameterization is a likely problem:

\begin{itemize}
	\item $\dim_{CBC} \tilde{A}_\lambda > \dim M_\Omega$ indicates underparameterization.
	
	\item $\dim_{CBC} \tilde{A}_\lambda \approx \dim M_\Omega$ indicates a suitable balance. 
	
	\item $\dim_{CBC} \tilde{A}_\lambda < \dim M_\Omega$ indicates overparameterization.
\end{itemize}

Here $\dim_{CBC}$ means the contracted block-counting dimension (\cf Section 2.1) and $\dim$ the regular dimension. These are only indications, because extreme cases, such as space filling curves, may break this intuition. They are nonetheless useful in making an educated guess and we revisit these measures in our meta-learning pipeline.

\section{Meta-Learning}

In this section we describe the meta-learning agent and how it interacts and gives instructions to the training process discussed in the previous sections. It is far too difficult and too much effort to define the entire grammar with all rules and constraints from scratch to generate a complete capsule network. Instead, our approach is bottom-up and involves only the definition of the terminal symbols (primitive capsules). The meta-learning agent is then responsible for learning all non-terminal symbols (semantic capsules) and the rules (routes) connecting them. Apart from expanding the capsule network, the meta-learning agent is also in charge of the extension of the capsule's attributes. To make sense semantically of what it has learned, an oracle is questioned, the answer of which influences the training process. This is also referred to as active learning.

\subsection{One Observed Axiom}

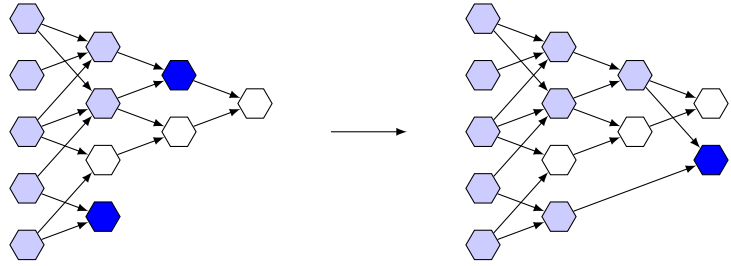
\begin{figure}[H]
	\centering
	\begin{adjustbox}{max width=1.0\textwidth}
		\begin{tikzpicture}
		
		%- First Caps Net
		\node (C01) [regular polygon,regular polygon sides=6, draw=black, fill=blue!20!white] at (0,-1.5) {};
		\node (C02) [regular polygon,regular polygon sides=6, draw=black, fill=blue!20!white] at (0,-0.75) {};
		\node (C03) [regular polygon,regular polygon sides=6, draw=black, fill=blue!20!white] at (0,0) {};
		\node (C04) [regular polygon,regular polygon sides=6, draw=black, fill=blue!20!white] at (0,0.75) {};
		\node (C05) [regular polygon,regular polygon sides=6, draw=black, fill=blue!20!white] at (0,1.5) {};
		
		\node (C11) [regular polygon,regular polygon sides=6, draw=black, fill=blue] at (1,-1.125) {};
		\node (C12) [regular polygon,regular polygon sides=6, draw=black, fill=white] at (1,-0.375) {};
		\node (C13) [regular polygon,regular polygon sides=6, draw=black, fill=blue!20!white] at (1,0.375) {};
		\node (C14) [regular polygon,regular polygon sides=6, draw=black, fill=blue!20!white] at (1,1.125) {};
		
		\node (C22) [regular polygon,regular polygon sides=6, draw=black, fill=white] at (2,0.0) {};
		\node (C23) [regular polygon,regular polygon sides=6, draw=black, fill=blue] at (2,0.75) {};
		
		\node (C31) [regular polygon,regular polygon sides=6, draw=black, fill=white] at (3,0.375) {};
		
		\draw[tochild, black] (C01) -- (C11);
		\draw[tochild, black] (C01) -- (C12);
		\draw[tochild, black] (C02) -- (C11);
		\draw[tochild, black] (C02) -- (C13);
		\draw[tochild, black] (C03) -- (C13);
		\draw[tochild, black] (C03) -- (C14);
		\draw[tochild, black] (C04) -- (C14);
		\draw[tochild, black] (C05) -- (C14);
		\draw[tochild, black] (C05) -- (C13);
		\draw[tochild, black] (C03) -- (C12);
		
		\draw[tochild, black] (C12) -- (C22);
		\draw[tochild, black] (C13) -- (C22);
		\draw[tochild, black] (C13) -- (C23);
		\draw[tochild, black] (C14) -- (C23);
		
		\draw[tochild, black] (C22) -- (C31);
		\draw[tochild, black] (C23) -- (C31);
		
		%- Transform
		
		\draw[tochild, black] (4,0) -- (5,0);
		
		%- New Caps Net
		\node (C01) [regular polygon,regular polygon sides=6, draw=black, fill=blue!20!white] at (6,-1.5) {};
		\node (C02) [regular polygon,regular polygon sides=6, draw=black, fill=blue!20!white] at (6,-0.75) {};
		\node (C03) [regular polygon,regular polygon sides=6, draw=black, fill=blue!20!white] at (6,0) {};
		\node (C04) [regular polygon,regular polygon sides=6, draw=black, fill=blue!20!white] at (6,0.75) {};
		\node (C05) [regular polygon,regular polygon sides=6, draw=black, fill=blue!20!white] at (6,1.5) {};
		
		\node (C11) [regular polygon,regular polygon sides=6, draw=black, fill=blue!20!white] at (7,-1.125) {};
		\node (C12) [regular polygon,regular polygon sides=6, draw=black, fill=white] at (7,-0.375) {};
		\node (C13) [regular polygon,regular polygon sides=6, draw=black, fill=blue!20!white] at (7,0.375) {};
		\node (C14) [regular polygon,regular polygon sides=6, draw=black, fill=blue!20!white] at (7,1.125) {};
		
		\node (C22) [regular polygon,regular polygon sides=6, draw=black, fill=white] at (8,0.0) {};
		\node (C23) [regular polygon,regular polygon sides=6, draw=black, fill=blue!20!white] at (8,0.75) {};
		
		\node (C31) [regular polygon,regular polygon sides=6, draw=black, fill=white] at (9,0.375) {};
		\node (C32) [regular polygon,regular polygon sides=6, draw=black, fill=blue] at (9,-0.375) {};
		
		\draw[tochild, black] (C01) -- (C11);
		\draw[tochild, black] (C01) -- (C12);
		\draw[tochild, black] (C02) -- (C11);
		\draw[tochild, black] (C02) -- (C13);
		\draw[tochild, black] (C03) -- (C13);
		\draw[tochild, black] (C03) -- (C14);
		\draw[tochild, black] (C04) -- (C14);
		\draw[tochild, black] (C05) -- (C14);
		\draw[tochild, black] (C05) -- (C13);
		\draw[tochild, black] (C03) -- (C12);
		
		\draw[tochild, black] (C12) -- (C22);
		\draw[tochild, black] (C13) -- (C22);
		\draw[tochild, black] (C13) -- (C23);
		\draw[tochild, black] (C14) -- (C23);
		
		\draw[tochild, black] (C22) -- (C31);
		\draw[tochild, black] (C23) -- (C31);
		
		\draw[tochild, black] (C11) -- (C32); 
		\draw[tochild, black] (C23) -- (C32);
		\end{tikzpicture}
	\end{adjustbox}
	\caption{A capsule network with all activated capsules in blue (left). Here, the topmost activated capsules (dark blue) do not have a common parent capsule that activated, \ie, a single observed axiom. In this case, the meta-learning agent decided to add a common parent as the new axiom (right).} \label{fig:DanglingCaps}
\end{figure}

Our generative grammar creates an image based on a single input symbol and its attributes. We showed that all images can be generated in such a way (\cf Section 4.2.3) and its inverse also holds true, \ie, each image can be described by a single symbol and its attributes (\cf Section 4.3.3). Thus, if there are multiple activated capsules with no common parent or a parent that did not activate, we interpret this as the observed grammar being incomplete due to the fact that the observed parse-trees don't have a common axiom. Figure \ref{fig:DanglingCaps} gives an example of such a situation and its remedy. 

This does not mean that the capsule network may only have one such axiom overall, but rather, that there must be only one observed axiom for each observation. We will refer to the state of a missing observed axiom as an incomplete observed grammar. Including the case depicted in Figure \ref{fig:DanglingCaps}, there are four causes based on which the agent detects an incomplete observed grammar (we include a verbal description for illustration):

\begin{enumerate}[leftmargin=1.3cm]
	\item[\textbf{A.1}] A non-activated parent capsule is lacking a route. \\
	\textit{"What existing symbol best describes these parts?"}
	\item[\textbf{A.2}] A parent capsule is missing. \\
	\textit{"What new symbol best describes these parts?"}
	\item[\textbf{B.1}] An attribute is lacking training data. \\
	\textit{"What existing attribute best describes this style or pose?"}
	\item[\textbf{B.2}] An attribute is missing. \\
	\textit{"What new attribute best describes this style or pose?"}
\end{enumerate}

Each of these causes can be remedied as indicated in the description, by adding a new route or capsule (\textbf{(A.1)}, \textbf{(A.2)}) or by training an existing or adding a new attribute (\textbf{(B.1)}, \textbf{(B.2)}). As discussed in detail in Sections \ref{sec:Covering} and \ref{sec:Fitting}, these remedies are equivalent to covering and fitting the configuration spaces. Our proposed meta-learning pipeline to rectify this problem consists of the following steps:

\begin{enumerate}
	\item The meta-learning agent detects an incomplete observed grammar.
	\item A weighted list of causes and supplementary information is [presented to an oracle as a question 
	$\vert$ used to make a decision].
	\item The meta-learning agent acts on the decision and extends the capsule network by either:
	\begin{enumerate}
		\item Adding a new route to a capsule with the same attribute setup as the previous routes.
		\item Adding a new capsule and connecting it, making sure it has the same set of attributes as its parts.
		\item Further training of a capsule's route with new training data for a specific attribute.
		\item Adding a new attribute to the capsule and all its ancestors. 
	\end{enumerate}
\end{enumerate}

Once the pipeline finishes, the capsule network may still be in a state of an incomplete observed grammar. To overcome this, the meta-learning process is repeated until there is a single observed axiom for the entire observed scene. 

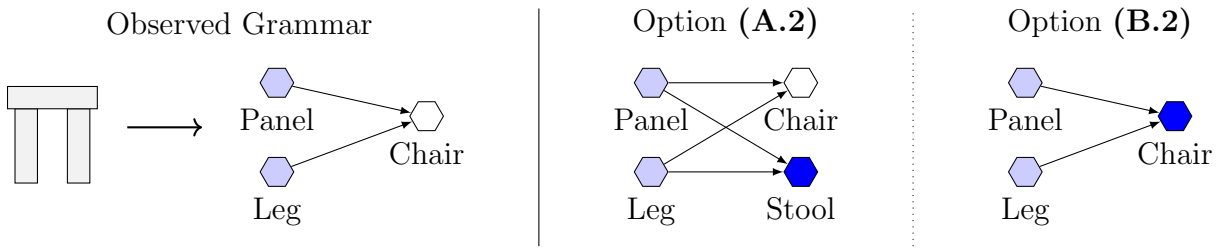
\begin{figure}[H]
	\centering
	\begin{adjustbox}{max width=1.0\textwidth}
		\begin{tikzpicture}
				
		\node (A11) [regular polygon,regular polygon sides=6, draw=black, fill=blue!20!white, label=below:Leg] at (0.0,0.0) {};
		\node (A12) [regular polygon,regular polygon sides=6, draw=black, fill=blue!20!white, label=below:Panel] at (0.0,1.2) {};
		\node (A13) [regular polygon,regular polygon sides=6, draw=black, fill=white, label=below:Chair] at (2.0,0.75) {};
		
		\draw[tochild, black] (A11) -- (A13);
		\draw[tochild, black] (A12) -- (A13);

		\node (A21) [regular polygon,regular polygon sides=6, draw=black, fill=blue!20!white, label=below:Leg] at (5.0,0.0) {};
		\node (A22) [regular polygon,regular polygon sides=6, draw=black, fill=blue!20!white, label=below:Panel] at (5.0,1.2) {};
		\node (A24) [regular polygon,regular polygon sides=6, draw=black, fill=blue, label=below:Stool] at (7.0,0.0) {};
		\node (A23) [regular polygon,regular polygon sides=6, draw=black, fill=white, label=below:Chair] at (7.0,1.2) {};
		
		\draw[tochild, black] (A21) -- (A23);
		\draw[tochild, black] (A22) -- (A23);
		\draw[tochild, black] (A21) -- (A24);
		\draw[tochild, black] (A22) -- (A24);

		\node (A31) [regular polygon,regular polygon sides=6, draw=black, fill=blue!20!white, label=below:Leg] at (10.0,0.0) {};
		\node (A32) [regular polygon,regular polygon sides=6, draw=black, fill=blue!20!white, label=below:Panel] at (10.0,1.2) {};
		\node (A33) [regular polygon,regular polygon sides=6, draw=black, fill=blue, label=below:Chair] at (12.0,0.75) {};

		\draw[tochild, black] (A31) -- (A33);
		\draw[tochild, black] (A32) -- (A33);

		\draw[black, fill=black!05!white] (-3.5,-0.15) rectangle (-3.2,0.85);
		\draw[black, fill=black!05!white] (-2.5,-0.15) rectangle (-2.8,0.85);
		\draw[black, fill=black!05!white] (-3.6,0.850) rectangle (-2.4,1.15);
				
		\draw[->,thick] (-2.0,0.6) -- (-1.0,0.6);
				
		%- Dividers
		\draw [black] (3.5,-1.0) -- (3.5,2.2);
		\draw [black, dotted] (8.5,-1.0) -- (8.5,2.2);
		
		\node at (-0.5, 2.0) {Observed Grammar};
		\node at (6.0, 2.0) {Option \textbf{(A.2)}};
		\node at (11.0, 2.0) {Option \textbf{(B.2)}};
				
		\end{tikzpicture}
	\end{adjustbox}
	\caption{The observed grammar (left) is incomplete, because both the \symb{panel} and \symb{leg} capsule activated (light blue), but no common parent capsule activated (blue). Two possible remedies are presented, option \textbf{(A.2)} is to add a new parent capsule \symb{stool} and option \textbf{(B.2)} is to add a new attribute.} \label{fig:StoolChair}
\end{figure}

The decision of which remedy to choose is either performed by an oracle or, if available, past oracle decisions are used to automatically come to a conclusion. Thus, the decision is subjective. As an example, consider the situation presented in Figure \ref{fig:StoolChair}: a capsule network with \symb{leg}, \symb{panel} and \symb{chair} capsules and route  \symb{leg}\symb{leg}\symb{panel} $\to$ \symb{chair} makes a new observation triggering the meta-learning agent. It observed a scene of \symb{leg}\symb{leg}\symb{panel}, which did not activate the \symb{chair} capsule. The oracle now must decide, if this is a \symb{chair} of unknown style \textbf{(B.2)} or if it should be described by a different type of symbol, such as \symb{stool} or \symb{table} \textbf{(A.2)}. Also \textbf{(A.1)} and \textbf{(B.1)} may be possibilities that need to be considered.

\subsection{The Decision Process}

Each time the capsule network is confronted with an unknown situation (\ie, multiple top-most capsules that activated competing to be the observed axiom), it has to make a decision what training process it needs to start (\textbf{(A.1)} - \textbf{(B.2)}), by inquiring an oracle. In order to make a decision, the oracle is given specific features of the current situation. Assuming the oracle is a human, it is difficult to identify what these features are, as they contain meta-information about the observed state not encoded in the attributes alone. To capture as much information about these observations as possible, we propose the features in Table \ref{tab:FeatureTable} with a short description of their interpretation.

\begin{table}[H]
	\begin{center}
		\renewcommand{\arraystretch}{1.2}
		\begin{tabularx}{\linewidth}{
				|>{\hsize=0.02\hsize}X|>{\hsize=0.49\hsize}X|>{\hsize=0.49\hsize}X|
			}
			\hline
			\# & Feature & Description \\
			\hline\hline			
			
			1 &
			Top-most capsules share an unactivated $\Omega$ as parent. & 
			The parts in the scene seem to belong to the same object $\Omega$, but are in an unknown configuration. \\ 
			\hline
			
			2 &
			Top-most capsules don't share an unactivated $\Omega$ as parent. & 
			The parts in the scene seem to belong to different objects. \\ 
			\hline
			
			3 &
			Parts tracked from previous scenes. & 
			All parts were in the previous frame and should still belong to the same object. \\ 
			\hline
			
			4 &
			$\Omega: Z(\vec{\alpha}, \vec{\tilde{\alpha}})$ indicates one attribute mismatch with no entry in memory $\alpha^i >\epsilon$.  & 
			It looks like the parts belong to object $\Omega$, but with an unexpected use of the $i$th adjective. \\ 
			\hline
			
			5 &
			$\Omega: Z(\vec{\alpha}, \vec{\tilde{\alpha}})$ indicates attribute mismatch for (position, rotation, size) only. & 
			It looks like the parts belong to object $\Omega$, but with an unexpected pose. \\ 
			\hline			
			
			6 &
			$\Omega: Z(\vec{\alpha}, \vec{\tilde{\alpha}})$ indicates attribute mismatch for more than half of all attributes. & 
			It looks like that the parts belong to object $\Omega$, but with very different attributes. \\ 
			\hline
			
			7 &
			$\dim_{CBC} \tilde{A}_\lambda > \dim M_\Omega$ indicating under-parameterization in $\Omega$. & 
			The object $\Omega$ is not sufficiently described by its current set of attributes.  \\ 
			\hline	
			
			8 &
			$\dim_{CBC} \tilde{A}_\lambda \approx \dim M_\Omega$ indicating a suitable balance in $\Omega$. & 
			The object $\Omega$ is suitably described by its current set of attributes.  \\ 
			\hline	
			
			9 &
			$\dim_{CBC} \tilde{A}_\lambda < \dim M_\Omega$ indicating over-parameterization in $\Omega$. & 
			The object $\Omega$ has more attributes than needed.  \\ 
			\hline

			 &
			\dots & 
			\dots  \\ 
			
			\hline
			
		\end{tabularx}
	\end{center}
	\caption{\label{tab:FeatureTable} Possible features that an oracle might base its decision on.}
\end{table}

Using these features and the semantic interpretation of the attributes (adjectives, verbs and prepositions) and capsules (nouns), we are able to formulate actual questions for the oracle. Consider our previous example with two \symb{leg} and one \symb{panel} capsules activating, but not \symb{chair}. Instead of presenting the oracle with this raw data, we may ask it based on Feature \ref{tab:FeatureTable} \textit{"Do these two \symb{leg}s and one \symb{panel} form a \symb{chair}?"} and highlight the area in the image using the grammar. The answer then corresponds to one of the four training regimes:

\begin{enumerate}[leftmargin=1.3cm]
	\item[\textbf{A.1}] A non-activated parent capsule is lacking a route. \\
	\textit{"Yes, this is \symb{chair}."}
	\item[\textbf{A.2}] A parent capsule is missing. \\
	\textit{"No, this is a \symb{stool}."}
	\item[\textbf{B.1}] An existing attribute is lacking training data. \\
	\textit{"Yes, this is a \attr{wooden} \symb{chair}."}
	\item[\textbf{B.2}] An attribute is missing. \\
	\textit{"Yes, this is a \attr{modern} \symb{chair}."}
\end{enumerate}

To avoid inquiring an oracle each time, we can use these features and the past decisions of the oracle as a data set and train a classification model, such as the simple decision matrix shown in Table \ref{tab:decisionMatrix}. By summing all the rows whose feature is true, we find the predicted decision (\textbf{(A.1)} - \textbf{(B.2)}) from the column with the highest value. Should the oracle make new decisions, the matrix is updated by adding $1$ to the respective columns of all rows that were true.

\begin{table}[H]
	\begin{center}
		\begin{tabular}{|| c | c | c | c | c ||} 
			\hline
			Feature & A.1 & A.2 & B.1 & B.2 \\
			\hline\hline			
			
			\makecell{Top-most capsules share an unactivated $\Omega$ as parent.} & \cellcolor{yellow!10!white} 4 & \cellcolor{yellow!10!white} 3 & \cellcolor{green!10!white} 14 & \cellcolor{green!10!white} 12  \\ 
			\hline
			
			\makecell{Top-most capsules don't share an unactivated $\Omega$ as parent.} & \cellcolor{yellow!10!white} 5 & \cellcolor{green!10!white} 19 & \cellcolor{red!10!white} 1 & \cellcolor{red!10!white} 0  \\ 
			\hline
			
			\makecell{Parts tracked from previous scenes.} & \cellcolor{green!10!white} 14 & \cellcolor{red!10!white} 1 & \cellcolor{green!10!white} 17 & \cellcolor{green!10!white} 12 \\ 
			\hline
			
			\makecell{$\Omega: Z(\vec{\alpha}, \vec{\tilde{\alpha}})$ indicates one attribute mismatch \\ with no entry in memory $\alpha^i >\epsilon$. } & \cellcolor{red!10!white} 1 & \cellcolor{red!10!white} 0 & \cellcolor{green!10!white} 12 & \cellcolor{red!10!white} 2  \\ 
			\hline
			
			\makecell{$\Omega: Z(\vec{\alpha}, \vec{\tilde{\alpha}})$ indicates attribute mismatch \\ for (position, rotation, size) only.} & \cellcolor{yellow!10!white} 4 & \cellcolor{yellow!10!white} 3 & \cellcolor{green!10!white} 13 & \cellcolor{green!10!white} 10  \\ 
			\hline			
			
			\makecell{$\Omega: Z(\vec{\alpha}, \vec{\tilde{\alpha}})$ indicates attribute mismatch \\ for more than half of all attributes.} & \cellcolor{green!10!white} 12 & \cellcolor{green!10!white} 14 & \cellcolor{yellow!10!white} 4 & \cellcolor{yellow!10!white} 4  \\ 
			\hline	
			
			\makecell{$\dim_{CBC} \tilde{A}_\lambda > \dim M_\Omega$ indicating under-parameterization in $\Omega$.} & \cellcolor{red!10!white} 1 & \cellcolor{red!10!white} 2 & \cellcolor{yellow!10!white} 5 & \cellcolor{green!10!white} 12  \\ 
			\hline	
			
			\makecell{$\dim_{CBC} \tilde{A}_\lambda \approx \dim M_\Omega$ indicating a suitable balance in $\Omega$.} & \cellcolor{green!10!white} 12 & \cellcolor{yellow!10!white} 5 & \cellcolor{green!10!white} 10 & \cellcolor{yellow!10!white} 6  \\ 
			\hline	
			
			\makecell{$\dim_{CBC} \tilde{A}_\lambda < \dim M_\Omega$ indicating over-parameterization in $\Omega$.} & \cellcolor{yellow!10!white} 5 & \cellcolor{green!10!white} 14 & \cellcolor{green!10!white} 12 & \cellcolor{yellow!10!white} 7  \\ 
			\hline

			\makecell{$\cdots$} &  &  &  &   \\ 
			\hline

			\hline
			
		\end{tabular}
	\end{center}
	\caption{\label{tab:decisionMatrix} Example of a trained decision matrix with an excerpt of features derived from the observed parse-trees and what cause they indicate (number of past oracle decisions). Here $\Omega$ is the capsule with the highest $p_\Omega$ that did not activate.}
\end{table}

The overall process is comparable to how an infant learns about the world, by first inquiring its parent and slowly learning the subjective way they classify and describe new objects.

\section{Implementation and Results}

For an example, we start with a capsule network that consists of three different primitive capsules, \symb{square}, \symb{triangle} and \symb{circle}. Each of them is implemented using signed distance fields \citep{Hart:1989, Hart:1993, Quilez:2017} as their renderer $g$. For the encoder $\gamma$ we initially employed using neural architecture search, however realized that an AlexNet-like CNN \citep{Krizhevsky:2012} for regression works well enough as encoder $\gamma$. 

The semantic capsules are implemented using a 4-layer deep dense neural network for regression with $\tanh$ activation functions as their encoder $\gamma$ and decoder $g$, where the width is dependent on the number of attributes. During training, we employ dropout \citep{Hinton:2012b} on individual input capsules to simulate occlusions. The full implementation is available as a Github repository at \url{https://github.com/Kayzaks/VividNet}.

To test our meta-learning implementation, we use a toy example based on an environment visually similar to Asteroids (Atari 2600). By presenting the capsule network with a scene, the meta-learning pipeline is immediately triggered, as a lot of \symb{square}, \symb{triangle} and \symb{circle} symbols are detected, but without a common parent as their observed axiom.

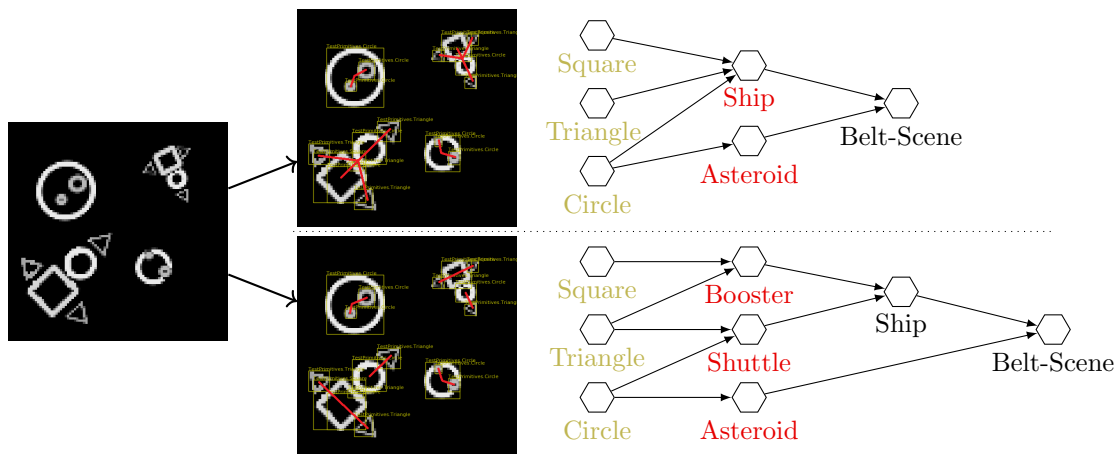
\begin{figure}[H]
	\centering
	\begin{adjustbox}{max width=1.0\textwidth}
		\begin{tikzpicture}
		\node[inner sep=0pt] (A0) at (-0.3,-0.2) {\includegraphics[width=.18\textwidth]{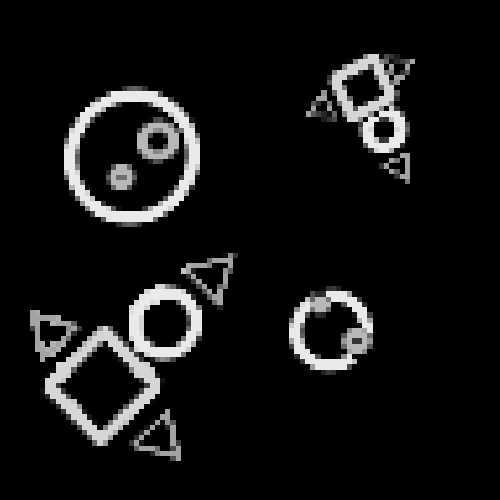}};
		\node[inner sep=0pt] (A2) at (3.5,1.3) {\includegraphics[width=.18\textwidth]{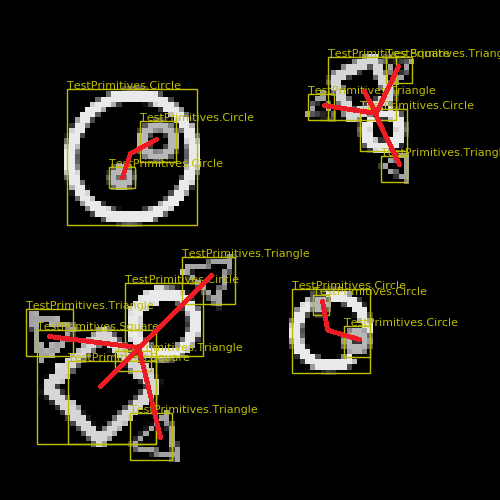}};
		\node[inner sep=0pt] (A3) at (3.5,-1.7) {\includegraphics[width=.18\textwidth]{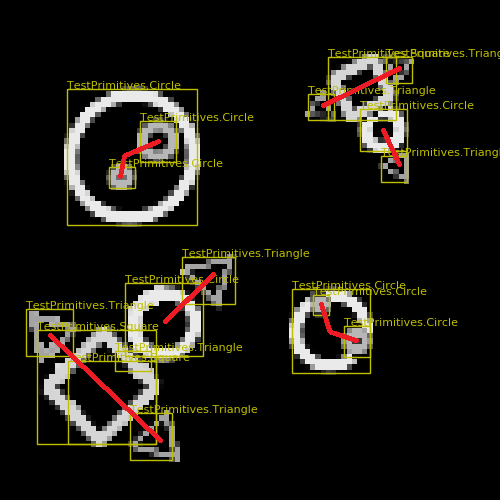}};
		
		\draw[->,thick] (A0) -- (A2);
		\draw[->,thick] (A0) -- (A3);
		
		%- Top Caps Net
		%- Square, Triangle, Circle
		\node (C01) [label={[shift={(0.0,-0.9)}]\textcolor{black!30!yellow}{\footnotesize{Square}}}, regular polygon,regular polygon sides=6, draw=black, fill=white] at (6,2.4) {};
		\node (C02) [label={[shift={(-0.05,-0.9)}]\textcolor{black!30!yellow}{\footnotesize{Triangle}}}, regular polygon,regular polygon sides=6, draw=black, fill=white] at (6,1.5) {};
		\node (C03) [label={[shift={(0.0,-0.9)}]\textcolor{black!30!yellow}{\footnotesize{Circle}}}, regular polygon,regular polygon sides=6, draw=black, fill=white] at (6,0.6) {};
		
		\node (C11) [label={[shift={(0.0,-0.95)}]\textcolor{black!10!red}{\footnotesize{Ship}}}, regular polygon,regular polygon sides=6, draw=black, fill=white] at (8,2.0) {};
		\node (C12) [label={[shift={(0.0,-0.9)}]\textcolor{black!10!red}{\footnotesize{Asteroid}}}, regular polygon,regular polygon sides=6, draw=black, fill=white] at (8,1.0) {};
		
		\node (C21) [label={[shift={(0.0,-0.9)}]\textcolor{black!30!black}{\footnotesize{Belt-Scene}}}, regular polygon,regular polygon sides=6, draw=black, fill=white] at (10,1.5) {};
		
		\draw[tochild, black] (C01) -- (C11);
		\draw[tochild, black] (C02) -- (C11);
		\draw[tochild, black] (C03) -- (C11);
		\draw[tochild, black] (C03) -- (C12);
		
		\draw[tochild, black] (C11) -- (C21);
		\draw[tochild, black] (C12) -- (C21);
		
		%- Divider
		\draw [dotted] (2,-0.2) -- (12,-0.2);
		
		%- Bottom Caps Net
		%- Square, Triangle, Circle
		\node (D01) [label={[shift={(0.0,-0.9)}]\textcolor{black!30!yellow}{\footnotesize{Square}}}, regular polygon,regular polygon sides=6, draw=black, fill=white] at (6,-0.6) {};
		\node (D02) [label={[shift={(0.0,-0.9)}]\textcolor{black!30!yellow}{\footnotesize{Triangle}}}, regular polygon,regular polygon sides=6, draw=black, fill=white] at (6,-1.5) {};
		\node (D03) [label={[shift={(0.0,-0.9)}]\textcolor{black!30!yellow}{\footnotesize{Circle}}}, regular polygon,regular polygon sides=6, draw=black, fill=white] at (6,-2.4) {};
		
		%- Rocket, Shuttle, Asteroid
		\node (D11) [label={[shift={(0.0,-0.9)}]\textcolor{black!10!red}{\footnotesize{Booster}}}, regular polygon,regular polygon sides=6, draw=black, fill=white] at (8,-0.6) {};
		\node (D12) [label={[shift={(0.0,-0.9)}]\textcolor{black!10!red}{\footnotesize{Shuttle}}}, regular polygon,regular polygon sides=6, draw=black, fill=white] at (8,-1.5) {};
		\node (D13) [label={[shift={(0.0,-0.9)}]\textcolor{black!10!red}{\footnotesize{Asteroid}}}, regular polygon,regular polygon sides=6, draw=black, fill=white] at (8,-2.4) {};
		
		\node (D21) [label={[shift={(0.0,-0.95)}]\textcolor{black!30!black}{\footnotesize{Ship}}}, regular polygon,regular polygon sides=6, draw=black, fill=white] at (10,-1.0) {};
		
		\node (D31) [label={[shift={(0.0,-0.9)}]\textcolor{black!30!black}{\footnotesize{Belt-Scene}}}, regular polygon,regular polygon sides=6, draw=black, fill=white] at (12,-1.5) {};
		
		\draw[tochild, black] (D01) -- (D11);
		\draw[tochild, black] (D02) -- (D11);
		\draw[tochild, black] (D02) -- (D12);
		\draw[tochild, black] (D03) -- (D12);
		\draw[tochild, black] (D03) -- (D13);
		
		\draw[tochild, black] (D11) -- (D21);
		\draw[tochild, black] (D12) -- (D21);
		
		\draw[tochild, black] (D21) -- (D31);
		\draw[tochild, black] (D13) -- (D31);
		
		\end{tikzpicture}
	\end{adjustbox}
	\caption{Two of many possible capsule network configurations the meta-learning agent might end up with, depending on the oracle and the decision matrix.} \label{fig:ResultsMeta}
\end{figure}

The oracle is queried to resolve these issues. However, depending on the subjective answer the oracle gives, the capsule network ends up with different configurations. In Figure \ref{fig:ResultsMeta} we present two such networks reached in one-shot. Note that the information content of each is slightly different, yet they are both able to correctly identify all objects in the scene as well as their relations.

\clearpage

\chapter{Intuitive Physics}\label{sec:Physics}\thispagestyle{empty}

\begin{wrapfigure}{R}{0.4\textwidth}
	\centering
	\begin{adjustbox}{max width=0.4\textwidth}
		\begin{tikzpicture}
		
		\vividnet{black!20!white}{black!20!white}{black!20!white}{black!10!blue}{black!20!white}{black!20!white}{black!20!white}

		\end{tikzpicture}
	\end{adjustbox}
\end{wrapfigure}

One of the advantages of a scene-graph representation of the visual semantics is that it is very versatile structure which allows for further exploration. In this chapter we discuss how to extract physical properties from an observation and how to predict future events based on this knowledge\footnotemark. These processes of intuitive physics are a vital component to achieve our overall goal of mental simulation. 

We approach the problem of predicting physics in three steps. First, we analyze the observations found in episodic memory to determine what objects in the scene are interactable. For this set of objects, we determine what kind of interaction they allow. This comprises, for example, is the object elastic or rigid or is the object connected to another by a joint? Finally, these properties allow us to predict future events using interaction networks.

\footnotetext{A preprint of the results in this chapter was previously uploaded to arXiv \citep{Kissner2:2019}.}
\clearpage

\section{Interaction Networks}

Interaction networks were designed to learn and predict the relations and physical behavior of interacting bodies \citep{Battaglia:2016}. The overall idea is to learn this interaction for pairs of objects and to aggregate the results to determine the final effect these interactions have on an object.

To model an interaction pair, one object $o_i$ is labeled as the \textit{sender} and the other object $o_j$ as the \textit{receiver} of some physical effect $e_{k}$. To calculate this effect, more information about the interaction is required, such as all possible constraints between sender and receiver, which we encode in $\rho$. This gives us an interaction triplet $\langle o_i, o_j, \rho \rangle_k$. The resulting physical effect $e_{k}$ of this interaction is calculated using a \textit{relational model}:
\begin{equation}
e_{k} = \phi_R \lbrack \langle o_i, o_j, \rho \rangle_k \rbrack \;\;\; .
\end{equation}

In order to make a prediction for the next time-step of a receiver $o_j$, all interactions with all possible senders $\{ o_i \}$ must be calculated to find the full set of effects $\{e_k\}$. Not all physical effects, however, arise from pairwise interactions. We must, therefore, include external effects $x_j$ that may apply to $o_j$, such as gravity. 

To aggregate all interaction effects that apply to the receiver, an aggregation function $a(\cdot)$ is introduced. In general, a linear combination of all effects $\{e_k\}$ is sufficient. To apply these effects to the $o_j$ in a meaningful way, an \textit{object model} is introduced as
\begin{equation}\label{eq:ExtForces}
o_j^{t+1} = \phi_O \lbrack o_j, a(\{e_k\}), x_j \rbrack \;\;\; ,
\end{equation}
which calculates the receiver's attributes at the next time-step.

Both relational models $\phi_R$ and $\phi_O$ are the central part of interaction networks and are implemented using regression models trained on previously observed interactions. This allows us to both learn and predict intuitive physics. 

\section{Interactable Neural-Symbolic Capsules}

Our observations in episodic memory are ideal candidates for interaction networks, as each object is a discrete entity and the most relevant physical properties are either known (position, rotation, \dots), can be inferred explicitly (speed, \dots) or estimated implicitly (mass, \dots). In the following, we explore how to integrate interaction networks with our neural-symbolic capsule network.

For the physical interactions which we want to extract from the visual data, the distance and the orientation of the surfaces are important. While these quantities are, in some sense, encoded in the attributes, such as \attr{size} and \attr{position}, and through their relation to other objects in the observed parse-tree, we extract this information explicitly using the object's visual representation, as:

\begin{enumerate}
	\item Render objects $A$ and $B$.
	\item Find the minimal distance between the pixels (2D) or voxels (3D) of the two renderings using
	\begin{equation}\label{eq:continue3}
	d_{A,B} = \inf\left\{ d(\vec{a}_i^{pos} - \vec{b}_j^{pos}) \;\lvert \; \vec{a}_i \in A,\; \vec{b}_j \in B; \;  a_i^{int} > 0,\; b_j^{int} > 0 \right\} \;\;\; ,
	\end{equation}
	where $\vec{a}, \vec{b}$ are the attributes of individual pixels or voxels, $pos$ and $int$ refer to their positions and intensities and $A,B$ on the sets of all pixels or voxels for each object. 
	\item Calculate the boundary normals $\vec{N}_A$ and $\vec{N}_B$ for the points of minimal distance between $A$ and $B$.
\end{enumerate}

\section{Interactable Objects}

Some features we require for the interaction networks must be extracted from past observation sequences and can't be inferred from a single image without prior information. We need to know about the object's ability to deform and move, in order to make accurate physical predictions.

For this analysis we need to identify those objects in the scene that are either non-interactable, independently interactable or constrained by a joint. For example, while a standard \symb{chair} is made up of multiple parts, we only consider the \symb{chair} as a whole interactable, as moving one \symb{leg} causes the \symb{chair} in its entirety to move. However, for an \symb{office-chair} the \symb{chair-base} and the \symb{seat} are connected by a joint. Here, we consider \symb{chair-base} and \symb{seat} as interactable, because they have a degree-of-freedom (DOF) by which they can move independently of each other (namely rotation), but consider \symb{office-chair} not to be interactable, as its interactions are fully determined by the two interactable parts. We see that the presence of a verb attribute, such as \attr{swivel}, is the main factor that determines this joint/interactable relationship.

To find the interactable objects and the DOF of their movement, we propose the following steps (note that we assume $g$ is continuous, as is the case for regression models based on neural networks):

\begin{enumerate}
	\item Find a semantic capsule $\Omega$ that has verb attributes $\vec{\alpha}_\Omega^{k_1,\cdots,k_n}$ and one or more child capsules $\lambda_i$ that do not have verb attributes. All child capsules $\lambda_i$ that satisfy this condition form the set of interactable objects $O$.
	
	\item Vary $\vec{\alpha}_\Omega^{k_1,\cdots,k_n}$ in the range $[0,1]^n$ using $g(\vec{\alpha}_\Omega) = \vec{\alpha}_{1,\cdots,\vert\lambda\vert}$ to find the spaces for scale, pose (position and rotation) and collision
	\begin{equation}
	{\mathbf{S}}_i = \{ \vec{\alpha}_i^{\mathit{size}}\} \;\;\; ,
	\end{equation}
	\begin{equation}
	{\mathbf{Q}}_i = \{ \vec{\alpha}_i^{\mathit{pos}} \oplus \vec{\alpha}_i^{\mathit{rot}} \} \;\;\; ,
	\end{equation}
	\begin{equation}
	{\mathbf{Q}}_{ij} = \{ (\textbf{R}_i \cdot (\vec{\alpha}_j^{\mathit{pos}} - \vec{\alpha}_i^{\mathit{pos}})) \oplus (\vec{\alpha}_j^{\mathit{rot}} - \vec{\alpha}_i^{\mathit{rot}}) \}  \;\;\; ,
	\end{equation}
	\begin{equation}
	\mathbf{D}_{ij} = \{ d(\lambda_i, \lambda_j) \}  \;\;\; ,
	\end{equation}
	where $\textbf{R}_i$ is the Euler rotation matrix for $-\vec{\alpha}_i^{\mathit{rot}}$, which is used to move the two objects into the same reference frame. ${\mathbf{Q}}_{ij}$ is also defined for two capsules which do not share a common parent, by instead finding the next common ancestor and successively applying $g$ to vary over $[0,1]^n$.
	
	\item For the spaces ${\mathbf{S}}_i$, ${\mathbf{Q}}_i$ and ${\mathbf{Q}}_{ij}$ we find the contracted block-counting dimensions $\dim_{CBC} {S_i}$, $\dim_{CBC} {Q_i}$ and $\dim_{CBC} {Q_{ij}}$. 	
		
\end{enumerate}

In step 1, we are essentially splitting the objects in an observation into two groups. The first group consists of all capsules whose parts have been found to take on different poses described by a verb. An example is \symb{laptop}, where the \symb{screen} and the \symb{keyboard} poses are dependent on some \attr{open} verb attribute. The second group consists of all capsules whose parts do not take on individual poses, such as a laptop's \symb{screen} which itself is made out of parts, but their pose is fully described by the screen's pose and, thus, the \symb{screen} does not have a verb attribute.

Following the identification of all the interactable objects, we explore how they move in relation to each other in step 2 and explicitly compute the DOF in step 3. From the derived quantities, we are able to infer some of the physical properties of the objects, such as:

\begin{enumerate}
	\item[A] A verb-less child $\lambda_i$ is a \textbf{rigid body} if $\dim_{CBC} {S_i} = 0$.\\
	\textit{$\to$ The size has never been observed to change over time.}
	\item[B] A verb-less child $\lambda_i$ is an \textbf{elastic/plastic body} if $\dim_{CBC} {S_i} > 0$.\\
	\textit{$\to$ The size has been observed to change over time.}
	\item[C] A verb-less child $\lambda_i$ is \textbf{static} if $\dim_{CBC} {Q_i} = 0$.\\
	\textit{$\to$ The position or rotation has never been observed to change over time.}
	\item[D] A verb-less child $\lambda_i$ is \textbf{dynamic} if $\dim_{CBC} {Q_i} > 0$.\\
	\textit{$\to$ The position or rotation has been observed to change over time.}
	\item[E] Two verb-less descendants $\lambda_i$ and $\lambda_j$ are connected by a \textbf{joint} if they collide ($\sup\mathbf{D}_{ij} < \epsilon$) and their total DOF $\dim_{CBC} {Q_{ij}}$ in relation to each other is $3 > \dim_{CBC} {Q_{ij}} > 0$ (2D) or $6 > \dim_{CBC} {Q_{ij}} > 0$ (3D).\\
	\textit{$\to$ The two objects have been observed to move in relation to each other over time in a constrained way.}
\end{enumerate}

Consider again our earlier example of the \symb{office-chair} with the \attr{swivel} verb attribute. Its parts, \symb{base} and \symb{seat}, do not have this verb and are, according to step (1.) of our algorithm, interactable, whereas their parent \symb{office-chair} is not. Now, by further analyzing the properties of \symb{base} and \symb{seat} using steps (2.) and (3.), we find that they are connected by a joint with one DOF ($\dim_{CBC} {Q_{\mathit{base},\mathit{seat}}} = 1$). Further, both of these objects are dynamic ($\dim_{CBC} {Q_{\mathit{base}}} = 3$ and $\dim_{CBC} {Q_{\mathit{seat}}} = 3$), as they both are able to move around the room. Note, however, that we identify $3$ DOF here: 2D movement across the floor and rotation around its own axis. This is the case, if the capsule network has never actually seen the \symb{office-chair} being lifted up or tilted to the side and assumes that it is, in some sense, stuck to the floor (\ie, if \symb{office-chair} were interactable, it would be considered to be connected by a joint to the floor). Assume for now, that the \symb{office-chair} has a second verb attribute \attr{kicked}, which is just a fast swivel. Now, we need to vary across two verbs using $[0,1]^2$ in step (2.) of the algorithm and one could assume that this traces out a plane in $\mathbf{Q}_{\mathit{base},\mathit{seat}}$. However, by closer inspection we find that it does indeed still trace out the same curve as \attr{swivel}, thus preserving our single DOF for this hinge.

\section{Defining Relations}

With all the required features in place, we are able to form the sender-receiver-relation triplets $\langle o_i, o_j, \rho \rangle_k$ that serve as the input for $\phi_R$. In Table \ref{tab:relation}, the full definition of such a triplet is given, where we one-hot encode the symbol $\lambda_i$ and encode the attributes such that across all objects, the position in the vector is the same if they have the same name. This is important, as the type of attribute might give indications of certain physical properties, such as \attr{metallic} for magnetism.

{
	\renewcommand{\arraystretch}{2}
	\begin{table}[H]
		\begin{center}
			\resizebox{\textwidth}{!}{
				\begin{tabular}{ c  c  c  c  c : c  c  c  c  c : c  c  c  c } 
					\hline
					\multicolumn{5}{c :}{\cellcolor{green!10!white} Sender $o_i$} & \multicolumn{5}{c :}{\cellcolor{blue!10!white} Receiver $o_j$} & \multicolumn{4}{c}{\cellcolor{red!10!white} Relation $\rho$} \\
					
					\hline
					$\lambda_i$ & $\vec{\alpha}_i$ & $\frac{d \vec{\alpha}_i}{d t}$ & \makecell{static / \\ dynamic} & \makecell{rigid / \\ elastic} &  
					$\lambda_j$ & $\vec{\alpha}_j$ & $\frac{d \vec{\alpha}_j}{d t}$ & \makecell{static / \\ dynamic} & \makecell{rigid / \\ elastic} &
					$d_{o_i, o_j}$ & joint & $\vec{N}_{o_i}$ & $\vec{N}_{o_j} $ \\
					\hline

			\end{tabular}}
		\end{center}
		\caption{\label{tab:relation} Full relation triplet used as input for $\phi_R$.}
	\end{table}
}

Next, we aggregate the results of $\phi_R$ using $a(\cdot)$ and employ the format given in Table \ref{tab:aggregate} as input for $\phi_O$. 

{
	\renewcommand{\arraystretch}{2}
	\begin{table}[H]
		\begin{center}
			\begin{tabular}{ c  c  c  c  c : c  c } 
				\hline
				\multicolumn{5}{c :}{\cellcolor{blue!10!white} Receiver $o_j$} & \multicolumn{2}{c}{\cellcolor{yellow!10!white} Effects}  \\
				
				\hline
				$\lambda_i$ & $\vec{\alpha}_i$ & $\frac{d \vec{\alpha}_i}{d t}$ & \makecell{static / \\ dynamic} & \makecell{rigid / \\ elastic} &  
				$\Sigma e_k$ & $X$ \\
				\hline
				
			\end{tabular}
		\end{center}
		\caption{\label{tab:aggregate} Aggregated data $a(\cdot)$ used as input for $\phi_O$.}
	\end{table}
}

By being very explicit about the attributes and their lexical interpretation, the resulting interaction of $\phi_O$ exhibits the predicted characteristics: The regression model learns to associate attributes with physical properties. Revisiting our example, if the object has a high value for its \attr{metallic} attribute, it is more likely heavy or magnetic, changing its physical interaction with the environment compared to an object with a low \attr{metallic} value.

\section{Implementation and Results}

We implement the interaction network using a neural network for regression with six dense layers for $\phi_R$ and one with two dense layers for $\phi_O$. For our test, we train the model on synthetic video data of collisions between circles. To calculate the contracted box-counting dimensions for $\textbf{S}_i, \textbf{Q}_i, \textbf{Q}_{ij}$ and $\textbf{D}_{ij}$, we employ a Monte-Carlo approach to avoid varying over all of $[0,1]^n$. By randomly selecting $[x_1, x_1+\Delta x]\times\cdots\times[x_n, x_n+\Delta x]\subseteq [0,1]^n$ boxes and feeding them into the decoder $g$, we get transformed volumes $\hat{V}$, which we contract using Equation \ref{eq:contract} to find $V$. Repeating this process, we find a distribution for $\dim_{CBC} V$ and choose the maximum of this distribution as the true dimension for the space we are analyzing.

The distance in Equation \ref{eq:continue3} is calculated explicitly through the signed distance functions used in our capsule implementation.

\begin{figure}[H]
	\centering
	\begin{adjustbox}{max width=1.0\textwidth}
		\begin{tikzpicture}
		\node[inner sep=0pt] at (3,0) {\includegraphics[width=.70\textwidth]{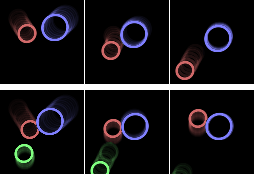}};
		\end{tikzpicture}
	\end{adjustbox}
	\caption{Top shows the prediction of two circles interacting. Bottom shows the prediction of three circles interacting, whereas the interaction network was only trained on two. Movement is illustrated using color and object trails.} \label{fig:results1}
\end{figure}

For testing, the interaction network is fed with an observation from episodic memory. Figure \ref{fig:results1} shows how our implementation predicts the interaction of two and three circles. This prediction is based on the capsule network having seen two frames with these circles and is able to simulate their interaction plausibly for $\sim 50$ frames thereafter.

\begin{figure}[H]
	\centering
	\begin{adjustbox}{max width=1.0\textwidth}
		\begin{tikzpicture}
		\node[inner sep=0pt] at (0,0) {\includegraphics[width=.70\textwidth]{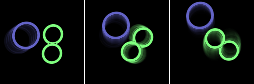}};
		\end{tikzpicture}
	\end{adjustbox}
	\caption{Interaction of a circle with a figure-eight, which is only able to rotate around its axis. Movement is illustrated using color and object trails.} \label{fig:results2}
\end{figure}

In the next experiment, we use a figure-eight that is unable to move horizontally and vertically and only rotates like a windmill ($\dim_{CBC} {Q_{\mathit{eight}}} = 1$). Next, we launch a circle towards the figure-eight and predict the physical interaction, as seen in Figure \ref{fig:results2}, and see that it indeed behaves plausibly by rotating on collision.

\clearpage

% TODO: Figure that explains how Fabricate works? How all three work?

\chapter{Mental Simulation}\label{sec:Sim}\thispagestyle{empty}

\begin{wrapfigure}{R}{0.4\textwidth}
	\centering
	\begin{adjustbox}{max width=0.4\textwidth}
		\begin{tikzpicture}
		
		\vividnet{black!20!white}{black!20!white}{black!20!white}{black!20!white}{black!20!white}{black!10!blue}{black!10!blue}

		\end{tikzpicture}
	\end{adjustbox}
\end{wrapfigure}

With all elements needed to construct our mental simulation framework in place, we tie them together into a querying process. In this chapter we describe the extraction of knowledge from our architecture in detail. We subdivide the used queries into three types, which can be recombined in any way:

\begin{itemize}
	\item \textbf{Replay:} Extracting existing information.
	\item \textbf{Predict:} Making a prediction based on a given state.
	\item \textbf{Fabricate:} Creating a fictive scenario.
\end{itemize}

Each of these query types is associated with one of the previously discussed parts: Replaying events targets episodic memory, predicting events relies on intuitive physics and fabricating events utilizes the capsule network itself. We begin by introducing a query language and proceed to explore some of the possible applications it enables. The presented framework is designed as a building block for planning agents \citep{Russel:2010} and the query language acts as its interface.\clearpage

\section{Querying}

Mental simulation is the act of imagining an action and simulating the results in order to make a decision. To design our query mechanism, we must break down this task into basic components. First, to imagine something we need a starting point and some way to \textbf{replay} it. This starting point has to be modifiable in order to \textbf{fabricate} the effect of an action or a new situation. Finally, we \textbf{predict} future outcomes on which an agent would base its decision. The entirety of this process is our mental simulation. We now discuss these three individual actions in detail.

\medskip
\noindent\textbf{Replay: }To replay an event, a sequence of time-ordered observations in episodic memory is chosen and returned or played. This is akin to video playback, but a replay on episodic memory is not limited to one type of observation or a single time-line.

\medskip
\noindent\textbf{Predict: }Intuitive physics is used to predict the next time-step of some prior observation. Even though this prediction itself already has the structure of an observed parse-tree ready to be saved, it is instead fed into the capsule network and only the resulting observation is stored as a prediction in episodic memory. This final step might seem redundant, however, we illustrate by an example that it is indeed important to let the capsule network know about the semantics. Consider a half open door and a ball flying towards it. Intuitive physics then predicts that the ball hits the door, closing it, and then flies off at an angle. Our capsule network may have never seen a door in the closed state before. By feeding these new semantics back into the network, we force the meta-learning pipeline to act and learn a potentially new data point, prior to even observing it in a real scene and purely from mental simulation.

\medskip
\noindent\textbf{Fabricate: }We fabricate events by altering given observations at certain nodes in the parse-tree that are of interest or by inventing completely new ones and feeding the result to the capsule network, as we did with predicted observations. For example, if the oracle received the description or proportions of a new object through a different source, such as a verbal explanation, the capsule network and its meta-learning pipeline are able to learn the new visual semantics by fabricating a scene and imagining it. 

\subsection{Expanded Episodic Memory}

We expand our episodic memory to handle, apart from perceived observations from actual visual input, two more types of entries: fabricated and predicted observations. 

The perceived observations form the main time-line of episodic memory. When we make a prediction based on such an observation, a new forked time-line is created. In this fork, new predictions can be made to further develop this time-line, or it may even be forked again. As predictions, fabrications are able to create a new fork, but may also spawn a completely separate time-line. Further, forks can be merged at any point. It is, however, forbidden that predictions and fabrications occur on the main time-line. Without such a restriction, it would be similarly to a human not able to distinguish reality from fiction.

To illustrate these three observation types, we consider the following example: A person sees a ball fly towards a wall (perceived observations). This person predicts that a few time-steps into the future the ball will bounce of the wall and fly towards the ground (predicted observations). If the predictions concur with what is observed shortly thereafter, a merge of the expected fork with the main time-line occurs, if not, the two branches diverge. Further, the person in our example may also imagine a second ball (fabricated observation) and how it flies and interacts with the first ball (predicted observations based on the fabricated observation). This branch begins completely independent of other time-lines, but may merge at some later point, if the set of events should ever occur in reality.

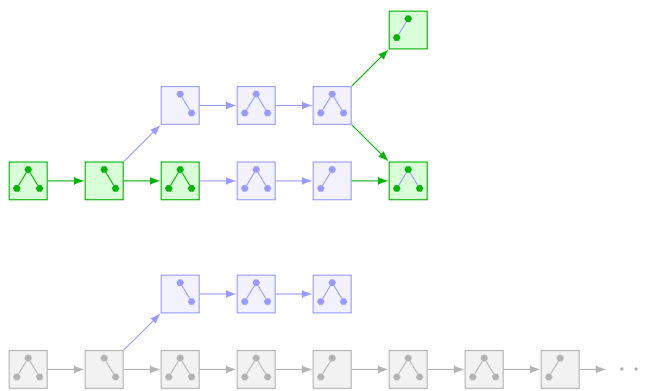
\begin{figure}[H]
	\centering
	\begin{adjustbox}{max width=1\textwidth}
		\begin{tikzpicture}

		%- Episodic Memory
		\node (E1) [fill=black!05!white, draw=black!30!white, minimum width=0.5cm, minimum height=0.5cm] at (-2.1cm, 0.3cm) {};
		\node (E2) [fill=black!05!white, draw=black!30!white, minimum width=0.5cm, minimum height=0.5cm] at (-1.1cm, 0.3cm) {};
		\node (E3) [fill=black!05!white, draw=black!30!white, minimum width=0.5cm, minimum height=0.5cm] at (-0.1cm, 0.3cm) {};
		\node (E4) [fill=black!05!white, draw=black!30!white, minimum width=0.5cm, minimum height=0.5cm] at (0.9cm, 0.3cm) {};
		\node (E5) [fill=black!05!white, draw=black!30!white, minimum width=0.5cm, minimum height=0.5cm] at (1.9cm, 0.3cm) {};
		\node (E6) [fill=black!05!white, draw=black!30!white, minimum width=0.5cm, minimum height=0.5cm] at (2.9cm, 0.3cm) {};
		\node (E7) [fill=black!05!white, draw=black!30!white, minimum width=0.5cm, minimum height=0.5cm] at (3.9cm, 0.3cm) {};
		\node (E8) [fill=black!05!white, draw=black!30!white, minimum width=0.5cm, minimum height=0.5cm] at (4.9cm, 0.3cm) {};
		\node (EX) [text=black!30!white, minimum width=0.5cm, minimum height=0.5cm] at (5.9cm, 0.3cm) {$\cdots$};
		
		\draw[tochild, black!30!white] (E1) -- (E2);
		\draw[tochild, black!30!white] (E2) -- (E3);
		\draw[tochild, black!30!white] (E3) -- (E4);
		\draw[tochild, black!30!white] (E4) -- (E5);
		\draw[tochild, black!30!white] (E5) -- (E6);
		\draw[tochild, black!30!white] (E6) -- (E7);
		\draw[tochild, black!30!white] (E7) -- (E8);
		\draw[tochild, black!30!white] (E8) -- (EX);
		
		\node (E1Cap1) [regular polygon,regular polygon sides=6, fill=black!30!white, minimum width=1cm, minimum height=1cm, scale=0.1] at (-2.1cm, 0.45cm) {};
		\node (E1Cap2) [regular polygon,regular polygon sides=6, fill=black!30!white, minimum width=1cm, minimum height=1cm, scale=0.1] at (-2.25cm, 0.2cm) {};
		\node (E1Cap3) [regular polygon,regular polygon sides=6, fill=black!30!white, minimum width=1cm, minimum height=1cm, scale=0.1] at (-1.95cm, 0.2cm) {};
		\draw[black!30!white, scale=0.1] (E1Cap1) -- (E1Cap2);
		\draw[black!30!white, scale=0.1] (E1Cap1) -- (E1Cap3);		
		
		\node (E2Cap1) [regular polygon,regular polygon sides=6, fill=black!30!white, minimum width=1cm, minimum height=1cm, scale=0.1] at (-1.1cm, 0.45cm) {};
		\node (E2Cap3) [regular polygon,regular polygon sides=6, fill=black!30!white, minimum width=1cm, minimum height=1cm, scale=0.1] at (-0.95cm, 0.2cm) {};
		\draw[black!30!white, scale=0.1] (E2Cap1) -- (E2Cap3);
		
		\node (E3Cap1) [regular polygon,regular polygon sides=6, fill=black!30!white, minimum width=1cm, minimum height=1cm, scale=0.1] at (-0.1cm, 0.45cm) {};
		\node (E3Cap2) [regular polygon,regular polygon sides=6, fill=black!30!white, minimum width=1cm, minimum height=1cm, scale=0.1] at (-0.25cm, 0.2cm) {};
		\node (E3Cap3) [regular polygon,regular polygon sides=6, fill=black!30!white, minimum width=1cm, minimum height=1cm, scale=0.1] at (0.05cm, 0.2cm) {};
		\draw[black!30!white, scale=0.1] (E3Cap1) -- (E3Cap2);
		\draw[black!30!white, scale=0.1] (E3Cap1) -- (E3Cap3);
		
		\node (E4Cap1) [regular polygon,regular polygon sides=6, fill=black!30!white, minimum width=1cm, minimum height=1cm, scale=0.1] at (0.9cm, 0.45cm) {};
		\node (E4Cap2) [regular polygon,regular polygon sides=6, fill=black!30!white, minimum width=1cm, minimum height=1cm, scale=0.1] at (0.75cm, 0.2cm) {};
		\node (E4Cap3) [regular polygon,regular polygon sides=6, fill=black!30!white, minimum width=1cm, minimum height=1cm, scale=0.1] at (1.05cm, 0.2cm) {};
		\draw[black!30!white, scale=0.1] (E4Cap1) -- (E4Cap2);
		\draw[black!30!white, scale=0.1] (E4Cap1) -- (E4Cap3);
		
		\node (E5Cap1) [regular polygon,regular polygon sides=6, fill=black!30!white, minimum width=1cm, minimum height=1cm, scale=0.1] at (1.9cm, 0.45cm) {};
		\node (E5Cap2) [regular polygon,regular polygon sides=6, fill=black!30!white, minimum width=1cm, minimum height=1cm, scale=0.1] at (1.75cm, 0.2cm) {};
		\draw[black!30!white, scale=0.1] (E5Cap1) -- (E5Cap2);
		
		\node (E6Cap1) [regular polygon,regular polygon sides=6, fill=black!30!white, minimum width=1cm, minimum height=1cm, scale=0.1] at (2.9cm, 0.45cm) {};
		\node (E6Cap2) [regular polygon,regular polygon sides=6, fill=black!30!white, minimum width=1cm, minimum height=1cm, scale=0.1] at (2.75cm, 0.2cm) {};
		\node (E6Cap3) [regular polygon,regular polygon sides=6, fill=black!30!white, minimum width=1cm, minimum height=1cm, scale=0.1] at (3.05cm, 0.2cm) {};
		\draw[black!30!white, scale=0.1] (E6Cap1) -- (E6Cap2);
		\draw[black!30!white, scale=0.1] (E6Cap1) -- (E6Cap3);
		
		\node (E7Cap1) [regular polygon,regular polygon sides=6, fill=black!30!white, minimum width=1cm, minimum height=1cm, scale=0.1] at (3.9cm, 0.45cm) {};
		\node (E7Cap2) [regular polygon,regular polygon sides=6, fill=black!30!white, minimum width=1cm, minimum height=1cm, scale=0.1] at (3.75cm, 0.2cm) {};
		\node (E7Cap3) [regular polygon,regular polygon sides=6, fill=black!30!white, minimum width=1cm, minimum height=1cm, scale=0.1] at (4.05cm, 0.2cm) {};
		\draw[black!30!white, scale=0.1] (E7Cap1) -- (E7Cap2);
		\draw[black!30!white, scale=0.1] (E7Cap1) -- (E7Cap3);
		
		\node (E8Cap1) [regular polygon,regular polygon sides=6, fill=black!30!white, minimum width=1cm, minimum height=1cm, scale=0.1] at (4.9cm, 0.45cm) {};
		\node (E8Cap2) [regular polygon,regular polygon sides=6, fill=black!30!white, minimum width=1cm, minimum height=1cm, scale=0.1] at (4.75cm, 0.2cm) {};
		\draw[black!30!white, scale=0.1] (E8Cap1) -- (E8Cap2);
		
		%- Prediction of main branch
		
		\node (P3) [fill=blue!05!white, draw=blue!40!white, minimum width=0.5cm, minimum height=0.5cm] at (-0.1cm, 1.3cm) {};
		\node (P4) [fill=blue!05!white, draw=blue!40!white, minimum width=0.5cm, minimum height=0.5cm] at (0.9cm, 1.3cm) {};
		\node (P5) [fill=blue!05!white, draw=blue!40!white, minimum width=0.5cm, minimum height=0.5cm] at (1.9cm, 1.3cm) {};

		\node (P3Cap1) [regular polygon,regular polygon sides=6, fill=blue!40!white, minimum width=1cm, minimum height=1cm, scale=0.1] at (-0.1cm, 1.45cm) {};
		\node (P3Cap3) [regular polygon,regular polygon sides=6, fill=blue!40!white, minimum width=1cm, minimum height=1cm, scale=0.1] at (0.05cm, 1.2cm) {};
		\draw[blue!40!white, scale=0.1] (P3Cap1) -- (P3Cap3);
		
		\node (P4Cap1) [regular polygon,regular polygon sides=6, fill=blue!40!white, minimum width=1cm, minimum height=1cm, scale=0.1] at (0.9cm, 1.45cm) {};
		\node (P4Cap2) [regular polygon,regular polygon sides=6, fill=blue!40!white, minimum width=1cm, minimum height=1cm, scale=0.1] at (0.75cm, 1.2cm) {};
		\node (P4Cap3) [regular polygon,regular polygon sides=6, fill=blue!40!white, minimum width=1cm, minimum height=1cm, scale=0.1] at (1.05cm, 1.2cm) {};
		\draw[blue!40!white, scale=0.1] (P4Cap1) -- (P4Cap2);
		\draw[blue!40!white, scale=0.1] (P4Cap1) -- (P4Cap3);
		
		\node (P5Cap1) [regular polygon,regular polygon sides=6, fill=blue!40!white, minimum width=1cm, minimum height=1cm, scale=0.1] at (1.9cm, 1.45cm) {};
		\node (P5Cap2) [regular polygon,regular polygon sides=6, fill=blue!40!white, minimum width=1cm, minimum height=1cm, scale=0.1] at (1.75cm, 1.2cm) {};
		\node (P5Cap3) [regular polygon,regular polygon sides=6, fill=blue!40!white, minimum width=1cm, minimum height=1cm, scale=0.1] at (2.05cm, 1.2cm) {};
		\draw[blue!40!white, scale=0.1] (P5Cap1) -- (P5Cap2);
		\draw[blue!40!white, scale=0.1] (P5Cap1) -- (P5Cap3);
		
		\draw[tochild, blue!40!white] (E2) -- (P3);
		\draw[tochild, blue!40!white] (P3) -- (P4);
		\draw[tochild, blue!40!white] (P4) -- (P5);

		%- Fabricated Timeline
		\node (S1) [fill=green!15!white, draw=black!30!green, minimum width=0.5cm, minimum height=0.5cm] at (-2.1cm, 2.8cm) {};
		\node (S2) [fill=green!15!white, draw=black!30!green, minimum width=0.5cm, minimum height=0.5cm] at (-1.1cm, 2.8cm) {};
		\node (S3) [fill=green!15!white, draw=black!30!green, minimum width=0.5cm, minimum height=0.5cm] at (-0.1cm, 2.8cm) {};
		\node (S4) [fill=blue!05!white, draw=blue!40!white, minimum width=0.5cm, minimum height=0.5cm] at (0.9cm, 2.8cm) {};
		\node (S5) [fill=blue!05!white, draw=blue!40!white, minimum width=0.5cm, minimum height=0.5cm] at (1.9cm, 2.8cm) {};
		\node (S6) [fill=green!15!white, draw=black!30!green, minimum width=0.5cm, minimum height=0.5cm] at (2.9cm, 2.8cm) {};
		\node (Sx6) [fill=green!15!white, draw=black!30!green, minimum width=0.5cm, minimum height=0.5cm] at (2.9cm, 4.8cm) {};
		
		\draw[tochild, black!30!green] (S1) -- (S2);
		\draw[tochild, black!30!green] (S2) -- (S3);
		\draw[tochild, blue!40!white] (S3) -- (S4);
		\draw[tochild, blue!40!white] (S4) -- (S5);
		\draw[tochild, black!30!green] (S5) -- (S6);
		
		\node (S1Cap1) [regular polygon,regular polygon sides=6, fill=black!30!green, minimum width=1cm, minimum height=1cm, scale=0.1] at (-2.1cm, 2.95cm) {};
		\node (S1Cap2) [regular polygon,regular polygon sides=6, fill=black!30!green, minimum width=1cm, minimum height=1cm, scale=0.1] at (-2.25cm, 2.7cm) {};
		\node (S1Cap3) [regular polygon,regular polygon sides=6, fill=black!30!green, minimum width=1cm, minimum height=1cm, scale=0.1] at (-1.95cm, 2.7cm) {};
		\draw[black!30!green, scale=0.1] (S1Cap1) -- (S1Cap2);
		\draw[black!30!green, scale=0.1] (S1Cap1) -- (S1Cap3);		
		
		\node (S2Cap1) [regular polygon,regular polygon sides=6, fill=black!30!green, minimum width=1cm, minimum height=1cm, scale=0.1] at (-1.1cm, 2.95cm) {};
		\node (S2Cap3) [regular polygon,regular polygon sides=6, fill=black!30!green, minimum width=1cm, minimum height=1cm, scale=0.1] at (-0.95cm, 2.7cm) {};
		\draw[black!30!green, scale=0.1] (S2Cap1) -- (S2Cap3);
		
		\node (S3Cap1) [regular polygon,regular polygon sides=6, fill=black!30!green, minimum width=1cm, minimum height=1cm, scale=0.1] at (-0.1cm, 2.95cm) {};
		\node (S3Cap2) [regular polygon,regular polygon sides=6, fill=black!30!green, minimum width=1cm, minimum height=1cm, scale=0.1] at (-0.25cm, 2.7cm) {};
		\node (S3Cap3) [regular polygon,regular polygon sides=6, fill=black!30!green, minimum width=1cm, minimum height=1cm, scale=0.1] at (0.05cm, 2.7cm) {};
		\draw[black!30!green, scale=0.1] (S3Cap1) -- (S3Cap2);
		\draw[black!30!green, scale=0.1] (S3Cap1) -- (S3Cap3);
		
		\node (S4Cap1) [regular polygon,regular polygon sides=6, fill=blue!40!white, minimum width=1cm, minimum height=1cm, scale=0.1] at (0.9cm, 2.95cm) {};
		\node (S4Cap2) [regular polygon,regular polygon sides=6, fill=blue!40!white, minimum width=1cm, minimum height=1cm, scale=0.1] at (0.75cm, 2.7cm) {};
		\node (S4Cap3) [regular polygon,regular polygon sides=6, fill=blue!40!white, minimum width=1cm, minimum height=1cm, scale=0.1] at (1.05cm, 2.7cm) {};
		\draw[blue!40!white, scale=0.1] (S4Cap1) -- (S4Cap2);
		\draw[blue!40!white, scale=0.1] (S4Cap1) -- (S4Cap3);
		
		\node (S5Cap1) [regular polygon,regular polygon sides=6, fill=blue!40!white, minimum width=1cm, minimum height=1cm, scale=0.1] at (1.9cm, 2.95cm) {};
		\node (S5Cap2) [regular polygon,regular polygon sides=6, fill=blue!40!white, minimum width=1cm, minimum height=1cm, scale=0.1] at (1.75cm, 2.7cm) {};
		\draw[blue!40!white, scale=0.1] (S5Cap1) -- (S5Cap2);
		
		\node (S6Cap1) [regular polygon,regular polygon sides=6, fill=black!30!green, minimum width=1cm, minimum height=1cm, scale=0.1] at (2.9cm, 2.95cm) {};
		\node (S6Cap2) [regular polygon,regular polygon sides=6, fill=black!30!green, minimum width=1cm, minimum height=1cm, scale=0.1] at (2.75cm, 2.7cm) {};
		\node (S6Cap3) [regular polygon,regular polygon sides=6, fill=black!30!green, minimum width=1cm, minimum height=1cm, scale=0.1] at (3.05cm, 2.7cm) {};
		\draw[blue!40!white, scale=0.1] (S6Cap1) -- (S6Cap2);
		\draw[blue!40!white, scale=0.1] (S6Cap1) -- (S6Cap3);
		
		\node (Sx6Cap1) [regular polygon,regular polygon sides=6, fill=black!30!green, minimum width=1cm, minimum height=1cm, scale=0.1] at (2.9cm, 4.95cm) {};
		\node (Sx6Cap2) [regular polygon,regular polygon sides=6, fill=black!30!green, minimum width=1cm, minimum height=1cm, scale=0.1] at (2.75cm, 4.7cm) {};
		\draw[blue!40!white, scale=0.1] (Sx6Cap1) -- (Sx6Cap2);
		
		%- Fabricated Prediction of Fabricated branch
		
		\node (PS3) [fill=blue!05!white, draw=blue!40!white, minimum width=0.5cm, minimum height=0.5cm] at (-0.1cm, 3.8cm) {};
		\node (PS4) [fill=blue!05!white, draw=blue!40!white, minimum width=0.5cm, minimum height=0.5cm] at (0.9cm, 3.8cm) {};
		\node (PS5) [fill=blue!05!white, draw=blue!40!white, minimum width=0.5cm, minimum height=0.5cm] at (1.9cm, 3.8cm) {};

		\node (PS3Cap1) [regular polygon,regular polygon sides=6, fill=blue!40!white, minimum width=1cm, minimum height=1cm, scale=0.1] at (-0.1cm, 3.95cm) {};
		\node (PS3Cap3) [regular polygon,regular polygon sides=6, fill=blue!40!white, minimum width=1cm, minimum height=1cm, scale=0.1] at (0.05cm, 3.7cm) {};
		\draw[blue!40!white, scale=0.1] (PS3Cap1) -- (PS3Cap3);
		
		\node (PS4Cap1) [regular polygon,regular polygon sides=6, fill=blue!40!white, minimum width=1cm, minimum height=1cm, scale=0.1] at (0.9cm, 3.95cm) {};
		\node (PS4Cap2) [regular polygon,regular polygon sides=6, fill=blue!40!white, minimum width=1cm, minimum height=1cm, scale=0.1] at (0.75cm, 3.7cm) {};
		\node (PS4Cap3) [regular polygon,regular polygon sides=6, fill=blue!40!white, minimum width=1cm, minimum height=1cm, scale=0.1] at (1.05cm, 3.7cm) {};
		\draw[blue!40!white, scale=0.1] (PS4Cap1) -- (PS4Cap2);
		\draw[blue!40!white, scale=0.1] (PS4Cap1) -- (PS4Cap3);
		
		\node (PS5Cap1) [regular polygon,regular polygon sides=6, fill=blue!40!white, minimum width=1cm, minimum height=1cm, scale=0.1] at (1.9cm, 3.95cm) {};
		\node (PS5Cap2) [regular polygon,regular polygon sides=6, fill=blue!40!white, minimum width=1cm, minimum height=1cm, scale=0.1] at (1.75cm, 3.7cm) {};
		\node (PS5Cap3) [regular polygon,regular polygon sides=6, fill=blue!40!white, minimum width=1cm, minimum height=1cm, scale=0.1] at (2.05cm, 3.7cm) {};
		\draw[blue!40!white, scale=0.1] (PS5Cap1) -- (PS5Cap2);
		\draw[blue!40!white, scale=0.1] (PS5Cap1) -- (PS5Cap3);
		
		\draw[tochild, blue!40!white] (S2) -- (PS3);
		\draw[tochild, blue!40!white] (PS3) -- (PS4);
		\draw[tochild, blue!40!white] (PS4) -- (PS5);
		\draw[tochild, black!30!green] (PS5) -- (S6);
		\draw[tochild, black!30!green] (PS5) -- (Sx6);
		
		\end{tikzpicture}
	\end{adjustbox}
	\caption{Example of episodic memory. The main time-line of perceived observations (gray) and a parallel time-line of fabricated observations (green), each with predicted observations (blue).} \label{fig:TimeLines}
\end{figure}

In Figure \ref{fig:TimeLines} a schematic depiction of how episodic memory stores these time-lines is given.

\subsection{Visual Knowledge}

Using a combination of the three queries (replay, predict, fabricate), we are able to extract visual knowledge that has been learned about the environment. For this task we outline a query syntax, inspired by other query languages such as SQL, GraphQL and SPARQL.

We first analyze in what tiers episodic memory is organized: 
\begin{itemize}
	\setlength{\itemindent}{1.0cm}
	\item [1. Tier:] Time-line data is stored as a disconnected network graph of perceived, predicted and fabricated observations (\cf Figure \ref{fig:TimeLines}). 
	\item [2. Tier:] Observations are stored together with meta-information and a tree-structure of observed capsules (\cf Figure \ref{fig:SemanticNetwork}).
	\item [3. Tier:] Observed capsules are stored with their activation probability and a collection of attributes (\cf Equation \ref{eq:memoryTrainingSet}).
\end{itemize}

\begin{figure}[H]
	\centering
	\begin{adjustbox}{max width=1\textwidth}
		\begin{tikzpicture}
		
		%- Knowledge Bubbles
		
		\node (R1) [regular polygon,regular polygon sides=6, fill=white!90!yellow, draw=black, minimum width=2cm] at (6.4cm, 0.3cm) {};
		\node (R2) [regular polygon,regular polygon sides=6, fill=white!90!yellow, draw=black, minimum width=2cm] at (9.4cm, 0.3cm) {};
		\node (R3) [regular polygon,regular polygon sides=6, fill=white!90!yellow, draw=black, minimum width=2cm] at (9.4cm, 3.0cm) {};
		\node (R4) [regular polygon,regular polygon sides=6, fill=white!90!yellow, draw=black, minimum width=2cm] at (6.4cm, 3.0cm) {};

		%- Knowledge
		\node (K1) [fill=green!15!white, draw=black!30!green, minimum width=0.5cm, minimum height=0.5cm] at (5.9cm, 0.3cm) {};
		\node (K2) [fill=black!05!white, draw=black!30!white, minimum width=0.5cm, minimum height=0.5cm] at (6.6cm, 0.7cm) {};
		\node (K3) [fill=green!15!white, draw=black!30!green, minimum width=0.5cm, minimum height=0.5cm] at (6.4cm, 3.0cm) {};
		\node (K4) [fill=black!05!white, draw=black!30!white, minimum width=0.5cm, minimum height=0.5cm] at (8.9cm, 0.3cm) {};
		\node (K5) [fill=black!05!white, draw=black!30!white, minimum width=0.5cm, minimum height=0.5cm] at (9.9cm, 0.3cm) {};
		\node (K6) [fill=black!05!white, draw=black!30!white, minimum width=0.5cm, minimum height=0.5cm] at (9.4cm, 3.0cm) {};
		\node (K8) [fill=blue!05!white, draw=blue!40!white, minimum width=0.5cm, minimum height=0.5cm] at (6.6cm, -0.1cm) {};
		
		\draw[tochild, black!30!white] (K2) -- (K4);
		\draw[tochild, black!30!white] (K4) -- (K5);
		\draw[tochild, black!30!white] (K5) -- (K6);
		\draw[tochild, black!30!green] (K3) -- (K1);
		\draw[tochild, blue!40!white] (K1) -- (K8);
		
		\node (K1Cap1) [regular polygon,regular polygon sides=6, fill=black!30!green, minimum width=1cm, minimum height=1cm, scale=0.1] at (5.9cm, 0.45cm) {};
		\node (K1Cap2) [regular polygon,regular polygon sides=6, fill=black!30!green, minimum width=1cm, minimum height=1cm, scale=0.1] at (5.75cm, 0.2cm) {};
		\node (K1Cap3) [regular polygon,regular polygon sides=6, fill=black!30!green, minimum width=1cm, minimum height=1cm, scale=0.1] at (6.05cm, 0.2cm) {};
		\draw[black!30!green, scale=0.1] (K1Cap1) -- (K1Cap2);
		\draw[black!30!green, scale=0.1] (K1Cap1) -- (K1Cap3);		
		
		\node (K2Cap1) [regular polygon,regular polygon sides=6, fill=black!30!white, minimum width=1cm, minimum height=1cm, scale=0.1] at (6.6cm, 0.85cm) {};
		\node (K2Cap2) [regular polygon,regular polygon sides=6, fill=black!30!white, minimum width=1cm, minimum height=1cm, scale=0.1] at (6.45cm, 0.6cm) {};
		\node (K2Cap3) [regular polygon,regular polygon sides=6, fill=black!30!white, minimum width=1cm, minimum height=1cm, scale=0.1] at (6.75cm, 0.6cm) {};
		\draw[black!30!white, scale=0.1] (K2Cap1) -- (K2Cap2);
		\draw[black!30!white, scale=0.1] (K2Cap1) -- (K2Cap3);
		
		\node (K3Cap1) [regular polygon,regular polygon sides=6, fill=black!30!green, minimum width=1cm, minimum height=1cm, scale=0.1] at (6.4cm, 3.15cm) {};
		\node (K3Cap2) [regular polygon,regular polygon sides=6, fill=black!30!green, minimum width=1cm, minimum height=1cm, scale=0.1] at (6.25cm, 2.9cm) {};
		\draw[black!30!green, scale=0.1] (K3Cap1) -- (K3Cap2);
		
		\node (K4Cap1) [regular polygon,regular polygon sides=6, fill=black!30!white, minimum width=1cm, minimum height=1cm, scale=0.1] at (8.9cm, 0.45cm) {};
		\node (K4Cap3) [regular polygon,regular polygon sides=6, fill=black!30!white, minimum width=1cm, minimum height=1cm, scale=0.1] at (9.05cm, 0.2cm) {};
		\draw[black!30!white, scale=0.1] (K4Cap1) -- (K4Cap3);
		
		\node (K5Cap1) [regular polygon,regular polygon sides=6, fill=black!30!white, minimum width=1cm, minimum height=1cm, scale=0.1] at (9.9cm, 0.45cm) {};
		\node (K5Cap2) [regular polygon,regular polygon sides=6, fill=black!30!white, minimum width=1cm, minimum height=1cm, scale=0.1] at (9.75cm, 0.2cm) {};
		\node (K5Cap3) [regular polygon,regular polygon sides=6, fill=black!30!white, minimum width=1cm, minimum height=1cm, scale=0.1] at (10.05cm, 0.2cm) {};
		\draw[black!30!white, scale=0.1] (K5Cap1) -- (K5Cap2);
		\draw[black!30!white, scale=0.1] (K5Cap1) -- (K5Cap3);
		
		\node (K6Cap1) [regular polygon,regular polygon sides=6, fill=black!30!white, minimum width=1cm, minimum height=1cm, scale=0.1] at (9.4cm, 3.15cm) {};
		\node (K6Cap2) [regular polygon,regular polygon sides=6, fill=black!30!white, minimum width=1cm, minimum height=1cm, scale=0.1] at (9.25cm, 2.9cm) {};
		\node (K6Cap3) [regular polygon,regular polygon sides=6, fill=black!30!white, minimum width=1cm, minimum height=1cm, scale=0.1] at (9.55cm, 2.9cm) {};
		\draw[black!30!white, scale=0.1] (K6Cap1) -- (K6Cap2);
		\draw[black!30!white, scale=0.1] (K6Cap1) -- (K6Cap3);
		
		\node (K8Cap1) [regular polygon,regular polygon sides=6, fill=blue!40!white, minimum width=1cm, minimum height=1cm, scale=0.1] at (6.6cm, 0.05cm) {};
		\node (K8Cap3) [regular polygon,regular polygon sides=6, fill=blue!40!white, minimum width=1cm, minimum height=1cm, scale=0.1] at (6.75cm, -0.2cm) {};
		\draw[blue!40!white, scale=0.1] (K8Cap1) -- (K8Cap3);
		
		\end{tikzpicture}
	\end{adjustbox}
	\caption{Creating a graph-of-graphs by clustering entire or partial observations filtered from episodic memory.} \label{fig:GraphOfGraphs}
\end{figure}
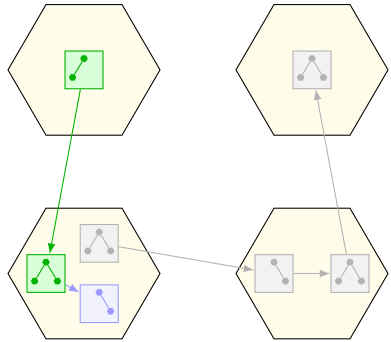

Due to the heterogeneous nature of each tier, the query syntax we introduce must be flexible enough to produce these different representations, so that we do not have to introduce a separate syntax for each tier. The output data structure is chosen to be a disconnected network graph, as it can be organized as a graph-of-graphs (1. Tier), a tree-structure (2. Tier), as well as a pure collection (3. Tier). The content of this graph depends on the tier the query operates on. An example for the highest tier can be seen in Figure \ref{fig:GraphOfGraphs}.

Next, we outline a simple query specification to extract knowledge given one of the three query types. In terms of SQL, replay can be seen as a \texttt{SELECT}, fabricate as an \texttt{INSERT} and predict as an \texttt{UPDATE}, although this analogy isn't entirely true, as a predict adds an updated observation to the memory instead of updating it in place.

We do not specify the exact grammar for this language, but rather keep it on a pseudo-code level. The structure for the queries is as follows:

\begin{align}
\textit{Results}, \textit{Frames} \gets & \textsc{Replay}(\textit{Origin},\textit{Filters}, \textit{Mappings}, \textit{Groupings}) \\
\textit{Observation} \gets & \textsc{Predict}(\textit{Origin}, \textit{Duration}, \textit{Forces}) \\
\textit{Observation} \gets & \textsc{Fabricate}(\textit{Origin}, \textit{Changes})
\end{align}

\begin{itemize}
	
	\item \textit{Results: } The results from a replay query, the tier format of it being dynamic and determined by the query itself.
	
	\item \textit{Frames: } A series of frames rendered based on the results. If the results contain symbols instead of observations, each symbol is rendered individually, as if it were an observation of a single object. Queries that return non-renderable parts (attributes or meta-information) return a blank image.
	
	\item \textit{Observation: } The observation that results from a predict or fabricate query.
	
	\item \textit{Origin: } Starting point for a query, such as an observation, and acts like a \texttt{FROM} clause. May be empty for fabricate. In this case, a completely new time-line is created instead of a fork. For replay, it acts as a pre-filter and determines which observations are affected by the query.
	
	\item \textit{Filters: } Generally, each query is concerned with all information found in memory, \ie, every symbol, attribute, connection (\ie, the edge between two consecutive observations) and piece of meta-information. Filters are designed to exclude uninteresting or redundant information. For example, if we are only interested in recent events, we define a filter to exclude all parse-trees before some specific time-stamp or only include symbols whose \attr{red} attribute is larger than some threshold. These filters act like a \texttt{WHERE} clause.
	
	\item \textit{Mappings: } To allow for simpler grouping, we map semantic structures to new ones. For example, we could define an adverb-mapping that maps all adjectives and their values, such as $\{\textit{metallic}:0.8\}$, to an adverb-adjective combination if they exceed some threshold, \eg, $\{\textit{metallic}:0.8\}$ becomes $\{\textit{very metallic}\}$. These mappings are equivalent to functions in other query languages.
		
	\item \textit{Groupings: } For the final grouping, clauses may be added that act as a \texttt{GROUP BY} and a \texttt{HAVING} statement. They cluster together entire parse-trees or also parts of them in a graph-of-graphs structure.
			
	\item \textit{Duration: } The duration intuitive physics is supposed to predict starting from the origin.
	
	\item \textit{Forces: } External forces (\cf Equation \ref{eq:ExtForces}) applied to an object or scene.
	
	\item \textit{Changes: } Changes that are applied to the origin of a fabricated observation, or the new parse-tree that is created if no origin is provided.
		
\end{itemize}

Only a replay query returns processed information, namely the filtered, mapped and grouped data, as well as the corresponding rendered images. While the images aren't required, as all the requested information is provided through the semantics, they do serve a useful purpose to visualize the results and are employed in some of the applications we explore in the following section. Furthermore, they strengthen the notion of mental simulation and imagination.

\section{Mental Simulation}

With the ability to query our framework, it is able to function as a mental simulation engine. The capability for simulations is learned through observing the environment and occasionally interacting with an oracle. While our framework is not capable of performing any creative task by itself, its ability to simulate offers a platform for exploration for an oracle. We quote from \citep{Hamrick2:2017}:

\begin{quote}
\enquote{There is a sense in which mental simulation is not only useful for prediction and inference, but also for exploration. For example, \citet{Finke:1988} showed that people can use mental simulation to discover novel and creative combinations of simple objects, such as combining the shapes \texttt{V}, \texttt{O}, and \texttt{D} into an ice cream cone (Rotate the \texttt{D} by 90° left, place it on the \texttt{V}, and then place the \texttt{O} on the very top). Given mental simulation's importance in planning, which inherently must deal with the exploration-exploitation trade-off \citep{Sutton:1998}, it seems unsurprising that the mind might be adapted towards using mental simulation for exploratory thought more generally. Moreover, although the way we choose which mental simulations to run is biased towards what is normally expected, we are able to modify our simulations to explore just outside the bounds of everyday life. For example, I can imagine riding an animal (which is not too far out of the ordinary), but gradually change the features of my mental simulation to make it unordinary (such as imagining riding a giant cat, or a flying alligator). The fact that our mental simulations tend to encode so much that is true about the world \citep{Gendler:1998} may be what makes them so perfect for creative thought: by tweaking them just by a small amount, we hit the sweet spot between novelty and utility that seems crucial for marking something as “creative” \citep{Ward:1994}. [\,\ldots ]}
\end{quote}

Through the use of queries, we have abstracted away the need to visually understand a scene in many cases and an oracle can focus on analyzing answers to the questions it has based on symbolic information. In the following, we highlight some of the possible mental simulation tasks that an oracle might induce.

\subsection{Sorting}

\begin{algorithm}[H]
	\caption{Sorting apples}\label{alg:Sorting}
	\begin{algorithmic}[1]
		
		\State \textit{apples} $\gets$ \textsc{Replay}(\textbf{origin all}, \textbf{filter} \symb{apple} \textbf{and} $\textit{time-stamp}>t_0$ 
		\Statex[7] \textbf{and omit} \textit{connections}, 
		\Statex[7] \textbf{map} (\symb{apple}.\attr{red}$<0.5\to$ \attr{reject}) 
		\Statex[7] \textbf{and} (\symb{apple}.\attr{red}$\geq 0.5\to$ \attr{prefer}), \Statex[7] \textbf{group by} \attr{prefer}, \attr{reject})
		
	\end{algorithmic}
\end{algorithm}

This query replays every recent apple symbol (after $t_0$) and maps the red attribute to whether to prefer or reject it. Finally, all apples are grouped and two collections (instead of graphs, as connections were filtered out) are produced, one with all preferred apples and one with all rejected apples. This could be a reasonable query at the supermarket for an agent that does not like green or discolored apples.

\subsection{Mapping the Environment}

For our next query, let $G$ be some uniform grid in space given as a parameter to the query with $G_{i,j}$ its individual grid points.

\begin{algorithm}[H]
	\caption{Navigating by region}\label{alg:Navigating}
	\begin{algorithmic}[1]
		
		\State \textit{map-graph} $\gets$ \textsc{Replay}(\textbf{origin} \textit{perceived-observations}, \textbf{filter} $\textit{time-stamp}>t_0$,
		\Statex[7] \textbf{map none}, \textbf{group by} $\min_{i,j} \lVert\textit{camera-position} - G_{i,j} \rVert$)
		
	\end{algorithmic}
\end{algorithm}

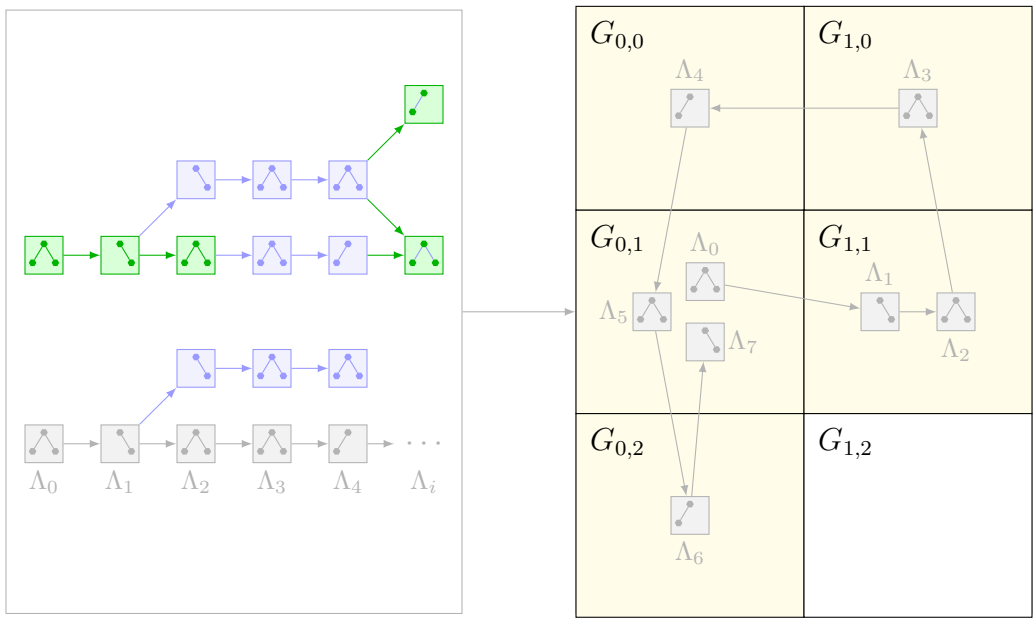
\begin{figure}[H]
	\centering
	\begin{adjustbox}{max width=1\textwidth}
		\begin{tikzpicture}
		
		%- Boxes
		
		\node (BEpisodic) [draw=black!30, minimum width=6cm, minimum height=8cm] at (0.4cm, 0.3cm) {};
		\node (BKnowledge) [draw=black!30, minimum width=6cm, minimum height=8cm] at (7.9cm, 0.3cm) {};
		
		%\node (BEx) at (0.4cm, 4.8cm) {Episodic Memory};
		%\node (BKx) at (7.9cm, 4.8cm) {Knowledge Graph-of-Graphs};
		
		\draw[tochild, black!30!white] (BEpisodic) -- (BKnowledge);

		%- Episodic Memory
		\node (E1) [fill=black!05!white, draw=black!30!white, minimum width=0.5cm, minimum height=0.5cm] at (-2.1cm, 0.3cm - 1.75cm) {};
		\node (E2) [fill=black!05!white, draw=black!30!white, minimum width=0.5cm, minimum height=0.5cm] at (-1.1cm, 0.3cm - 1.75cm) {};
		\node (E3) [fill=black!05!white, draw=black!30!white, minimum width=0.5cm, minimum height=0.5cm] at (-0.1cm, 0.3cm - 1.75cm) {};
		\node (E4) [fill=black!05!white, draw=black!30!white, minimum width=0.5cm, minimum height=0.5cm] at (0.9cm, 0.3cm - 1.75cm) {};
		\node (E5) [fill=black!05!white, draw=black!30!white, minimum width=0.5cm, minimum height=0.5cm] at (1.9cm, 0.3cm - 1.75cm) {};
		\node (E6) [text=black!30!white, minimum width=0.5cm, minimum height=0.5cm] at (2.9cm, 0.3cm - 1.75cm) {$\cdots$};
		
		\node[text=black!30!white] at (-2.1cm, -0.2cm - 1.75cm) {\footnotesize $\Lambda_0$};
		\node[text=black!30!white] at (-1.1cm, -0.2cm - 1.75cm) {\footnotesize $\Lambda_1$};
		\node[text=black!30!white] at (-0.1cm, -0.2cm - 1.75cm) {\footnotesize $\Lambda_2$};
		\node[text=black!30!white] at (0.9cm, -0.2cm - 1.75cm) {\footnotesize $\Lambda_3$};
		\node[text=black!30!white] at (1.9cm, -0.2cm - 1.75cm) {\footnotesize $\Lambda_4$};
		\node[text=black!30!white] at (2.9cm, -0.2cm - 1.75cm) {\footnotesize $\Lambda_i$};
		
		\draw[tochild, black!30!white] (E1) -- (E2);
		\draw[tochild, black!30!white] (E2) -- (E3);
		\draw[tochild, black!30!white] (E3) -- (E4);
		\draw[tochild, black!30!white] (E4) -- (E5);
		\draw[tochild, black!30!white] (E5) -- (E6);
		
		\node (E1Cap1) [regular polygon,regular polygon sides=6, fill=black!30!white, minimum width=1cm, minimum height=1cm, scale=0.1] at (-2.1cm, 0.45cm - 1.75cm) {};
		\node (E1Cap2) [regular polygon,regular polygon sides=6, fill=black!30!white, minimum width=1cm, minimum height=1cm, scale=0.1] at (-2.25cm, 0.2cm - 1.75cm) {};
		\node (E1Cap3) [regular polygon,regular polygon sides=6, fill=black!30!white, minimum width=1cm, minimum height=1cm, scale=0.1] at (-1.95cm, 0.2cm - 1.75cm) {};
		\draw[black!30!white, scale=0.1] (E1Cap1) -- (E1Cap2);
		\draw[black!30!white, scale=0.1] (E1Cap1) -- (E1Cap3);		
		
		\node (E2Cap1) [regular polygon,regular polygon sides=6, fill=black!30!white, minimum width=1cm, minimum height=1cm, scale=0.1] at (-1.1cm, 0.45cm - 1.75cm) {};
		\node (E2Cap3) [regular polygon,regular polygon sides=6, fill=black!30!white, minimum width=1cm, minimum height=1cm, scale=0.1] at (-0.95cm, 0.2cm - 1.75cm) {};
		\draw[black!30!white, scale=0.1] (E2Cap1) -- (E2Cap3);
		
		\node (E3Cap1) [regular polygon,regular polygon sides=6, fill=black!30!white, minimum width=1cm, minimum height=1cm, scale=0.1] at (-0.1cm, 0.45cm - 1.75cm) {};
		\node (E3Cap2) [regular polygon,regular polygon sides=6, fill=black!30!white, minimum width=1cm, minimum height=1cm, scale=0.1] at (-0.25cm, 0.2cm - 1.75cm) {};
		\node (E3Cap3) [regular polygon,regular polygon sides=6, fill=black!30!white, minimum width=1cm, minimum height=1cm, scale=0.1] at (0.05cm, 0.2cm - 1.75cm) {};
		\draw[black!30!white, scale=0.1] (E3Cap1) -- (E3Cap2);
		\draw[black!30!white, scale=0.1] (E3Cap1) -- (E3Cap3);
		
		\node (E4Cap1) [regular polygon,regular polygon sides=6, fill=black!30!white, minimum width=1cm, minimum height=1cm, scale=0.1] at (0.9cm, 0.45cm - 1.75cm) {};
		\node (E4Cap2) [regular polygon,regular polygon sides=6, fill=black!30!white, minimum width=1cm, minimum height=1cm, scale=0.1] at (0.75cm, 0.2cm - 1.75cm) {};
		\node (E4Cap3) [regular polygon,regular polygon sides=6, fill=black!30!white, minimum width=1cm, minimum height=1cm, scale=0.1] at (1.05cm, 0.2cm - 1.75cm) {};
		\draw[black!30!white, scale=0.1] (E4Cap1) -- (E4Cap2);
		\draw[black!30!white, scale=0.1] (E4Cap1) -- (E4Cap3);
		
		\node (E5Cap1) [regular polygon,regular polygon sides=6, fill=black!30!white, minimum width=1cm, minimum height=1cm, scale=0.1] at (1.9cm, 0.45cm - 1.75cm) {};
		\node (E5Cap2) [regular polygon,regular polygon sides=6, fill=black!30!white, minimum width=1cm, minimum height=1cm, scale=0.1] at (1.75cm, 0.2cm - 1.75cm) {};
		\draw[black!30!white, scale=0.1] (E5Cap1) -- (E5Cap2);
		
		%- Predicted Memory
		
		\node (P3) [fill=blue!05!white, draw=blue!40!white, minimum width=0.5cm, minimum height=0.5cm] at (-0.1cm, 1.3cm - 1.75cm) {};
		\node (P4) [fill=blue!05!white, draw=blue!40!white, minimum width=0.5cm, minimum height=0.5cm] at (0.9cm, 1.3cm - 1.75cm) {};
		\node (P5) [fill=blue!05!white, draw=blue!40!white, minimum width=0.5cm, minimum height=0.5cm] at (1.9cm, 1.3cm - 1.75cm) {};

		\node (P3Cap1) [regular polygon,regular polygon sides=6, fill=blue!40!white, minimum width=1cm, minimum height=1cm, scale=0.1] at (-0.1cm, 1.45cm - 1.75cm) {};
		\node (P3Cap3) [regular polygon,regular polygon sides=6, fill=blue!40!white, minimum width=1cm, minimum height=1cm, scale=0.1] at (0.05cm, 1.2cm - 1.75cm) {};
		\draw[blue!40!white, scale=0.1] (P3Cap1) -- (P3Cap3);
		
		\node (P4Cap1) [regular polygon,regular polygon sides=6, fill=blue!40!white, minimum width=1cm, minimum height=1cm, scale=0.1] at (0.9cm, 1.45cm - 1.75cm) {};
		\node (P4Cap2) [regular polygon,regular polygon sides=6, fill=blue!40!white, minimum width=1cm, minimum height=1cm, scale=0.1] at (0.75cm, 1.2cm - 1.75cm) {};
		\node (P4Cap3) [regular polygon,regular polygon sides=6, fill=blue!40!white, minimum width=1cm, minimum height=1cm, scale=0.1] at (1.05cm, 1.2cm - 1.75cm) {};
		\draw[blue!40!white, scale=0.1] (P4Cap1) -- (P4Cap2);
		\draw[blue!40!white, scale=0.1] (P4Cap1) -- (P4Cap3);
		
		\node (P5Cap1) [regular polygon,regular polygon sides=6, fill=blue!40!white, minimum width=1cm, minimum height=1cm, scale=0.1] at (1.9cm, 1.45cm - 1.75cm) {};
		\node (P5Cap2) [regular polygon,regular polygon sides=6, fill=blue!40!white, minimum width=1cm, minimum height=1cm, scale=0.1] at (1.75cm, 1.2cm - 1.75cm) {};
		\node (P5Cap3) [regular polygon,regular polygon sides=6, fill=blue!40!white, minimum width=1cm, minimum height=1cm, scale=0.1] at (2.05cm, 1.2cm - 1.75cm) {};
		\draw[blue!40!white, scale=0.1] (P5Cap1) -- (P5Cap2);
		\draw[blue!40!white, scale=0.1] (P5Cap1) -- (P5Cap3);
		
		\draw[tochild, blue!40!white] (E2) -- (P3);
		\draw[tochild, blue!40!white] (P3) -- (P4);
		\draw[tochild, blue!40!white] (P4) -- (P5);
		
		%- Fabricated Timeline
		\node (S1) [fill=green!15!white, draw=black!30!green, minimum width=0.5cm, minimum height=0.5cm] at (-2.1cm, 2.8cm - 1.75cm) {};
		\node (S2) [fill=green!15!white, draw=black!30!green, minimum width=0.5cm, minimum height=0.5cm] at (-1.1cm, 2.8cm - 1.75cm) {};
		\node (S3) [fill=green!15!white, draw=black!30!green, minimum width=0.5cm, minimum height=0.5cm] at (-0.1cm, 2.8cm - 1.75cm) {};
		\node (S4) [fill=blue!05!white, draw=blue!40!white, minimum width=0.5cm, minimum height=0.5cm] at (0.9cm, 2.8cm - 1.75cm) {};
		\node (S5) [fill=blue!05!white, draw=blue!40!white, minimum width=0.5cm, minimum height=0.5cm] at (1.9cm, 2.8cm - 1.75cm) {};
		\node (S6) [fill=green!15!white, draw=black!30!green, minimum width=0.5cm, minimum height=0.5cm] at (2.9cm, 2.8cm - 1.75cm) {};
		\node (Sx6) [fill=green!15!white, draw=black!30!green, minimum width=0.5cm, minimum height=0.5cm] at (2.9cm, 4.8cm - 1.75cm) {};
		
		\draw[tochild, black!30!green] (S1) -- (S2);
		\draw[tochild, black!30!green] (S2) -- (S3);
		\draw[tochild, blue!40!white] (S3) -- (S4);
		\draw[tochild, blue!40!white] (S4) -- (S5);
		\draw[tochild, black!30!green] (S5) -- (S6);
		
		\node (S1Cap1) [regular polygon,regular polygon sides=6, fill=black!30!green, minimum width=1cm, minimum height=1cm, scale=0.1] at (-2.1cm, 2.95cm - 1.75cm) {};
		\node (S1Cap2) [regular polygon,regular polygon sides=6, fill=black!30!green, minimum width=1cm, minimum height=1cm, scale=0.1] at (-2.25cm, 2.7cm - 1.75cm) {};
		\node (S1Cap3) [regular polygon,regular polygon sides=6, fill=black!30!green, minimum width=1cm, minimum height=1cm, scale=0.1] at (-1.95cm, 2.7cm - 1.75cm) {};
		\draw[black!30!green, scale=0.1] (S1Cap1) -- (S1Cap2);
		\draw[black!30!green, scale=0.1] (S1Cap1) -- (S1Cap3);		
		
		\node (S2Cap1) [regular polygon,regular polygon sides=6, fill=black!30!green, minimum width=1cm, minimum height=1cm, scale=0.1] at (-1.1cm, 2.95cm - 1.75cm) {};
		\node (S2Cap3) [regular polygon,regular polygon sides=6, fill=black!30!green, minimum width=1cm, minimum height=1cm, scale=0.1] at (-0.95cm, 2.7cm - 1.75cm) {};
		\draw[black!30!green, scale=0.1] (S2Cap1) -- (S2Cap3);
		
		\node (S3Cap1) [regular polygon,regular polygon sides=6, fill=black!30!green, minimum width=1cm, minimum height=1cm, scale=0.1] at (-0.1cm, 2.95cm - 1.75cm) {};
		\node (S3Cap2) [regular polygon,regular polygon sides=6, fill=black!30!green, minimum width=1cm, minimum height=1cm, scale=0.1] at (-0.25cm, 2.7cm - 1.75cm) {};
		\node (S3Cap3) [regular polygon,regular polygon sides=6, fill=black!30!green, minimum width=1cm, minimum height=1cm, scale=0.1] at (0.05cm, 2.7cm - 1.75cm) {};
		\draw[black!30!green, scale=0.1] (S3Cap1) -- (S3Cap2);
		\draw[black!30!green, scale=0.1] (S3Cap1) -- (S3Cap3);
		
		\node (S4Cap1) [regular polygon,regular polygon sides=6, fill=blue!40!white, minimum width=1cm, minimum height=1cm, scale=0.1] at (0.9cm, 2.95cm - 1.75cm) {};
		\node (S4Cap2) [regular polygon,regular polygon sides=6, fill=blue!40!white, minimum width=1cm, minimum height=1cm, scale=0.1] at (0.75cm, 2.7cm - 1.75cm) {};
		\node (S4Cap3) [regular polygon,regular polygon sides=6, fill=blue!40!white, minimum width=1cm, minimum height=1cm, scale=0.1] at (1.05cm, 2.7cm - 1.75cm) {};
		\draw[blue!40!white, scale=0.1] (S4Cap1) -- (S4Cap2);
		\draw[blue!40!white, scale=0.1] (S4Cap1) -- (S4Cap3);
		
		\node (S5Cap1) [regular polygon,regular polygon sides=6, fill=blue!40!white, minimum width=1cm, minimum height=1cm, scale=0.1] at (1.9cm, 2.95cm - 1.75cm) {};
		\node (S5Cap2) [regular polygon,regular polygon sides=6, fill=blue!40!white, minimum width=1cm, minimum height=1cm, scale=0.1] at (1.75cm, 2.7cm - 1.75cm) {};
		\draw[blue!40!white, scale=0.1] (S5Cap1) -- (S5Cap2);
		
		\node (S6Cap1) [regular polygon,regular polygon sides=6, fill=black!30!green, minimum width=1cm, minimum height=1cm, scale=0.1] at (2.9cm, 2.95cm - 1.75cm) {};
		\node (S6Cap2) [regular polygon,regular polygon sides=6, fill=black!30!green, minimum width=1cm, minimum height=1cm, scale=0.1] at (2.75cm, 2.7cm - 1.75cm) {};
		\node (S6Cap3) [regular polygon,regular polygon sides=6, fill=black!30!green, minimum width=1cm, minimum height=1cm, scale=0.1] at (3.05cm, 2.7cm - 1.75cm) {};
		\draw[blue!40!white, scale=0.1] (S6Cap1) -- (S6Cap2);
		\draw[blue!40!white, scale=0.1] (S6Cap1) -- (S6Cap3);
		
		\node (Sx6Cap1) [regular polygon,regular polygon sides=6, fill=black!30!green, minimum width=1cm, minimum height=1cm, scale=0.1] at (2.9cm, 4.95cm - 1.75cm) {};
		\node (Sx6Cap2) [regular polygon,regular polygon sides=6, fill=black!30!green, minimum width=1cm, minimum height=1cm, scale=0.1] at (2.75cm, 4.7cm - 1.75cm) {};
		\draw[blue!40!white, scale=0.1] (Sx6Cap1) -- (Sx6Cap2);
		
		%- Fabricated Prediction of Fabricated branch
		
		\node (PS3) [fill=blue!05!white, draw=blue!40!white, minimum width=0.5cm, minimum height=0.5cm] at (-0.1cm, 3.8cm - 1.75cm) {};
		\node (PS4) [fill=blue!05!white, draw=blue!40!white, minimum width=0.5cm, minimum height=0.5cm] at (0.9cm, 3.8cm - 1.75cm) {};
		\node (PS5) [fill=blue!05!white, draw=blue!40!white, minimum width=0.5cm, minimum height=0.5cm] at (1.9cm, 3.8cm - 1.75cm) {};

		\node (PS3Cap1) [regular polygon,regular polygon sides=6, fill=blue!40!white, minimum width=1cm, minimum height=1cm, scale=0.1] at (-0.1cm, 3.95cm - 1.75cm) {};
		\node (PS3Cap3) [regular polygon,regular polygon sides=6, fill=blue!40!white, minimum width=1cm, minimum height=1cm, scale=0.1] at (0.05cm, 3.7cm - 1.75cm) {};
		\draw[blue!40!white, scale=0.1] (PS3Cap1) -- (PS3Cap3);
		
		\node (PS4Cap1) [regular polygon,regular polygon sides=6, fill=blue!40!white, minimum width=1cm, minimum height=1cm, scale=0.1] at (0.9cm, 3.95cm - 1.75cm) {};
		\node (PS4Cap2) [regular polygon,regular polygon sides=6, fill=blue!40!white, minimum width=1cm, minimum height=1cm, scale=0.1] at (0.75cm, 3.7cm - 1.75cm) {};
		\node (PS4Cap3) [regular polygon,regular polygon sides=6, fill=blue!40!white, minimum width=1cm, minimum height=1cm, scale=0.1] at (1.05cm, 3.7cm - 1.75cm) {};
		\draw[blue!40!white, scale=0.1] (PS4Cap1) -- (PS4Cap2);
		\draw[blue!40!white, scale=0.1] (PS4Cap1) -- (PS4Cap3);
		
		\node (PS5Cap1) [regular polygon,regular polygon sides=6, fill=blue!40!white, minimum width=1cm, minimum height=1cm, scale=0.1] at (1.9cm, 3.95cm - 1.75cm) {};
		\node (PS5Cap2) [regular polygon,regular polygon sides=6, fill=blue!40!white, minimum width=1cm, minimum height=1cm, scale=0.1] at (1.75cm, 3.7cm - 1.75cm) {};
		\node (PS5Cap3) [regular polygon,regular polygon sides=6, fill=blue!40!white, minimum width=1cm, minimum height=1cm, scale=0.1] at (2.05cm, 3.7cm - 1.75cm) {};
		\draw[blue!40!white, scale=0.1] (PS5Cap1) -- (PS5Cap2);
		\draw[blue!40!white, scale=0.1] (PS5Cap1) -- (PS5Cap3);
		
		\draw[tochild, blue!40!white] (S2) -- (PS3);
		\draw[tochild, blue!40!white] (PS3) -- (PS4);
		\draw[tochild, blue!40!white] (PS4) -- (PS5);
		\draw[tochild, black!30!green] (PS5) -- (S6);
		\draw[tochild, black!30!green] (PS5) -- (Sx6);
		
		%- Knowledge Bubbles
		
		\node (R1) [fill=white!90!yellow, draw=black, minimum width=3cm, minimum height=2.7cm] at (6.4cm, 0.3cm) {};
		\node (R2) [fill=white!90!yellow, draw=black, minimum width=3cm, minimum height=2.7cm] at (9.4cm, 0.3cm) {};
		\node (R3) [fill=white!90!yellow, draw=black, minimum width=3cm, minimum height=2.7cm] at (9.4cm, 3.0cm) {};
		\node (R4) [fill=white!90!yellow, draw=black, minimum width=3cm, minimum height=2.7cm] at (6.4cm, 3.0cm) {};
		\node (R5) [fill=white!90!yellow, draw=black, minimum width=3cm, minimum height=2.7cm] at (6.4cm, -2.4cm) {};
		\node (R6) [fill=white, draw=black, minimum width=3cm, minimum height=2.7cm] at (9.4cm, -2.4cm) {};
		
		\node (G1) [text=black] at (5.45cm, 3.95cm) {$G_{0,0}$};
		\node (G2) [text=black] at (8.45cm, 3.95cm) {$G_{1,0}$};
		\node (G3) [text=black] at (5.45cm, 1.25cm) {$G_{0,1}$};
		\node (G4) [text=black] at (8.45cm, 1.25cm) {$G_{1,1}$};
		\node (G5) [text=black] at (5.45cm, -1.45cm) {$G_{0,2}$};
		\node (G6) [text=black] at (8.45cm, -1.45cm) {$G_{1,2}$};

		%- Knowledge
		\node (K1) [fill=black!05!white, draw=black!30!white, minimum width=0.5cm, minimum height=0.5cm] at (5.9cm, 0.3cm) {};
		\node (K2) [fill=black!05!white, draw=black!30!white, minimum width=0.5cm, minimum height=0.5cm] at (6.6cm, 0.7cm) {};
		\node (K3) [fill=black!05!white, draw=black!30!white, minimum width=0.5cm, minimum height=0.5cm] at (6.4cm, 3.0cm) {};
		\node (K4) [fill=black!05!white, draw=black!30!white, minimum width=0.5cm, minimum height=0.5cm] at (8.9cm, 0.3cm) {};
		\node (K5) [fill=black!05!white, draw=black!30!white, minimum width=0.5cm, minimum height=0.5cm] at (9.9cm, 0.3cm) {};
		\node (K6) [fill=black!05!white, draw=black!30!white, minimum width=0.5cm, minimum height=0.5cm] at (9.4cm, 3.0cm) {};
		\node (K7) [fill=black!05!white, draw=black!30!white, minimum width=0.5cm, minimum height=0.5cm] at (6.4cm, -2.4cm) {};
		\node (K8) [fill=black!05!white, draw=black!30!white, minimum width=0.5cm, minimum height=0.5cm] at (6.6cm, -0.1cm) {};
		
		\draw[tochild, black!30!white] (K2) -- (K4);
		\draw[tochild, black!30!white] (K4) -- (K5);
		\draw[tochild, black!30!white] (K5) -- (K6);
		\draw[tochild, black!30!white] (K6) -- (K3);
		\draw[tochild, black!30!white] (K3) -- (K1);
		\draw[tochild, black!30!white] (K1) -- (K7);
		\draw[tochild, black!30!white] (K7) -- (K8);

		\node[text=black!30!white] at (6.6cm, 0.7cm + 0.5cm) {\footnotesize $\Lambda_0$};
		\node[text=black!30!white] at (8.9cm, 0.3cm + 0.5cm) {\footnotesize $\Lambda_1$};
		\node[text=black!30!white] at (9.9cm, 0.3cm - 0.5cm) {\footnotesize $\Lambda_2$};
		\node[text=black!30!white] at (9.4cm, 3.0cm + 0.5cm) {\footnotesize $\Lambda_3$};
		\node[text=black!30!white] at (6.4cm, 3.0cm + 0.5cm) {\footnotesize $\Lambda_4$};
		\node[text=black!30!white] at (5.9cm - 0.5cm, 0.3cm) {\footnotesize $\Lambda_5$};
		\node[text=black!30!white] at (6.4cm, -2.4cm - 0.5cm) {\footnotesize $\Lambda_6$};
		\node[text=black!30!white] at (6.6cm + 0.5cm, -0.1cm) {\footnotesize $\Lambda_7$};

		\node (K1Cap1) [regular polygon,regular polygon sides=6, fill=black!30!white, minimum width=1cm, minimum height=1cm, scale=0.1] at (5.9cm, 0.45cm) {};
		\node (K1Cap2) [regular polygon,regular polygon sides=6, fill=black!30!white, minimum width=1cm, minimum height=1cm, scale=0.1] at (5.75cm, 0.2cm) {};
		\node (K1Cap3) [regular polygon,regular polygon sides=6, fill=black!30!white, minimum width=1cm, minimum height=1cm, scale=0.1] at (6.05cm, 0.2cm) {};
		\draw[black!30!white, scale=0.1] (K1Cap1) -- (K1Cap2);
		\draw[black!30!white, scale=0.1] (K1Cap1) -- (K1Cap3);		
		
		\node (K2Cap1) [regular polygon,regular polygon sides=6, fill=black!30!white, minimum width=1cm, minimum height=1cm, scale=0.1] at (6.6cm, 0.85cm) {};
		\node (K2Cap2) [regular polygon,regular polygon sides=6, fill=black!30!white, minimum width=1cm, minimum height=1cm, scale=0.1] at (6.45cm, 0.6cm) {};
		\node (K2Cap3) [regular polygon,regular polygon sides=6, fill=black!30!white, minimum width=1cm, minimum height=1cm, scale=0.1] at (6.75cm, 0.6cm) {};
		\draw[black!30!white, scale=0.1] (K2Cap1) -- (K2Cap2);
		\draw[black!30!white, scale=0.1] (K2Cap1) -- (K2Cap3);
		
		\node (K3Cap1) [regular polygon,regular polygon sides=6, fill=black!30!white, minimum width=1cm, minimum height=1cm, scale=0.1] at (6.4cm, 3.15cm) {};
		\node (K3Cap2) [regular polygon,regular polygon sides=6, fill=black!30!white, minimum width=1cm, minimum height=1cm, scale=0.1] at (6.25cm, 2.9cm) {};
		\draw[black!30!white, scale=0.1] (K3Cap1) -- (K3Cap2);
		
		\node (K4Cap1) [regular polygon,regular polygon sides=6, fill=black!30!white, minimum width=1cm, minimum height=1cm, scale=0.1] at (8.9cm, 0.45cm) {};
		\node (K4Cap3) [regular polygon,regular polygon sides=6, fill=black!30!white, minimum width=1cm, minimum height=1cm, scale=0.1] at (9.05cm, 0.2cm) {};
		\draw[black!30!white, scale=0.1] (K4Cap1) -- (K4Cap3);
		
		\node (K5Cap1) [regular polygon,regular polygon sides=6, fill=black!30!white, minimum width=1cm, minimum height=1cm, scale=0.1] at (9.9cm, 0.45cm) {};
		\node (K5Cap2) [regular polygon,regular polygon sides=6, fill=black!30!white, minimum width=1cm, minimum height=1cm, scale=0.1] at (9.75cm, 0.2cm) {};
		\node (K5Cap3) [regular polygon,regular polygon sides=6, fill=black!30!white, minimum width=1cm, minimum height=1cm, scale=0.1] at (10.05cm, 0.2cm) {};
		\draw[black!30!white, scale=0.1] (K5Cap1) -- (K5Cap2);
		\draw[black!30!white, scale=0.1] (K5Cap1) -- (K5Cap3);
		
		\node (K6Cap1) [regular polygon,regular polygon sides=6, fill=black!30!white, minimum width=1cm, minimum height=1cm, scale=0.1] at (9.4cm, 3.15cm) {};
		\node (K6Cap2) [regular polygon,regular polygon sides=6, fill=black!30!white, minimum width=1cm, minimum height=1cm, scale=0.1] at (9.25cm, 2.9cm) {};
		\node (K6Cap3) [regular polygon,regular polygon sides=6, fill=black!30!white, minimum width=1cm, minimum height=1cm, scale=0.1] at (9.55cm, 2.9cm) {};
		\draw[black!30!white, scale=0.1] (K6Cap1) -- (K6Cap2);
		\draw[black!30!white, scale=0.1] (K6Cap1) -- (K6Cap3);
		
		\node (K7Cap1) [regular polygon,regular polygon sides=6, fill=black!30!white, minimum width=1cm, minimum height=1cm, scale=0.1] at (6.4cm, -2.25cm) {};
		\node (K7Cap2) [regular polygon,regular polygon sides=6, fill=black!30!white, minimum width=1cm, minimum height=1cm, scale=0.1] at (6.25cm, -2.5cm) {};
		\draw[black!30!white, scale=0.1] (K7Cap1) -- (K7Cap2);
		
		\node (K8Cap1) [regular polygon,regular polygon sides=6, fill=black!30!white, minimum width=1cm, minimum height=1cm, scale=0.1] at (6.6cm, 0.05cm) {};
		\node (K8Cap3) [regular polygon,regular polygon sides=6, fill=black!30!white, minimum width=1cm, minimum height=1cm, scale=0.1] at (6.75cm, -0.2cm) {};
		\draw[black!30!white, scale=0.1] (K8Cap1) -- (K8Cap3);
		
		\end{tikzpicture}
	\end{adjustbox}
	\caption{Transformation of episodic memory using Algorithm \ref{alg:Navigating} to form a graph-of-graphs that represents the spatial map of the environment (size and distances not to scale), by only clustering perceived observations (gray) with a camera position similar to the grid points $G_{0,0}, \cdots, G_{1,2}$ (filled grid cells in yellow, empty grid cell in white).} \label{fig:SpatialMapping}
\end{figure}

In Algorithm \ref{alg:Navigating} we cluster all observations that are close to each of these grid points (\cf Figure \ref{fig:SpatialMapping}). As all connections between the networks are preserved, the graph-of-graphs this query produces is essentially a navigation graph of walkable space \citep{Werner:2014} explored since $t_0$. We could extend this map purely based on "imagination" by performing predict-actions and including the predicted states. The map could be further improved by assuming that the aggregation of the connections between the nodes in the final graph-of-graphs represents the certainty that this route actually exists. Using this mental map, the oracle may then perform graph traversal to find the shortest route to some location based on the knowledge it has gained. Note that Algorithm \ref{alg:Navigating} is overly simplified and does not take more complex effects into consideration, such as drift when it comes to loop closing.

\subsection{Will They Collide?}

Given a starting observation (\textit{ball-obs}) of two moving balls (\symbind{ball}{A} and \symbind{ball}{B}), we may ask ourselves, if they will collide in the next second? Employing the distance function used in intuitive physics (Equation \ref{eq:continue3}), we incrementally predict the next time step of the two balls flying towards each other and check if they collide. This process is highlighted in Algorithm \ref{alg:DoTheyCollide}.

\begin{algorithm}[H]
	\caption{Will they collide?}\label{alg:DoTheyCollide}
	\begin{algorithmic}[1]
		\State \textit{step} $\gets$ 0
		\Repeat
		
		\State \textit{ball-obs} $\gets$ \textsc{Predict}(\textbf{origin} \textit{ball-obs}, \textbf{duration} \SI{5}{\milli\second}, \textbf{force none})
		
		\State \textit{distance} $\gets$ \textsc{Replay}(\textbf{origin} \textit{ball-obs}, \textbf{filter} \symbind{ball}{A} \textbf{or} \symbind{ball}{B},
		\Statex[7] \textbf{map} d(\symbind{ball}{A},\symbind{ball}{B}), \textbf{group by none})
		
		\State \textit{step} $\gets$ \textit{step} + 1
		
		\Until{$distance < \epsilon$ \textbf{or} \textit{step} $\geq$ 20}
	\end{algorithmic}
\end{algorithm}

\subsection{Would it Fit?}

Another common mental simulation task is to check if an \symb{object} fits inside some \symb{container}. We may do this by fabricating a scene with both objects in it and placing one inside the other. The distance function will then allow us to perform a simple check if the object is colliding, \ie, doesn't fit (\cf Algorithm \ref{alg:WouldItFit}).

\begin{algorithm}[H]
	\caption{Would it fit?}\label{alg:WouldItFit}
	\begin{algorithmic}[1]		
		
		\State \textit{fabricate-obs} $\gets$ \textsc{Fabricate}(\textbf{origin none}, \textbf{change add} \symb{object} \textbf{and} \symb{container})
		
		\State \textit{distance} $\gets$ \textsc{Replay}(\textbf{origin} \textit{fabricate-obs}, \textbf{filter} \symb{container} \textbf{or} \symb{object},
		\Statex[7] \textbf{map} d(\symb{container},\symb{object}), \textbf{group by none})		
		
		\If {\textit{distance} $> \epsilon$}
			\State \textsc{DoesFit}()
		\Else
			\State \textsc{DoesNotFit}()
		\EndIf
		
	\end{algorithmic}
\end{algorithm}

\subsection{Style Transfer}

Thus far, our example queries have focused on extracting knowledge. We are, however, also able to input knowledge into our framework using queries. This is important, as an oracle or a planning agent may come to realize something through other means, which it then wishes to share with the framework.

We consider the following situation: a \symb{stool} with the attribute \attr{modern} produces \symb{leg} symbols with a very \attr{wooden} appearance and a \symb{top} with a \attr{metallic} appearance. The oracle has learned that this modern style is not unique to stools and also applies to lamps. So far, however, the capsule network has only encountered \symb{lamp} symbols that produce a \attr{wooden} \symb{stand} and \symb{shade} or a \attr{metallic} \symb{stand} and \symb{shade}, but never a mixture of both. The oracle may now impose the knowledge of a \attr{modern} \symb{lamp} through a query, such as in Algorithm \ref{alg:StyleTransfer}.

\begin{algorithm}[H]
	\caption{Style Transfer}\label{alg:StyleTransfer}
	\begin{algorithmic}[1]

		\State \textit{all-modern-stools} $\gets$ \textsc{Replay}(\textbf{origin none}, \textbf{filter} \symb{stool} \textbf{and} \attr{modern} $>0.8$,
		\Statex[8] \textbf{map none}, \textbf{group by none})	
		
		\ForEach {\textit{modern-stool} $\in$ \textit{all-modern-stools}}
		\State \dots $\gets$ \textsc{Fabricate}(\textbf{origin} \symb{lamp}, \textbf{change copy attribute values} \textit{modern-stool})
		\EndFor
		
	\end{algorithmic}
\end{algorithm}

We are in a sense implicitly copying the configuration space topology of one capsule into another by forcing the \symb{lamp} capsule to simulate itself with the attributes of a \attr{modern} \symb{stool}. This copy process may also be applied to entire parts. We may, for example, simulate a lamp with its \symb{stand} replaced by a \symb{leg} trough fabrication. Here, not only is the style transferred, but also its configuration.

The truly interesting aspect to this is, that we may automate such a query and execute it for a wide range of capsules each time a new style is learned. This would mean an automated style transfer, but with the ability to control exactly the classes of objects it applies to.

\subsection{Moving and Fixing Errors}

For a robotics application, it is useful that a robot is able to simulate an action before executing it, in order to apply the correct movements with adequate forces. This is especially true if it needs to interact with its environment, such as moving an object without damaging it.

In our example, an arm needs to be rotated to a certain \textit{target-angle}. Our framework has seen this arm in action before and learned some of the physics of it. Using this knowledge, the robot is able to perform a mental simulation of the intended action before executing it and assess an error it made afterwards for a possible correction (\cf Algorithm \ref{alg:FixingErrors}).

\begin{algorithm}[H]
	\caption{Moving and Fixing Errors}\label{alg:FixingErrors}
	\begin{algorithmic}[1]	
		
		\State \textit{final-torque} $\gets$ \SI{0}{\newton.\meter}
		
		\Repeat
		
		\State \textit{final-torque} $\gets$ \textit{final-torque} + \SI{0.1}{\newton.\meter}
		
		\State \textit{predicted-obs} $\gets$ \textsc{Predict}(\textbf{origin} \textit{current-obs}, \textbf{duration} \SI{5}{\milli\second}, 
		\Statex[8]\textbf{force apply} \textit{final-torque} \textbf{to} \symb{arm})
		
		\Until{\textit{predicted-obs}.\symb{arm}.\attr{angle} $\geq$ target-angle}
		
		\State \textsc{PerformAction}(\textit{final-torque})
		
		\State \textit{angle} $\gets$ \textsc{Replay}(\textbf{origin none}, \textbf{filter newest},
		\Statex[6] \textbf{map} \symb{arm}.\attr{angle}, \textbf{group by none})
		
		\State \textit{angle-error} $\gets$ \textit{target-angle} - \textit{angle}
		
	\end{algorithmic}
\end{algorithm}

\subsection{Game Engine}

As final application that combines all aspects of the framework, we investigate its use as a game engine. Based purely on the observations of its environment, it can simulate both the visual appearance, in addition to the physics, as shown in Algorithm \ref{alg:GameEngine}.

\begin{algorithm}[H]
	\caption{Basic game engine}\label{alg:GameEngine}
	\begin{algorithmic}[1]		
		
		\State game-environment $\gets$ \textsc{Fabricate}(\textbf{origin} \textit{real-environment}, \textbf{change none})
				
		\While{\textit{game-is-running} \textbf{is} \textit{true}}

		\State \textit{game-camera}, \textit{modify-objects}, \textit{forces} $\gets$ \textsc{GameActions}(\textit{game-environment})
		
		\State \textit{game-environment} $\gets$ \textsc{Fabricate}(\textbf{origin} \textit{game-environment},
		\Statex[11]\textbf{change apply} \textit{modify-objects})
				
		\State \textit{game-environment} $\gets$ \textsc{Predict}(\textbf{origin} \textit{game-environment}, \textbf{duration} \SI{5}{\milli\second}, 
		\Statex[11]\textbf{force apply} \textit{forces})

		\State \dots, \textit{frame} $\gets$ \textsc{Replay}(\textbf{origin} \textit{game-environment},
		\Statex[8]\textbf{filter} $arg\,min_{\textit{obs}} \lVert \textit{obs}.\textit{camera} - \textit{game-camera} \rVert$,
		\Statex[8] \textbf{map \textit{obs}.\textit{camera} $\gets$ \textit{game-camera}}, \textbf{group by none})
		
		\State \textsc{Present}(\textit{frame})	
		
		\EndWhile
		
	\end{algorithmic}
\end{algorithm}

We begin by defining a subset of all observations as the \textit{game-environment} to improve performance. For example, we may restrict the environment to the interior of some apartment. While the game is running, the player performs actions, such as moving around, throwing, adding or removing an object. What actions are performed is given by the \textsc{GameActions} function. Using these, the game environment is altered accordingly and the next time step is predicted. To simulate the player's movement, the observation from \textit{game-environment} is selected whose camera position and rotation fit as good as possible and is used to provide the new parameters for rendering. Finally, the resulting frame is presented to the player (\textsc{Present}) and the next frame of the game loop starts.

\section{Implementation and Results}

We have implemented the inverse game engine as outlined in Algorithm \ref{alg:GameEngine} using our VividNet framework and Python mimicking the presented querying language. 

\begin{figure}[H]
	\centering
	\begin{adjustbox}{max width=1.0\textwidth}
		\begin{tikzpicture}
		\node[inner sep=0pt] at (0,0) {\includegraphics[width=0.7\textwidth]{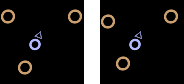}};
		\end{tikzpicture}
	\end{adjustbox}
	\caption{The two frames of slightly moving asteroids observed before starting the simulation. Asteroids are shown in brown and the shuttle in blue.} \label{fig:simStart}
\end{figure}

Our test is conducted by training our framework on an Asteroids-like environment, letting it observe a limited amount of interactions of asteroids with the space shuttle and other asteroids. We then show it two new frames, in which the space shuttle is not moving, but the asteroids are, as shown in Figure \ref{fig:simStart}. These new frames are sufficient for the framework to infer all the required information to begin simulation starting at that point. 

We produce three different simulation scenarios with 14 frames each. The oracle's input only controls the movement of the shuttle and nothing else. The shuttle's physics and all asteroids are simulated based solely on our VividNet framework and its past observations. All output is rendered using the feed-backward process of VividNet.

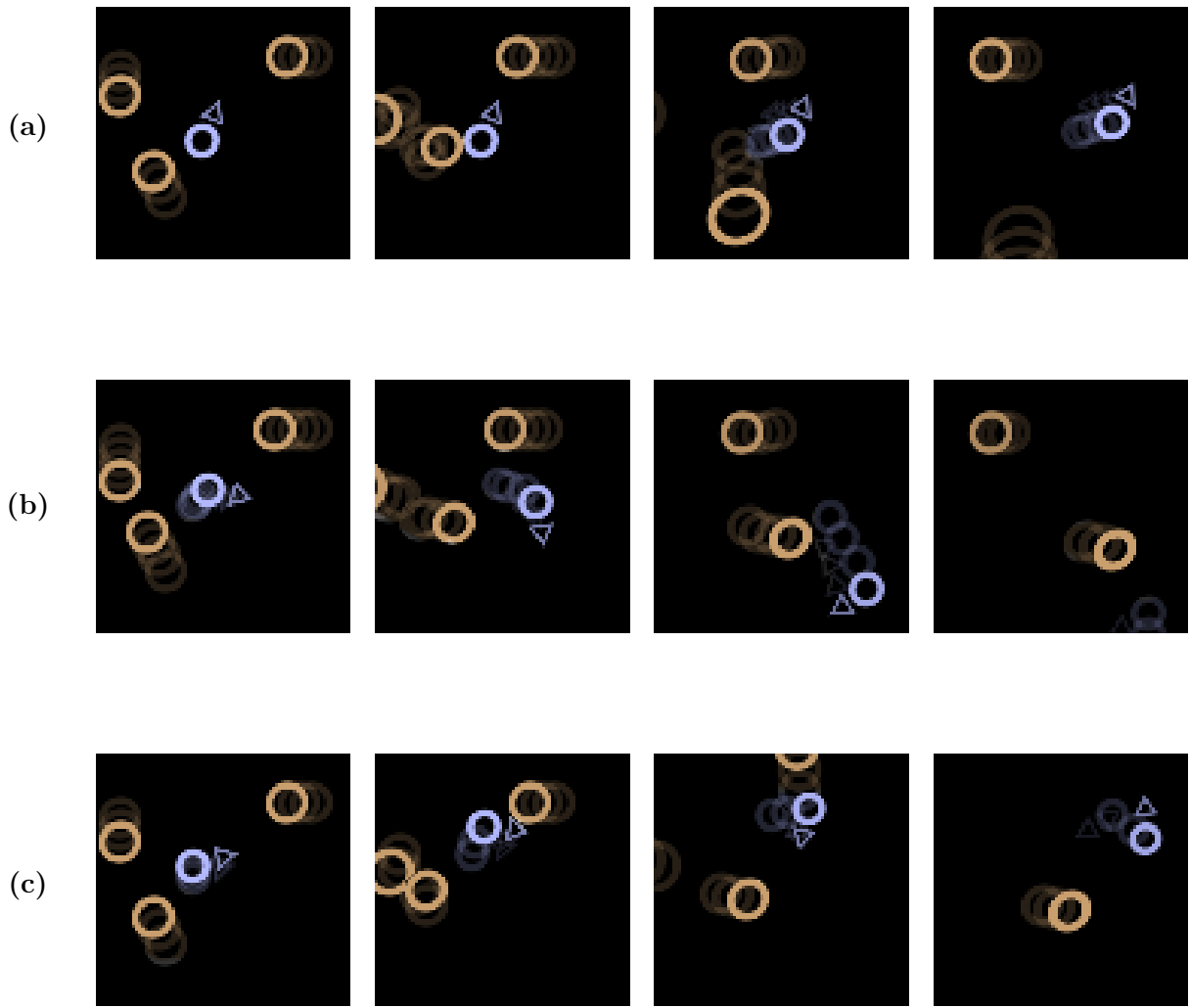
\begin{figure}
	\centering
	\begin{adjustbox}{max width=1.0\textwidth}
		\begin{tikzpicture}
		\node[inner sep=0pt] at (0,0) {\includegraphics[width=1.0\textwidth]{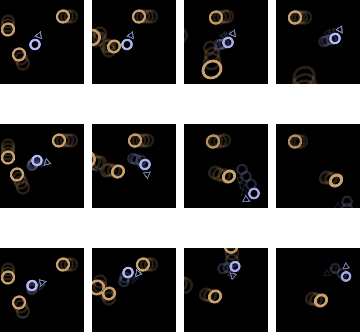}};
		
		\node at (-9.0,5.6) {\textbf{(a)}};
		\node at (-9.0,0.0) {\textbf{(b)}};
		\node at (-9.0,-5.6) {\textbf{(c)}};
		\end{tikzpicture}
	\end{adjustbox}
	\caption{Three possible simulation scenarios with different behavior of the shuttle (blue) and its interaction with moving asteroids (brown). \textbf{(a)} shows the shuttle not moving, but eventually being hit and moved by an asteroid. \textbf{(b)} shows the shuttle moving out of the way of all asteroids. \textbf{(c)} shows the shuttle deliberately hitting an asteroid to move it. Trails of past frames are shown.} \label{fig:simResults}
\end{figure}

The first simulation is a simple prediction of what happens if nothing is changed, \ie, the asteroids move on their predicted paths, but the shuttle does not move (\cf Figure \ref{fig:simResults} \textbf{(a)}). Here we see the asteroids mutually bouncing off and eventually colliding with the ship, sending it to a new trajectory. One of the asteroids collides two times in this prediction causing this asteroid to grow in size because of the accumulated numerical errors of our implementation of the interaction network's predictor $\phi_R$.

We repeat the same simulation, but this time take control of the shuttle and move it out of the way of all of the asteroids to avoid collision (\cf Figure \ref{fig:simResults} \textbf{(b)}). All asteroids fly on their expected paths, ignoring the shuttle.

In a final experiment, the shuttle deliberately strikes an asteroid to push it off its flight path (\cf Figure \ref{fig:simResults} \textbf{(c)}). The trajectories of the shuttle and asteroid after the collision correspond to the expected outcome.

\clearpage

\chapter{Discussion}\label{sec:Discuss}\thispagestyle{empty}

Our approach differs too much from current classification methods for a reasonable direct numerical comparison. The neural-symbolic capsule network expresses confidence, but has no notion of accuracy, as any inaccuracies are remedied by the meta-learning pipeline over time. A comparison would be to a subjective configuration of the capsule network at that point, eliminating the benefits of lifelong meta-learning.

Because of this, we abstain from a numerical comparison and instead discuss strengths and limitations of our method and compare them to the classical utilization of neural networks.

\section{Versatility}

We begin by giving a short definition of how we use the terms \index{Interpretability}interpretable and \index{Explainability}explainable, analogous to \citep{Gilpin:2018}:
\begin{itemize}
	\item Interpretability is to understand how a result is reached and the mechanisms employed on a mathematical or procedural level.
	\item Explainability allows a model's mechanisms to be explained to and understood by a non-expert.
\end{itemize}

Even though both terms are often used interchangeably in current literature, the subtle difference is important for our discussion. Often an explainable model is interpretable, but the opposite is seldom true.

\subsection{Modularity}

To avoid the need to constantly redesign and retrain models, the concept of \index{Modularity}modularity offers a way to simply combine existing parts in a novel way. As we discussed in previous sections, this is one of the key features of our neural-symbolic capsule's meta-learning algorithm. All attributes, routes and capsules can be recombined and allow for a dynamic structure.

For classical neural networks, modularity strongly hinges on interpretability and explainability. An example for interpretable modularity is \index{Neural Architecture Search}neural architecture search (NAS) \citep{Elsken:2019}. In NAS, the idea is to have building blocks consisting of smaller neural networks (modules) and to use algorithms, such as evolutionary computation, to find the best architecture for a given use-case. While a NAS can be interpretable, the resulting architecture, however, is seldom explainable. Thus, while highly modular, we are so far unable to infer useful conclusions from the final design and need to repeat the costly process should the conditions change.

Explainable modularity for neural networks would eliminate these downsides. However, we are not aware of any classical architectures that are both modular and explainable. Current efforts have instead been to find explainable architectures and to use \index{Transfer Learning}transfer learning \citep{Tan:2018} to take advantage of it in other domains. 

Another challenge for the explainability of neural networks is the difficulty to pinpoint the neurons or region of neurons responsible for a specific result. This can be, for example, studied using \index{Saliency Map}saliency maps \citep{Zeiler:2014, Bach:2015, Bojarski:2017}, class activation maps \citep{Zhou:2016} or by identifying critical routing paths in the network \citep{Wang:2018}. The results are often diffuse and span large regions of the network with only a small contribution from each neuron. This is not necessarily a bad property, because it is argued that this is what makes these networks so powerful by not focusing on exact features and instead also incorporating the context around these features. However, it does make it harder to explain in a reasonable matter why this is happening. With regard to this sort of explainability, \citet{Jaeger:2014} showed that it is indeed possible to extract symbolic meaning from the inner workings of recurrent neural networks, but so far this has not been transferred to convolutional neural networks. 

In contrast, neural-symbolic capsules allow to precisely identify which region was influenced by which capsule, as the saliency maps (\ie, graphically rendering an object and using its silhouette) are exact. Together with the lexical interpretation of each parameter, we can explain the results in a more complete manner. Further, as each capsule is trained in isolation, we are able to add, remove, modify or move any capsule or attribute in the network, resulting in full modularity with an explainable outcome.

\subsection{Adaptability}

A neural network is believed to generalize well, if it has reached a \index{Locally Flat Minimum}locally flat minimum \citep{Hinton:1993, Hochreiter:1996}, where the information in the weights is limited and small perturbations barely influence the loss. New studies \citep{Wu3:2017, Novak:2018, Li:2018} have shown that this intuition goes in the right direction. However, computing the \index{Loss Landscape,}loss landscape for a large model and a wide variety of data points is computationally very expensive. In most cases, such a check is performed only after a model has been found using algorithms (such as NAS) to confirm its feasibility and ability to generalize.

By splitting the monolithic approach to neural networks into smaller networks for individual capsules, an analysis of the loss landscape becomes computationally much more feasible. Furthermore, as it is possible that all capsules use the same architecture internally, we only need to perform such a check on a single or some capsules. Through the modularity discussed above, we are also able to replace any neural network used inside the capsules at any point in the future with one that generalizes better.

Yet, we note that this only covers the generalization of individual capsules and not the capsule network as a whole. The network itself is very limited in its ability to generalize and delegates this task to the meta-learning pipeline. The meta-learning pipeline, however, is able to generalize well, as it simply attempts to learn the features of anything new.

By construction, our neural-symbolic capsules are able to learn the representation of any object or scene if an adequate set of primitive capsules was chosen. This is especially true if an atomic set of primitives was selected (for example edges combined with discrete cosine transforms) which also is suitable for organic surfaces. For the latter, we refer to the work by \citet{Omran:2018}, which use an architecture similar to that of the routes found in our primitive capsules to perform pose and shape estimation of humans. As such, their implementation could be directly used as a route for a \symb{human} primitive capsule. Further, again taking advantage of modularity, it is always possible to add new primitive capsules at a later point by hand, expanding the set of learnable objects. 

We also saw that through querying we are able to perform \index{Style Transfer}style transfers (\cf Section 7.2.5) and noted how this could also be automated. We consider the ability to explicitly perform style transfers as a strength, as it allows meta-learning or the oracle to control the process more accurately, instead of happening in unpredictable ways. 

Our framework in its current state is, however, limited in its ability to handle transparency, refraction and reflection. We believe that this needs to be handled on a meta-level similar to intuitive physics, instead of directly inside our capsules. The objects still get detected, albeit with a different color grading (transparency), wrong position and inverted structure (reflection) or distortions of its shape (refraction), and must be put into the correct context through such a meta-process. 

\subsection{Context}

The context of an object in a scene is often important or even necessary to infer its meaning or its type. A table will have the same appearance if it is in a room or in a miniature play-set, but they represent two very distinct object classes. CNNs are able to infer some context from nearby features on different scales, but it has been observed that this can become problematic on larger distances. Recently, there have been efforts to augment these networks to make them more context-aware \citep{Liu2:2019, Alamri:2019}.

In our framework, context is checked and provided by ancestor capsules. In our table example, a \symb{table} capsule activates but we do not yet know if it is a life-sized or a play-set table. However, depending on the other objects in the scene, \ie, its context, either the \symb{room} or \symb{play-set} parent capsule will activate. This provides us with the answer to the context and what kind of table we were looking at.

A more concrete approach is to use a \index{Context-Sensitive Grammar}context-sensitive grammar as the basis for the neural-symbolic capsules, which we did not explore here. By context-sensitive we mean a grammar that allows production rules of the form
\begin{equation}\label{eq:ContextSensitive}
r : \lambda_l\Omega\lambda_r \to \lambda_1\cdots\lambda_n  \;\;\;\; \text{where} \; \Omega\in V, \lambda\in \bigcup_{l,r,i\in\mathbb{N}\setminus\{0\}}(V\cup\Sigma) \;\;\;,
\end{equation}
where the left hand side needs a certain context given by $\lambda_l$ and/or $\lambda_r$ for the production rule to be valid. Such a grammar would result in a capsule's route not just checking if all parts are in the correct configuration, but also if the route itself fits into the current context. While this is certainly sensible, it is somewhat redundant in the feed-forward process of our capsules. In our context-free construction, the aforementioned route's check is performed by one of its ancestor capsules instead and even though our grammar is "context-free", a lot of the important context is still encoded this way. Yet, in a feed-backward operation, context-sensitivity does indeed become relevant, as it makes the choice of rule/route less ambiguous during a full rendering, producing more accurate images. 

Furthermore, context provides additional semantic information by enabling the capsule network to fill in the gaps. As an example, consider a scene with \symb{house}\symb{car}\symb{house}. If the two \symb{house} symbols describe buildings in a wealthy neighborhood using its routes, it is expected that the \symb{car} should also be using a route describing an expensive variety. In case the \symb{car} should be occluded, through context-sensitivity it is possible to infer properties that might not be visible. This is in a sense akin to the ability to piece together a full picture from incomplete information, \ie, \index{Autoassociative Memory}autoassociative memory. In our context-free grammar, we are also able to infer such properties if they are encoded in an attribute \attr{expensive}, but not if it is described by a route itself. However, in spite of this we decided not to explore the option of context-sensitive grammars, as it would make the discussion of our approach much more complex.

\section{The Binding Problem}

In cognitive science, the \index{Binding Problem}binding problem \citep{Revonsuo:1999} plays an important role and has also gained some interest in the field of deep learning. Here, we address a part of the binding problem known as the segregation problem, which asks the following question:

\begin{quote}
Which internal neurons or mechanisms encode the features of external physical objects?
\end{quote}

This is similar to explainability in deep learning with a focus on identifying regions in the network that are specific to an object, as discussed in Section 8.1.1. However, it goes a step further, as the binding problem incorporates all sensory inputs. Two individual neural networks, one for image classification and one for sound classification, will even when running at the same time have their own understanding of the concept "dog" and there is no binding between the two. In such a case, the binding is performed using human input in hindsight.

We have addressed the binding problem early in the design of the capsules by letting each represent one specific symbol. As an example, let us assume we have a primitive capsule for every character \symb{a} to \symb{z}. We require a semantic capsule for every word in a text. Yet, instead of adding a new capsule for a new word, we may also add it as a route to the existing semantic capsule that represents the object. For example, a \symb{cat} capsule may have a route for its visual representation and one for its text representation. The upside of this is that we have an explicit correspondence between the word and its image, something that is lacking in current neural network approaches. We may even go a step further and name the capsule itself by its text representation instead of querying the oracle, strengthening the autonomy of our meta-learning pipeline, as there is no need to ask for a name for every object. 

Further, we may also add routes to a capsule that take an audio signal as input. A cow's "moo" and its image are then processed in two different routes of the same \symb{cow} capsule, which effectively binds these two concepts to the same physical entity. This is also the reason why we chose to incorporate the possibility of different inputs to our framework in Figure \ref{fig:AIArchitectureIntro} in the introduction of this thesis. 

\section{Training}

Training a neural network usually requires vast amounts of labeled data. Once it is finished, the network can be used in a production setting. For CNNs used in image classification tasks, this usually means fixing it in its current state and, thus, keeping it from learning anything new. Methods have been derived to overcome this problem, such as transfer learning \citep{Tan:2018} and \index{Lifelong Machine-Learning}lifelong machine-learning \citep{Parisi:2019}. However, most of these methods require large amounts of fully annotated data for continued learning and, generally, are not capable to learn directly from their environment.

Our neural-symbolic capsule network is designed to learn primarily through observation of its environment and the interaction with the oracle in the early phase. Through the proposed augmentation strategy we are able to do this with a \index{Few-Shot Learning}one-shot / few-shot approach. Learning through observation of the environment is neither better nor worse than through pre-annotated data. On one hand, we forgo the costly process of hand-labeling data, on the other hand, we need to have an oracle present and a predefined set of primitive capsules. \citet{Simard:2017} propose that the future of machine learning lies in simplifying the teaching process and making it accessible to non-experts. For this, our framework would be very suitable, as it asks simple-to-answer questions and handles all complex issues internally. Depending on the application, learning from large amounts of pre-annotated data might be preferred over the interaction with an oracle or vice versa. For example, a classifier intended to process well-defined data will not benefit as much from our approach compared to a robotics system that may find itself in unexpected situations.  

One major hindrance for any neural network that attempts lifelong learning is \index{Catastrophic Interference}catastrophic interference \citep{McCloskey:1989, Ratcliff:1990}, \ie, the possibility of unlearning previous behavior through further training. A capsule's route based on neural networks is as susceptible to this phenomenon as any other network. However, as we only employ small neural networks and each capsule acts in isolation, capsules are only affected on an individual basis instead of on the network level and we can forgo the problem of catastrophic forgetting in an economical way, by retraining it from scratch or through \index{Rehearsing}rehearsing \citep{Robins:1995, Robins:1996}. 

\section{Performance}
 
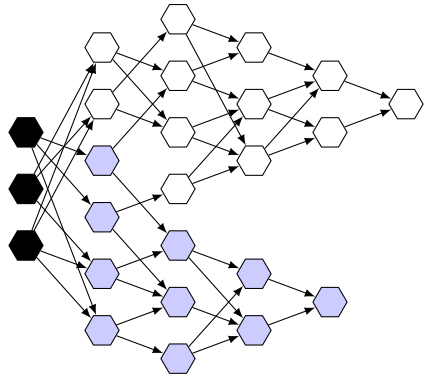
\begin{figure}[H]
	\centering
	\begin{adjustbox}{max width=1.0\textwidth}
		\begin{tikzpicture}
		
		%- Primitive Capsules
		\node (C01) [regular polygon,regular polygon sides=6, draw=black, fill=black] at (0,0.75) {};
		\node (C02) [regular polygon,regular polygon sides=6, draw=black, fill=black] at (0,0.0) {};
		\node (C03) [regular polygon,regular polygon sides=6, draw=black, fill=black] at (0,-0.75) {};
		
		% First Layer
		\node (C11) [regular polygon,regular polygon sides=6, draw=black, fill=blue!20!white] at (1,-1.875) {};
		\node (C12) [regular polygon,regular polygon sides=6, draw=black, fill=blue!20!white] at (1,-1.125) {};
		\node (C13) [regular polygon,regular polygon sides=6, draw=black, fill=blue!20!white] at (1,-0.375) {};
		\node (C14) [regular polygon,regular polygon sides=6, draw=black, fill=blue!20!white] at (1,0.375) {};
		\node (C15) [regular polygon,regular polygon sides=6, draw=black, fill=white] at (1,1.125) {};
		\node (C16) [regular polygon,regular polygon sides=6, draw=black, fill=white] at (1,1.875) {};
		
		% Second Layer
		\node (C21) [regular polygon,regular polygon sides=6, draw=black, fill=blue!20!white] at (2,-2.25) {};
		\node (C22) [regular polygon,regular polygon sides=6, draw=black, fill=blue!20!white] at (2,-1.5) {};
		\node (C23) [regular polygon,regular polygon sides=6, draw=black, fill=blue!20!white] at (2,-0.75) {};
		\node (C24) [regular polygon,regular polygon sides=6, draw=black, fill=white] at (2,0.0) {};
		\node (C25) [regular polygon,regular polygon sides=6, draw=black, fill=white] at (2,0.75) {};
		\node (C26) [regular polygon,regular polygon sides=6, draw=black, fill=white] at (2,1.5) {};
		\node (C27) [regular polygon,regular polygon sides=6, draw=black, fill=white] at (2,2.25) {};
		
		% Third Layer
		\node (C31) [regular polygon,regular polygon sides=6, draw=black, fill=blue!20!white] at (3,-1.875) {};
		\node (C32) [regular polygon,regular polygon sides=6, draw=black, fill=blue!20!white] at (3,-1.125) {};
		\node (C33) [regular polygon,regular polygon sides=6, draw=black, fill=white] at (3,0.375) {};
		\node (C34) [regular polygon,regular polygon sides=6, draw=black, fill=white] at (3,1.125) {};
		\node (C35) [regular polygon,regular polygon sides=6, draw=black, fill=white] at (3,1.875) {};
		
		% Fourth Layer
		\node (C41) [regular polygon,regular polygon sides=6, draw=black, fill=blue!20!white] at (4,-1.5) {};
		\node (C42) [regular polygon,regular polygon sides=6, draw=black, fill=white] at (4,0.75) {};
		\node (C43) [regular polygon,regular polygon sides=6, draw=black, fill=white] at (4,1.5) {};
		
		% Fifth Layer
		\node (C51) [regular polygon,regular polygon sides=6, draw=black, fill=white] at (5,1.125) {};

		\draw[tochild, black] (C01) -- (C11);
		\draw[tochild, black] (C01) -- (C13);
		\draw[tochild, black] (C01) -- (C14);
		\draw[tochild, black] (C02) -- (C12);
		\draw[tochild, black] (C02) -- (C15);
		\draw[tochild, black] (C02) -- (C16);
		\draw[tochild, black] (C03) -- (C11);
		\draw[tochild, black] (C03) -- (C12);
		\draw[tochild, black] (C03) -- (C15);
		\draw[tochild, black] (C03) -- (C16);
		
		\draw[tochild, black] (C11) -- (C21);
		\draw[tochild, black] (C11) -- (C22);
		\draw[tochild, black] (C12) -- (C22);
		\draw[tochild, black] (C12) -- (C23);
		\draw[tochild, black] (C13) -- (C22);
		\draw[tochild, black] (C13) -- (C24);
		\draw[tochild, black] (C14) -- (C23);
		\draw[tochild, black] (C14) -- (C26);
		\draw[tochild, black] (C15) -- (C25);
		\draw[tochild, black] (C15) -- (C27);
		\draw[tochild, black] (C16) -- (C25);
		\draw[tochild, black] (C16) -- (C26);
		
		\draw[tochild, black] (C21) -- (C31);
		\draw[tochild, black] (C21) -- (C32);
		\draw[tochild, black] (C22) -- (C31);
		\draw[tochild, black] (C23) -- (C31);
		\draw[tochild, black] (C23) -- (C32);
		\draw[tochild, black] (C24) -- (C33);
		\draw[tochild, black] (C24) -- (C34);
		\draw[tochild, black] (C25) -- (C34);
		\draw[tochild, black] (C25) -- (C33);
		\draw[tochild, black] (C26) -- (C35);
		\draw[tochild, black] (C26) -- (C34);
		\draw[tochild, black] (C27) -- (C33);
		\draw[tochild, black] (C27) -- (C35);

		\draw[tochild, black] (C31) -- (C41);
		\draw[tochild, black] (C32) -- (C41);
		\draw[tochild, black] (C33) -- (C42);
		\draw[tochild, black] (C33) -- (C43);
		\draw[tochild, black] (C34) -- (C42);
		\draw[tochild, black] (C34) -- (C43);
		\draw[tochild, black] (C35) -- (C43);
		
		\draw[tochild, black] (C42) -- (C51);
		\draw[tochild, black] (C43) -- (C51);
		\end{tikzpicture}
	\end{adjustbox}
	\caption{Only a part of the capsule network activates for each scene and needs computational resources. Primitive capsules are shown in black, activated semantic capsules in blue.} \label{fig:Performance}
\end{figure}

As \index{Performance}our meta-learning pipeline adapts to a wide range of different scenes throughout its lifetime, the capsule network will add more and more semantic capsules. For a wider variety of scenes a network will grow both in width and depth, clustering related concepts (\cf Figure \ref{fig:Performance}). For a particular scene, only capsules that are relevant to it need to activate which reduces the overall amount of computation required. This is in contrast to a block-style end-to-end network of neurons, where every node needs to be considered, even if some regions only give a negligible contribution to the result.

On the other hand we note, that while a classical neural network requires the computation of all nodes, it does not necessarily make it slower or use more resources than our framework. As there is essentially no branching taking place during execution, it is GPU-friendly and can be broken down mainly into simple linear algebra operations. With our framework, we do not only require branching operations, but also (in the case of a single GPU) context-switching between rendering and compute capabilities, as well as potentially switching to different kernels for each capsule. Here, our neural-symbolic capsules benefit from a multi-GPU setup not just from a raw computational perspective, but also by being able to dedicate specific tasks to each node.

\clearpage

\chapter{Conclusion}\thispagestyle{empty}

In this thesis, we presented a novel framework for mental simulation based on our interpretation of capsules using a neural-symbolic hybrid method that addresses the problems outlined in the introduction and repeated in the following. Through our VividNet implementation, we showed that our approach works on toy examples and is able to infer both the semantic information found in the scene as well as physical interactions (Problem 1: Unified Semantic Information). Based on the extracted knowledge, it is able to simulate the environment in different settings given by our querying algorithm (Problem 2: Querying and Simulation). Through the entire process, the binding between the semantic and visual representation is always clear and comprehensible (Problem 3: Explainability). We also showed that the meta-learning algorithm is capable of learning new semantics at any point in time by either inferring it automatically or asking the oracle about unknown elements in the scene (Problem 4: Lifelong Meta-Learning). Finally, we discussed the advantages and shortcomings of our approach and compared it, where possible, with classical neural networks.

Our neural-symbolic framework also tackles some of the other problems that plague deep learning in a novel way. We showed how our meta-learning process avoids the need for large amounts of data and how it can reduce the effects of catastrophic forgetting. Through the versatile nature of our capsule's routes, we were able to tackle the binding problem but also could explore how our framework is naturally invertible to be used both for rendering as well as inverse-graphics.

Looking forward, the true test for our framework would be to integrate it with an algorithm capable of replacing the oracle, such as the Rosie Soar cognitive architecture \citep{Laird:2012, Kirk:2014}, an imagination-based planner \citep{Pascanu:2017} or more general reinforcement learning algorithms. In such a pairing, the framework could potentially become fully unsupervised in its learning behavior. It would also be interesting to explore how the querying process can be further automated using recurrent neural networks, such as a neural Turing machine \citep{Graves:2014}, or if it can be paired with a partially observable Markov decision process \citep{Hamrick2:2017}. Further, disentangled variational auto-encoders \citep{Kingma:2014, Higgins:2017} could provide an interesting alternative for the capsule's internal encoder-decoder pair we currently employ. Also, our implementation has mainly focused on 2D toy examples and it would be interesting to see its performance on 3D datasets, such as CLEVR \citep{Johnson:2017}.

Finally, we believe that our approach is a suitable interface for any agent with its environment, so that this agent can perform all its actions based on purely symbolic information, \ie, as if in a classic symbolic artificial intelligence setting. As such, we hope to help bridge the gap between the connectionist and the symbolic approach to artificial intelligence.

\bibliographystyle{abbrvnat}
\bibliography{egbib}

%\printindex

\end{document}